\documentclass{article}

\usepackage{microtype}
\usepackage{graphicx}
\usepackage{subcaption}
\usepackage{booktabs} 

\usepackage{hyperref}

\usepackage[accepted]{icml2025}

\usepackage{amsmath}
\usepackage{amssymb}
\usepackage{mathtools}
\usepackage{amsthm}

\usepackage[capitalize,noabbrev]{cleveref}

\usepackage[space]{grffile}     

\usepackage{wrapfig}
\usepackage{algorithm}
\usepackage{algorithmic}
\usepackage[export]{adjustbox}
\usepackage[para]{footmisc}
\graphicspath{ {./figures/} }
\usepackage{hyperref}       

\theoremstyle{plain}

\theoremstyle{definition}

\theoremstyle{remark}

\usepackage[textsize=tiny]{todonotes}

\icmltitlerunning{Aligning Agents like Large Language Models}

\begin{document}

\twocolumn[
\icmltitle{Aligning Agents like Large Language Models}




\begin{icmlauthorlist}
\icmlauthor{Adam Jelley}{uoe}\hspace{-1mm}\textsuperscript{\dag}\hspace{1mm}
\icmlauthor{Yuhan Cao}{msr}
\icmlauthor{Dave Bignell}{msr}
\icmlauthor{Amos Storkey}{uoe}
\icmlauthor{Sam Devlin}{msr}
\icmlauthor{Tabish Rashid}{msr}
\end{icmlauthorlist}

\icmlaffiliation{uoe}{School of Informatics, University of Edinburgh, Edinburgh, United Kingdom}
\icmlaffiliation{msr}{Micosoft Research, Cambridge, United Kingdom}

\icmlcorrespondingauthor{Adam Jelley}{adam.jelley@ed.ac.uk}

\icmlkeywords{Machine Learning, ICML}

\vskip 0.2in
]



\printAffiliationsAndNotice{}  

\begin{abstract}
Training agents to act competently in complex 3D environments from high-dimensional visual information is challenging. Reinforcement learning is conventionally used to train such agents, but requires a carefully designed reward function, and is difficult to scale to obtain robust agents that generalize to new tasks. In contrast, Large Language Models (LLMs) demonstrate impressively general capabilities resulting from large-scale pre-training and post-training alignment, but struggle to act in complex environments. This position paper draws explicit analogies between decision-making agents and LLMs, and argues that agents should be trained like LLMs to achieve more general, robust, and aligned behaviors. We provide a proof-of-concept to demonstrate how the procedure for training LLMs can be used to train an agent in a 3D video game environment from pixels. We investigate the importance of each stage of the LLM training pipeline, while providing guidance and insights for successfully applying this approach to agents. Our paper provides an alternative perspective to contemporary LLM Agents on how recent progress in LLMs can be leveraged for decision-making agents, and we hope will illuminate a path towards developing more generally capable agents for video games and beyond. Project summary and videos: \url{https://adamjelley.github.io/aligning-agents-like-llms}.
\end{abstract}

\section{Introduction}
\label{introduction}

The optimal approach for training agents to act in complex 3D environments without well-defined reward functions is an open question. Many modern video games provide such environments, where the state space, action space, and time-horizons are large enough that reinforcement learning (RL) from scratch is usually infeasible. However, even if possible, the resulting agent would not generalize to new objectives or environments. In contrast, large language models (LLMs) demonstrate impressively general capabilities resulting from large-scale pre-training, but struggle to act in complex low-level environments \citep{paglieri_balrog_2024}. 

\begin{figure}[t]
  \centering
  \includegraphics[width=\columnwidth]{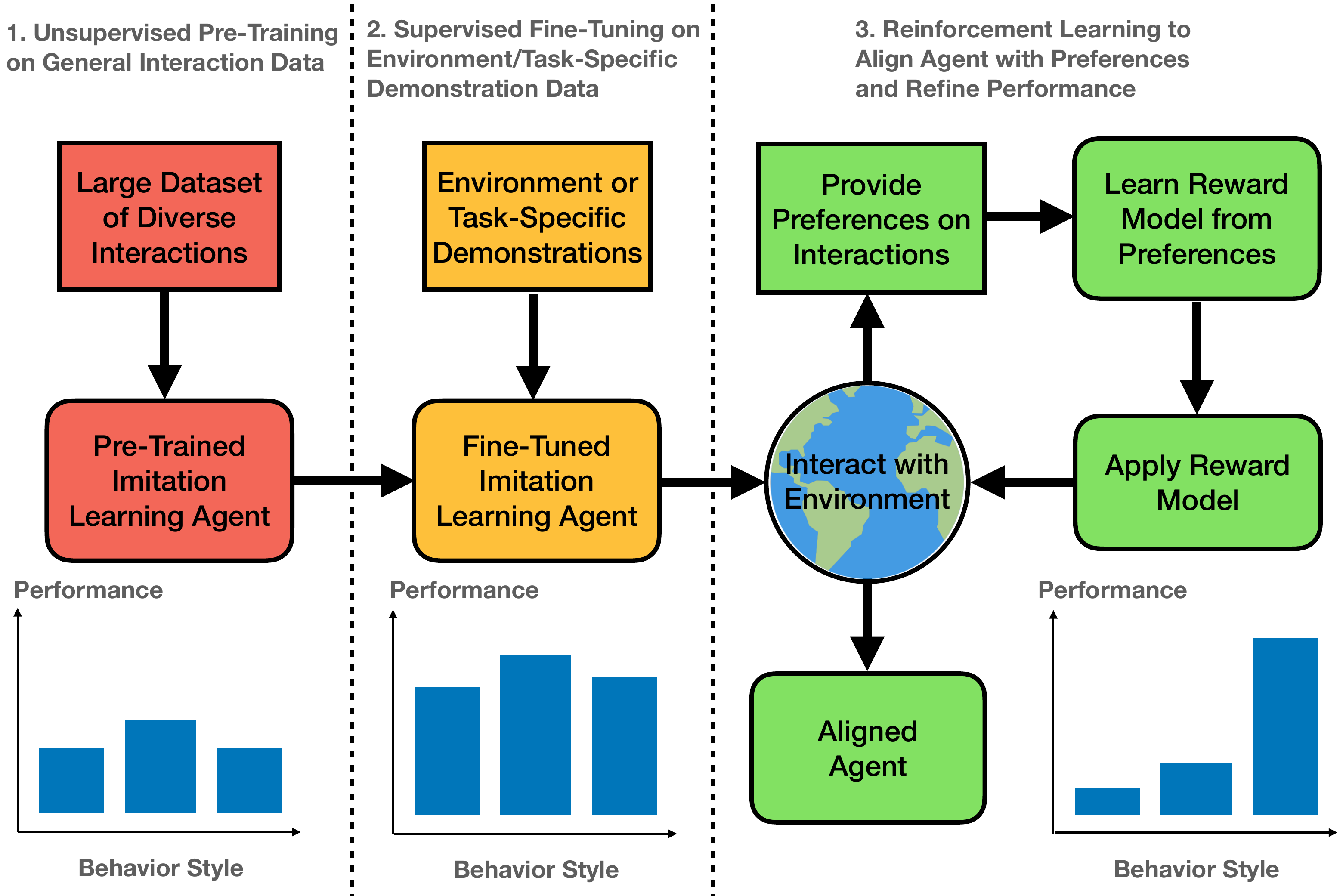} 
  \caption[Short Figure 1 caption]%
  {Illustration of our approach for training general agents in complex environments. Pre-training on a large, diverse dataset of interactions provides a base agent which can be more effectively fine-tuned with limited  demonstrations. This agent can be further refined with RL, using a reward model or external reward function. This approach is analogous to the training of modern LLMs.} 
  \label{fig:Figure1}
  \vspace{-4mm}
\end{figure}

An analogous approach in the context of training agents is to leverage a large dataset of general human gameplay to pre-train an agent with imitation learning (IL). This provides an agent with a general understanding of player behavior, and there is evidence of generalization benefits from scaling up data and compute \citep{lee_multi-game_2022, baker_video_2022, reed_generalist_2022, brohan_rt-2_2023, pearce_scaling_2024}. 

 However, such an agent will inevitably learn to imitate all behaviors found within the human gameplay, including undesirable behaviors of novice or malicious players, analogous to unhelpful or toxic responses of unaligned LLMs \citep{ziegler_fine-tuning_2020}. Additionally, game designers may want this general agent to perform a particular role, or act with a certain style or strategy \citep{aytemiz_acting_2021, devlin_navigation_2021}. There is a parallel between making such an agent useful and aligning LLMs. In the same way that general LLM base models have been aligned for various purposes \citep{banerjee_benchmarking_2023, spatharioti_comparing_2023, chen_evaluating_2021}, we can imagine aligning a general decision-making agent with various objectives or environments.
\newpage
\textbf{This position paper argues that decision-making agents should be trained in the same manner as large language models (LLMs) to obtain analogous generality and robustness to objectives and environments.} Specifically, agents should be trained and refined over multiple stages, typically involving: 1) \textit{Unsupervised pre-training} on general behavior or gameplay data to obtain a general behavioral prior, 2) \textit{Supervised fine-tuning} (SFT) on higher quality demonstration data of the relevant task or environment, 3) \textit{Reinforcement learning} (RL), potentially from human feedback (RLHF), to refine and improve the behavior or further align the agent with implicit behavior preferences. To provide evidence for our position, we consider an academically illustrative part of a modern video game where the human behavior distribution is distinctly multi-modal, but we desire our agent to imitate a single mode of this distribution. This particular mode of behavior is difficult to learn from scratch with RL from pixels due to the size of the exploration space, and there are insufficient clean demonstrations of this mode for robust imitation learning. We demonstrate that this behavior mode can be reliably achieved by following the modern LLM training pipeline, and illustrate the importance of each stage. We train our agent from high-dimensional visual input with full game controller action output, maintaining the generality of our approach to other game environments, and demonstrating that the modern LLM training paradigm can be successfully applied to training agents in complex 3D environments. Our proof of concept provides evidence that many of the recent findings in the training of LLMs do indeed transfer to training agents, and could enable the training of more general and robust decision-making agents. 

\section{Context and Background}

\textbf{Large Scale Imitation Learning:}
Scaling up imitation learning and offline reinforcement learning \citep{levine_offline_2020} has recently gained popularity with the aim of demonstrating similar scaling laws to LLMs \citep{kaplan_scaling_2020, hoffmann_training_2022} to obtain more general agents. For example, Decision Transformer \citep{chen_decision_2021} proposed imitation learning with a transformer policy that can be conditioned on a desired return. Multi-Game Decision Transformers \citep{lee_multi-game_2022} extended this approach to learn multiple game policies, and GATO \citep{reed_generalist_2022} demonstrated that a single model could learn hundreds of diverse tasks. RT1 \citep{brohan_rt-1_2022} shows similar scaling potential in robotics tasks, while RT2 \citep{brohan_rt-2_2023} integrated vision-language models and RoboCat \citep{bousmalis_robocat_2023} demonstrated promising adaptation to new tasks. SIMA \citep{sima-team_scaling_2024} incorporates language conditioning to enable generalization to new video games. Recently, \citet{pearce_scaling_2024} have confirmed the hypothesized scaling laws for imitation learning. While these approaches show the potential for scaling imitation learning, we argue that these models are analogous to GPT-2 \citep{radford_improving_2018} in the rise of modern LLMs and that additional alignment stages can similarly be used to specialize and improve these models to make them as useful as modern LLMs.


\textbf{Fine-tuning with Reinforcement Learning:} The use of imitation learning as pre-training for reinforcement learning was first investigated to improve the sample efficiency of deep RL algorithms by reducing the exploration space \citep{hester_deep_2017, vecerik_leveraging_2018}. AlphaGo \citep{silver_mastering_2016} and subsequently AlphaStar \citep{vinyals_grandmaster_2019} demonstrated that large-scale imitation learning could provide a strong behavioral prior for environments like StarCraft where RL from scratch is infeasible. VPT \citep{baker_video_2022} extended this paradigm to internet-scale Minecraft videos using an inverse dynamics model to enable imitation learning, before RL fine-tuning on task-specific rewards. While these works demonstrate the potential of fine-tuning large imitation models, they use hard-coded reward functions to maximize performance rather than align an agent's behavior with new objectives or subjective preferences.

\textbf{Reinforcement Learning from Human Feedback (RLHF):}
Training agents with human preferences has a long history, notably including \citet{knox_tamer_2008} and \citet{bennett_netflix_2009}, which has been reviewed by \citet{wirth_survey_2017} and \citet{zhang_recent_2021}. The popular modern formulation of RLHF for deep learning was proposed by \citet{christiano_deep_2017}. This idea was extended by \citet{ibarz_reward_2018} to include imitation learning as pre-training to improve the efficiency of reward modeling. PEBBLE \citep{lee_pebble_2021} instead uses unsupervised pre-training to increase the diversity of initial behaviors. 
\citet{abramson_improving_2022} and \citet{milani_towards_2023} are closest to our position in using RLHF to refine the performance of large imitation agents in 3D simulated worlds, but do not make an explicit analogy or investigate the full LLM training pipeline. 

\textbf{Large Language and Vision Models:} \citet{radford_improving_2018} demonstrated that pre-training with an unsupervised generative task (next token prediction) on a large diverse corpus of text, followed by fine-tuning on a specific task, led to performance benefits compared to fine-tuning alone. \citet{stiennon_learning_2020}, following \citet{ziegler_fine-tuning_2020}, then popularized the use of RLHF to further fine-tune these models to align their responses with human preferences. This procedure led to InstructGPT \citep{ouyang_training_2022}, Chat-GPT \citep{openai_introducing_2022}, and GPT-4 \citep{openai_gpt-4_2023} as well as a wide range of other popular language models, such as Claude \citep{bai_training_2022}, Llama \citep{touvron_llama_2023} and Gemini \citep{geminiteam2024geminifamilyhighlycapable}. Generative pre-training has been shown to have equivalent benefits for vision tasks \citep{chen_generative_2020}, and RLHF has similarly been used to refine image generation and vision-language models (VLMs) \citep{xu_imagereward_2023, esser_scaling_2024}. More recently, RL with pre-trained LLMs has been shown to improve reasoning capabilities in verifiable domains where external rewards can be defined \citep{openai_openai_2024, deepseek-ai_deepseek-r1_2025}. While components of the LLM training pipeline have been applied both individually and in combination to decision-making agents, in this paper we argue that the full pipeline (outlined in Appendix~\ref{app:procedure}) should be more widely and explicitly applied to agents to achieve similarly general agent models. We now turn to providing evidence to support this position with a concrete proof-of-concept.

\section{Proof of Concept}

\subsection{Environment and Alignment Goal}

As a proof of concept, we use the AAA video game Bleeding Edge\footnote{\url{www.bleedingedge.com/en}}, 
which was launched in 2020 for Xbox One\footnote{\url{www.pcgamingwiki.com/wiki/Bleeding_Edge}}. It is a team-based 4v4 online multi-player video game. The game is played with a third-person view, with the camera angle controlled by the player, so the environment is partially observable. Here, we focus 
on a single map, called Skygarden. This map is spread over three islands, including a main island and two launch islands (one for each team). 

At the beginning of a game, players spawn at one of four points on their team's launch island. From the spawn point, a player must navigate across the launch island to one of three jumppads that launch the player onto the main island. Depending on the jumppad selected, the player will be launched onto a different area of the main island, so players often take different jumppads. For the purpose of this work, we aim to train an agent to navigate across the launch island to a single one of the three jumppads. Since this navigation task requires around 10 seconds to complete for an optimal agent, without access to privileged information such as the agent location to shape a reward function, it would be extremely difficult to train an agent to complete this task with RL from visual input alone, as it would rarely leave the spawn area. On the other hand, an imitation learning agent will match the human distribution to reach all three of the jumppads. Therefore while the task is simple, it provides a clear motivation for alignment and a setting which we can quantitatively analyze to illustrate our position. 

\subsection{Implementation Details}
\label{subsec:environment}

\begin{figure}[h]
  \centering
  \includegraphics[width=0.49\columnwidth]{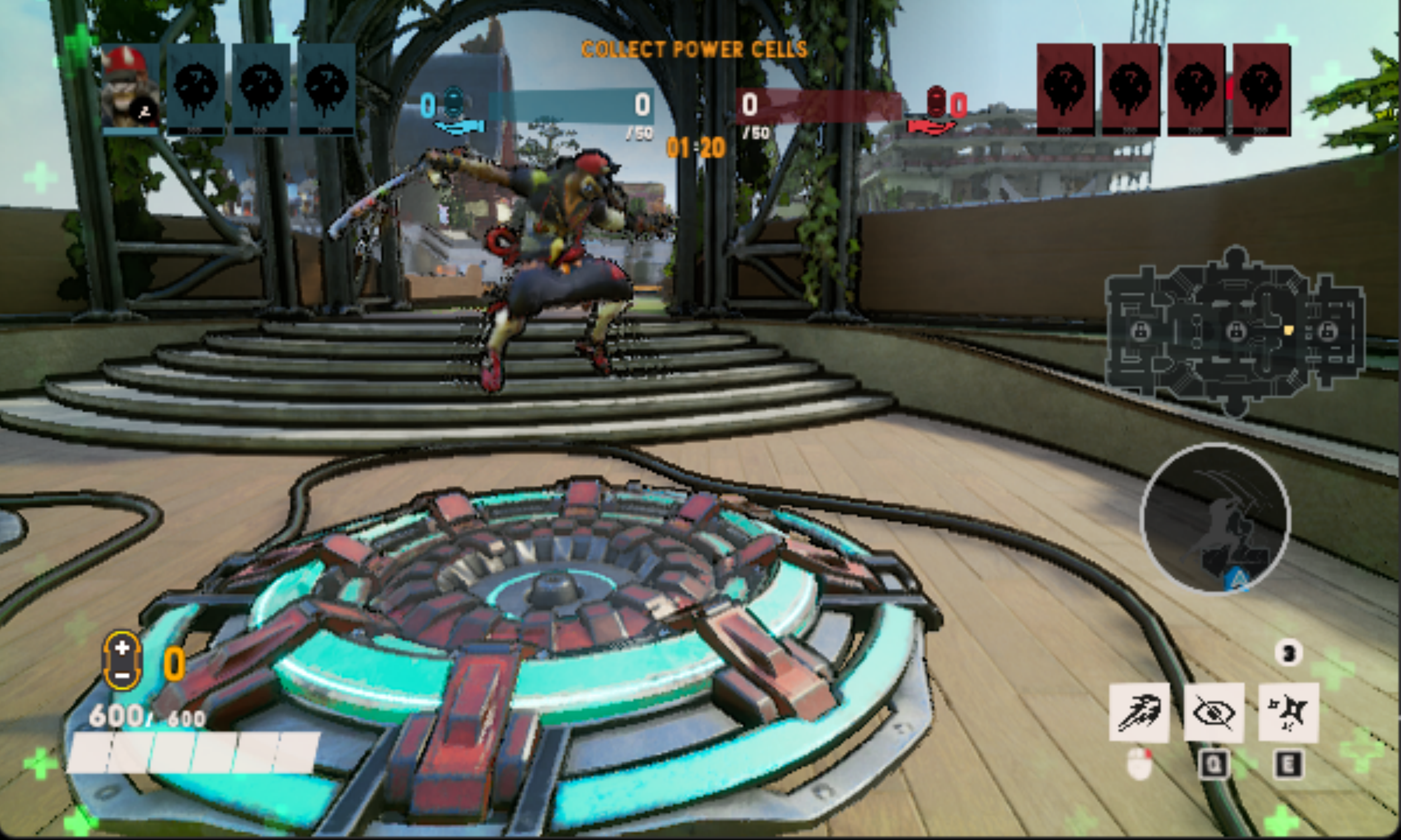} 
  \includegraphics[width=0.49\columnwidth]{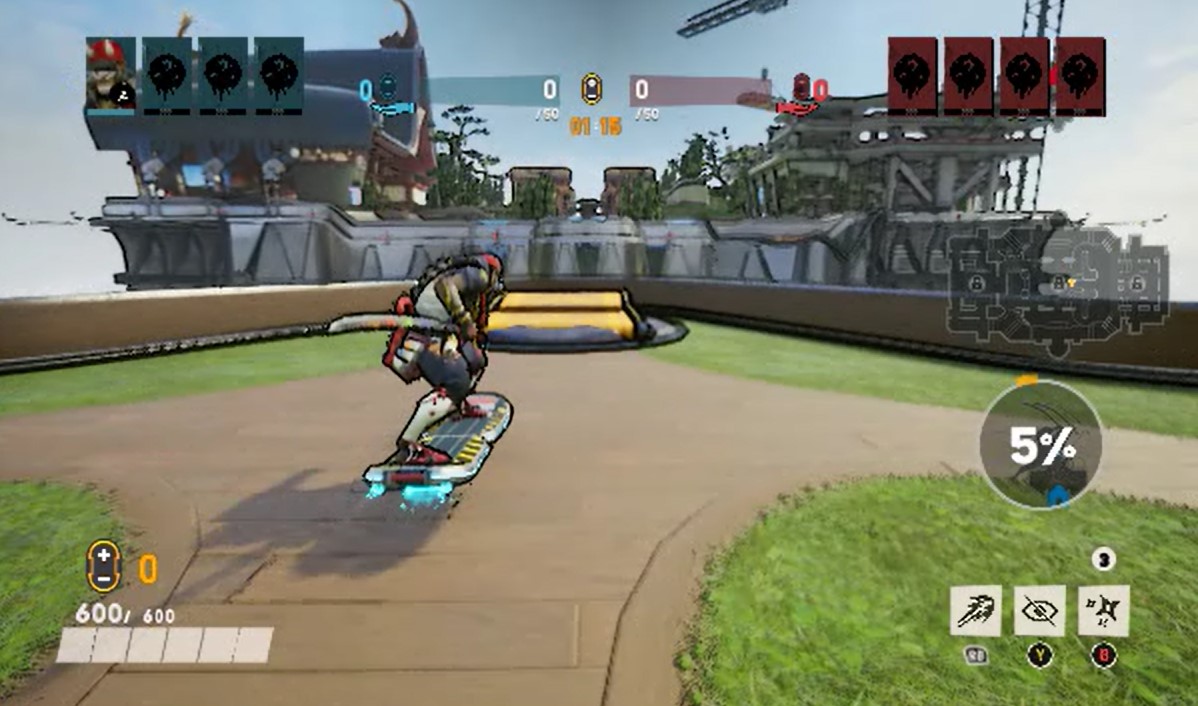} 

  \caption{Screenshots of the Ninja agent at a spawn point and heading towards the middle jumppad of the launch island.} 
  \vspace{-5mm}
  \label{fig:Screenshots}
\end{figure}

\textbf{Observation and Action Spaces:} To maintain the generality of our approach to any 3D environment, we take visual gameplay observations $\mathbf{o}_t \in \mathbb{R}^{H\times W\times 3}$ as input, sampled at 10Hz. We do not provide any privileged information, so the agent only has access to visual information, as shown in Figure~\ref{fig:Screenshots}. 
The action space consists of 12 buttons and 2 joysticks.
The left joystick controls movement, while the right controls the camera angle. 
We decompose each joystick into an $x$ and a $y$ component which are independently discretized into 11 buckets to enable better modeling of multi-modal behavior distributions (compared to directly using continuous values), which is essential for our setting. 

\textbf{Dataset:} For general pre-training, a dataset was extracted from recorded human gameplay, as described in Appendix~\ref{app:data}. This large unfiltered dataset consists of 71,940 individual player trajectories from 8788 matches recorded between 09-02-2020 and 10-19-2022, which amounts to 9875 hours (1.12 years) of individual gameplay.
For the purposes of this work, we use a base agent that was trained for roughly one epoch on this dataset. Considering the success of recent foundation models, we note that given our unified observation and action spaces, it may be possible to instead train across games to obtain this base model, as explored in previous work \citep{reed_generalist_2022, sima-team_scaling_2024}.

\textbf{Architecture and Training:} For the policy, we use a GPT-2 \citep{radford_language_2019} causal transformer architecture with 103M parameters, similar to that used by VPT \citep{baker_video_2022}. Observations from the human gameplay $\mathbf{o}_t \in \mathbb{R}^{H\times W\times 3}$ are taken directly as input to a convolutional encoder to give observation embeddings $\mathbf{z}_t$. 
The transformer is trained 
with a context window of $H=32$ timesteps (around 3s of gameplay given the 10Hz sampling). The context window is important since the game is partially observable: longer context provides the agent with a more Markovian state, but at the cost of quadratic computational complexity. We note that developments in extending the context length of transformers, such as modified attention mechanisms \citep{beltagy_longformer_2020}, or even Retrieval-Augmented Generation (RAG) \citep{lewis_retrieval-augmented_2021}, could also be investigated here to help address partial observability. The transformer and convolutional encoder are both trained end to end with a cross-entropy loss over all 16 discrete action components to provide a policy $\pi(a_t|o_{t},a_{t-1}...o_{t-H})$. While we only consider prior observation and action context for simplicity in this proof-of-concept, text embeddings $e$ corresponding to game or task instructions (likely to be an integral part of a general agent) could also be straightforwardly included as conditioning here, providing an instruction-conditioned policy $\pi(a_t|o_{t},a_{t-1}...o_{t-H};e_t)$ similar to SIMA \citep{sima-team_scaling_2024}. Full architecture and training details are provided in Appendix \ref{app:details}.


\textbf{Evaluation:} To evaluate our models, we run each agent online in the game environment and record which jumppad is reached over 1000 episodes. To run an agent online, we initialized the \textit{Ninja} character with an empty context buffer at one of the spawn points at random. We run the agent until it reaches one of the jumppads or times out.

\subsection{Does large scale pre-training provide generalization benefits in the context of agents?}
\label{subsec:pretraining}

\begin{figure}[h!]
  \centering
  \includegraphics[width=\columnwidth]{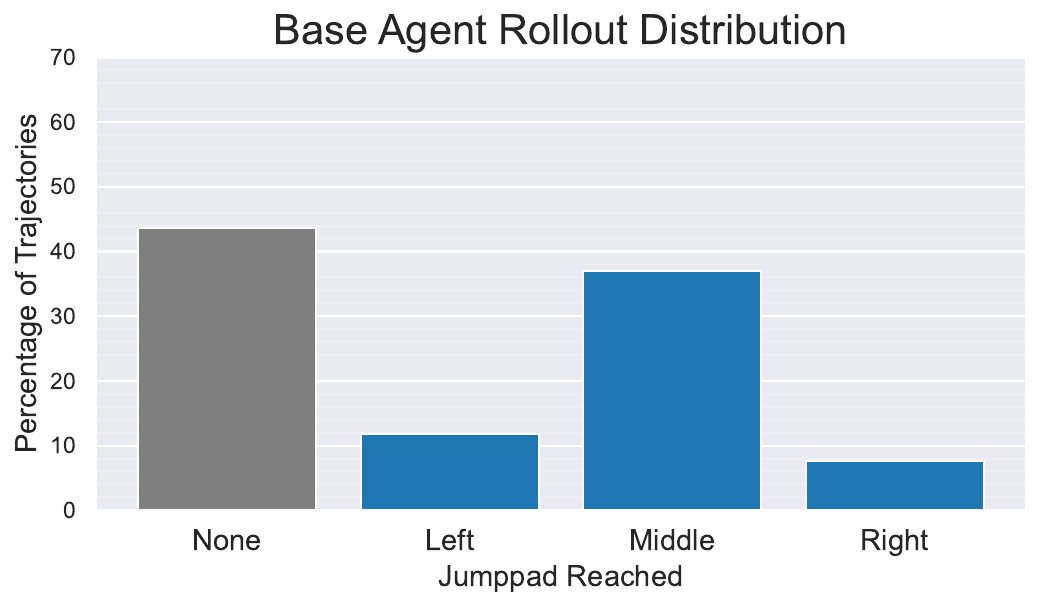}
  \caption{Distribution of jumppads reached by the base agent.}
  \label{fig:basejumppads}
\end{figure}

We begin by evaluating our base model, which achieves the jumppad distribution shown in Figure~\ref{fig:basejumppads}. We see that our agent fails to reach a jumppad $44\%$ of the time, and has a bias towards the middle jumppad, reflecting human behavior. Our base model therefore provides a reasonable behavioral prior, and is much better than a randomly initialized agent (which has a $0\%$ success rate). However, the success rate could be improved, which is partly due to distribution shift between the offline data and the online environment. For example, the base model was trained on data containing all thirteen possible characters, while our online evaluation only uses the default \textit{Ninja} character. Additionally, we initialize our agents online at the spawn point with an empty context buffer, while in training the agent will have access to context including a spawn animation and prior gameplay. While measures can be taken to avoid distribution shift, offline-only learning is inherently challenging \citep{ostrovski_difficulty_2021} as discussed in Appendix \ref{app:o2oshift}, demonstrating the need for some online behavior refinement (see Section \ref{subsec:alignment}).

\begin{figure}[h!]
  \centering
  \includegraphics[width=\columnwidth]{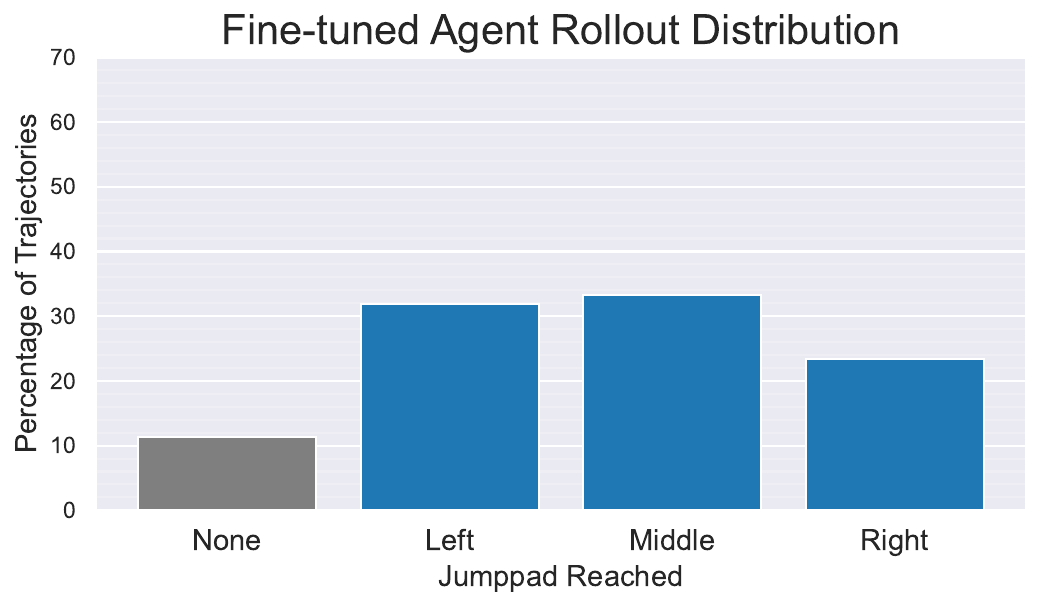}
  \caption{Distribution of jumppads reached by the base agent after supervised fine-tuning (SFT) on a task-specific dataset.}
  \label{fig:FTjumppads}
\end{figure}

Following the LLM pipeline, we now fine-tune our agent on a smaller curated dataset. We curated 300 successful trajectories (100 per jumppad) for fine-tuning, all of which involved the character Ninja and were filtered to contain only the first part of the trajectory until a jumppad was reached. We fine-tune until convergence as described in Appendix~\ref{app:details}. More generally, this stage could consist of fine-tuning on any smaller, high-quality demonstration data. We see in Figure \ref{fig:FTjumppads} that the success rate of reaching a jumppad is now substantially higher (with only an $11\%$ failure rate), and that the agent more evenly reaches all three jumppads (corresponding to the balanced fine-tuning dataset).

At this stage, it is natural to ask whether the general pre-training was beneficial or whether the model could have been trained from scratch on our task-relevant dataset directly. After all, if clean demonstrations of the desired behavior are available, why train the agent on messy, less-relevant data? To investigate this, we train an equivalent architecture model from scratch until convergence on the fine-tuning dataset only (see Appendix \ref{app:details} for details). The evaluation of this model is shown below in Figure \ref{fig:ablationFTjumppads}.

\begin{figure}[h]
\vspace{0mm}
    \includegraphics[width=\columnwidth]{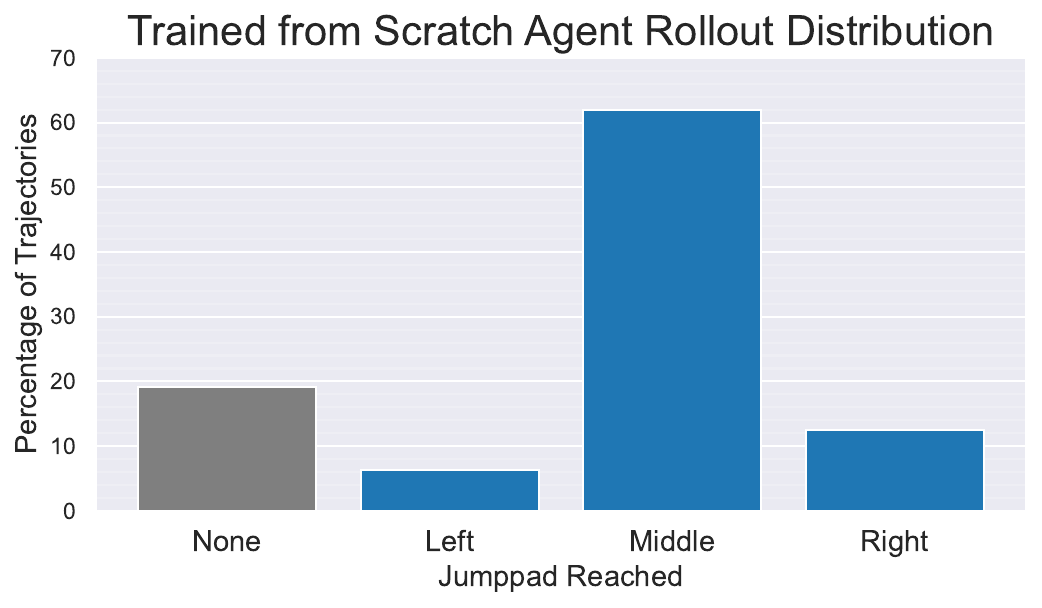}

      \vspace{0mm}
      \caption{Distribution of jumppads reached by an equivalent agent trained from scratch on the task-specific dataset used for supervised fine-tuning.}
\vspace{0mm}
\label{fig:ablationFTjumppads}
\end{figure}

Perhaps surprisingly, we find that the agent without pre-training has a less diverse jumppad distribution, and a higher failure rate (around $20\%$, almost twice that of the pre-trained agent). This finding seems to go against conventional wisdom that only clean, relevant demonstrations should be used from the available data. Instead,  these results corroborate similar findings for LLMs, in which pre-training on diverse data has been shown to enable more effective fine-tuning, effectively increasing the size of the fine-tuning data compared to training from scratch \citep{hernandez_scaling_2021}.

\begin{figure}[h]
  \centering
  \includegraphics[width=0.49\columnwidth]{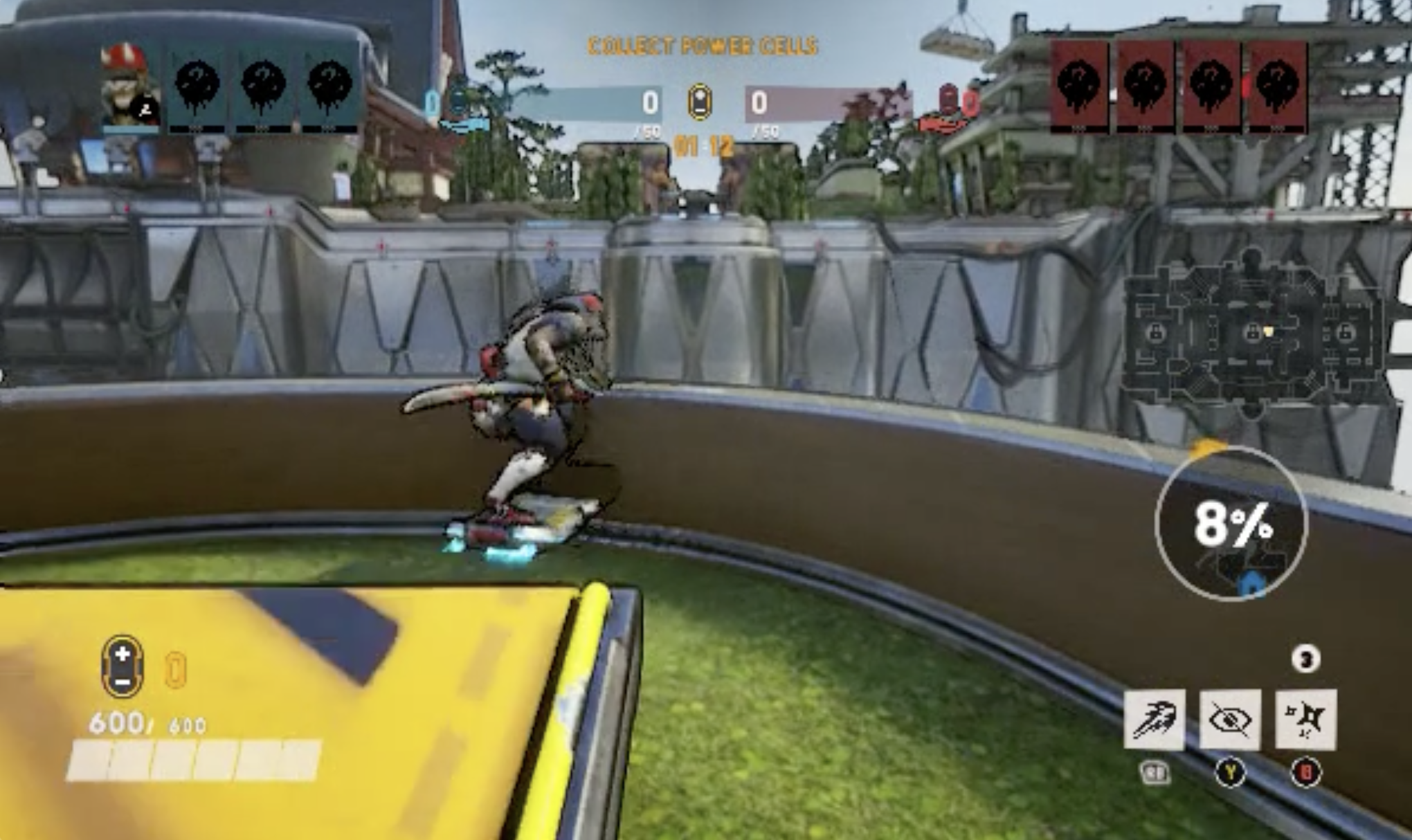} 
  \includegraphics[width=0.49\columnwidth]{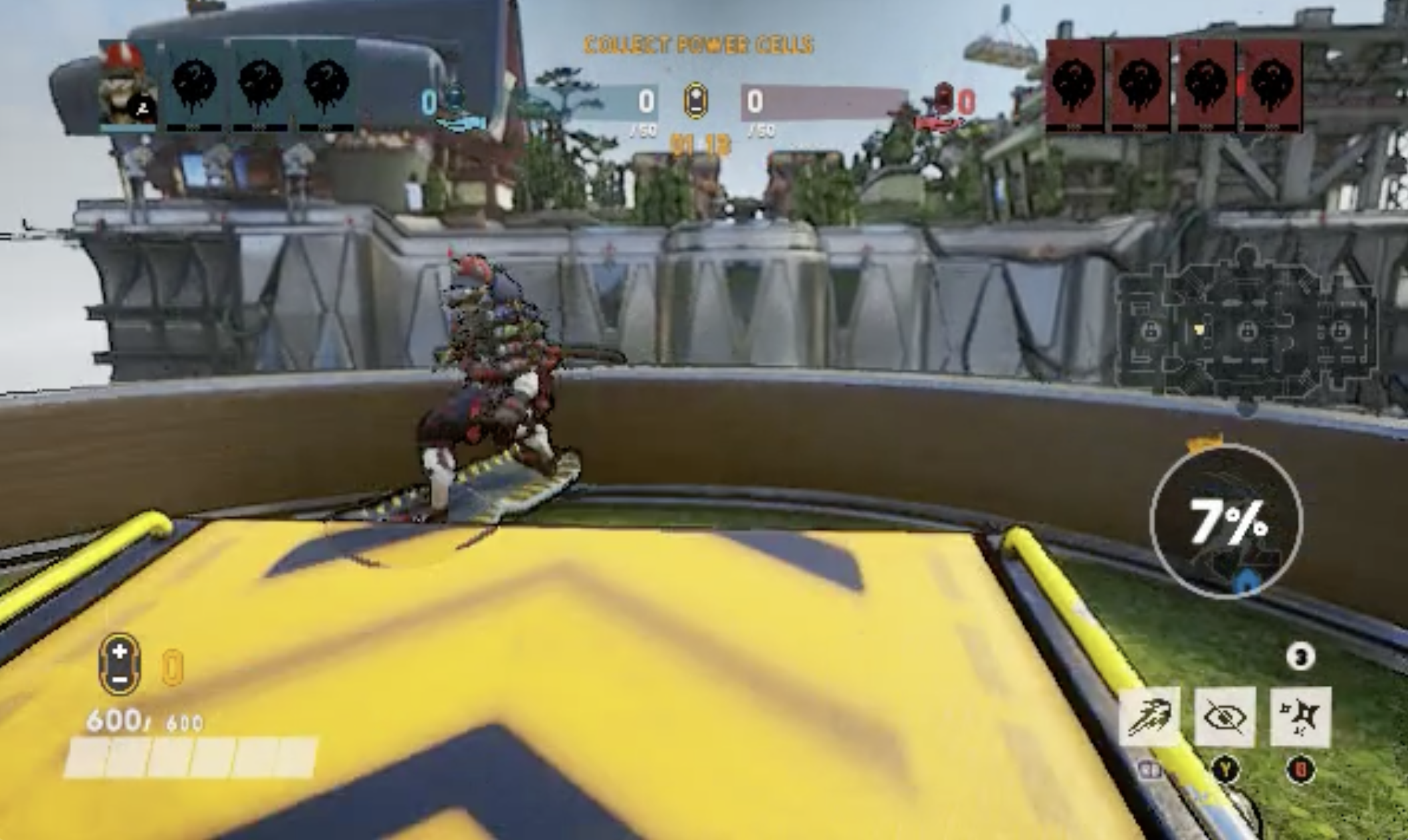} 

  \caption{Screenshots of the agent after having missed the intended jumppad. Left: Agent without pre-training continues into wall. Right: Agent with pre-training turns around to return to jumppad.} 
  \label{fig:agents_hitting_wall}
\end{figure}

To understand the reason behind this difference, we reviewed videos of the two agents navigating towards the jumppads. We noticed that failures would often arise when an agent accumulates action error over its trajectory, a common challenge with imitation learning \citep{ross_efficient_2010}, causing it to miss the intended jumppad as shown in Figure~\ref{fig:agents_hitting_wall}. Since the fine-tuning dataset consists of only a small number of trajectories that hit the jumppad directly, the agent trained from scratch (without pre-training) has never seen that observation before, so therefore chooses the modal action (which is to move forwards) as it is now out of distribution. This causes it to run into the wall continuously until it times out, leading to failure (Fig.~\ref{fig:agents_hitting_wall}, left). On the other hand, the pre-trained agent would often hit the wall, and then turn around and hit the jumppad from behind (Fig.~\ref{fig:agents_hitting_wall}, right). This is because the more diverse pre-training data provides a common-sense behavior prior, allowing the agent to return to the fine-tuning distribution. These behaviors are demonstrated in the videos on our project webpage, and provide anecdotal but intuitive evidence for the benefits of unsupervised pre-training in the context of agents.

We also perform a preliminary investigation of model scaling in Appendix \ref{app:scalingablation}, by training additional models with an equivalent transformer architecture from scratch on this smaller task-specific dataset. We find that smaller models have a higher failure rate, providing evidence that larger model sizes lead to improved performance in our imitation learning task, even when trained only on this relatively small fine-tuning dataset. These findings are in agreement with more comprehensive recent work by \citet{pearce_scaling_2024}. Furthermore, we see early indications that larger models also align more efficiently with reinforcement learning.

\textbf{In summary, we find that large-scale pre-training\\ provides general benefits to agents, analogous to LLMs.}

\subsection{Preference Modeling}\label{subsec:preferences}

The next stage in the standard procedure for training LLMs is to obtain human preference data on model responses. Here we deploy the fine-tuned agent in the environment to collect trajectories for preference labelling. This is analogous to generating multiple LLM responses to a prompt, but in this context the prompt becomes the initial observation and optional context of previous observations and actions (or instructional text embeddings) provided. 
Similarly to LLMs, there are tradeoffs to be made in the diversity, quality and quantity of preferences (discussed further in Appendix \ref{app:limitations}), but here we avoid such issues by utilizing synthetic preferences to isolate the effect of quantity. Additionally, this labeling could potentially be automated using recent VLMs, again leveraging ideas and developments from the LLM-as-a-Judge literature \citep{zheng_judging_2023, li_llms-as-judges_2024}.

\textbf{Online Rollouts:} We use the evaluation procedure described in Section~\ref{subsec:environment} to generate a dataset of 2400 on-policy trajectories, divided into 1000 trajectories for training and 1400 for evaluation. Similarly to LLMs, the temperature of the softmax sampling of the policy for action selection can be increased to generate more diverse behaviors from the agent for easier comparison, but here we found the behavior to be diverse enough with a default temperature of one.

\textbf{Generating Preferences:} We use synthetic preferences to rank trajectories based on the primary jumppad (JP) criteria:\newline

\vspace{-1.5mm}
\centerline{\textit{Given JP Reached $\succ$ Other JP Reached $\succ$ No JP Reached}}

Within each of these categories, we further rank by duration, preferring shorter trajectories. 
By selecting subsets of the training trajectories, we can investigate how reward model performance scales with number of comparisons (which is a proxy for human labeling time, often the main bottleneck for RLHF).
A preference dataset $P$ is then created by considering all pairwise comparisons of trajectories $\tau$ using the criteria above, i.e. $(\tau_A \succ \tau_B) \in P \ \forall \ \tau_A, \tau_B \in \tau \text{ if } \tau_A \succ \tau_B$.

\subsection{Do modern reward-modeling practices apply in the context of agents?}
\label{subsec:RM_training}

A reward model is then trained on this dataset to predict rewards in accordance with preferences, as described below. Previous work on RLHF for agents has generally relied on independent (often linear) reward models \citep{knox_tamer_2008, christiano_deep_2017}. However, recent work on LLMs has demonstrated that reward models using the pre-trained or fine-tuned policy model with the action classification head replaced with a scalar regression head generally perform better \citep{stiennon_learning_2020}, and also improve with scale \citep{ouyang_training_2022, touvron_llama_2023}. It is hypothesized that this is because the pre-training enables the reward model to better distinguish behaviors for reward assignment without the trajectory over-fitting that would occur with the equivalent model otherwise \citep{ziegler_fine-tuning_2020}. We investigate this potential for representation transfer in the context of visual input agents in this section.


\textbf{Procedure: } We use the standard Bradley-Terry \citep{bradley_rank_1952} modeling procedure to train a reward model $\hat r$ from these pairwise preferences, following \citet{christiano_deep_2017}. Specifically, we interpret trajectory rewards as preference rankings analogous to Elo ratings \citep{elo_rating_1978}  developed for chess, such that the annotator's probability of preferring a trajectory depends exponentially on the trajectory reward difference.
We can then fit the reward model by minimising the cross-entropy between these probabilities and the preference labels, which gives the loss function:
\begin{equation}\label{eqn:BTloss}
\mathcal{L}(\hat r) = \sum_{(\tau_w, \tau_l)\in P} -\log\bigg( \sigma\big(\hat r(\tau_w) - \hat r(\tau_l)\big)\bigg)
\end{equation}
where $\sigma$ is the sigmoid function and $(\tau_w, \tau_l)$ are the trajectories being compared, with $\tau_w$ being the winning (preferred) trajectory and $\tau_l$ being the losing trajectory. Since the reward model is only trained on comparisons, the scale of the predicted rewards is arbitrary. As a result, we found the need to apply a small amount of $L2$ regularization to prevent the scale of the rewards becoming overly large for the best and worst trajectories. We further empirically normalize the reward model after training using the max and min of predicted rewards over the training trajectories to scale our reward model output to be in the range $\hat r \in [0,1]$.

\textbf{Architecture: } Following the LLM procedure to use the agent model for the reward model initialization, we take the output embeddings of the fine-tuned transformer at each timestep as input to an MLP. This outputs a scalar return for use in the loss function in Equation \ref{eqn:BTloss}.
As a baseline, we considered training a reward model with an equivalent architecture end-to-end, but found this led to trivial over-fitting and negligible validation performance. Instead, we use a frozen randomly initialized encoder to provide embeddings for each timestep, which are fed into an equivalent MLP.

\begin{figure}[h!]
  \centering
  \includegraphics[width=\columnwidth]{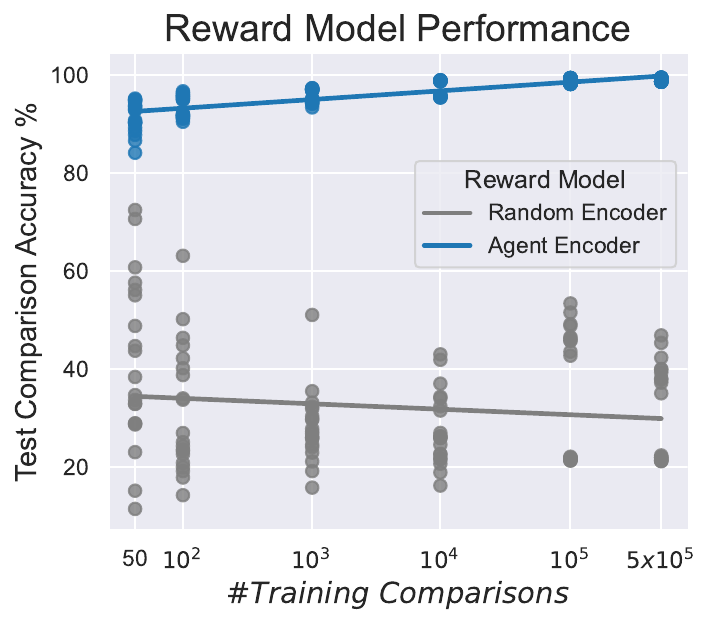}
  \vspace{-6mm}
  \caption{Reward model test accuracies against number of training comparisons for random and agent-initialized models.}
  \label{fig:RMs}
  \vspace{-3mm}
\end{figure}

\textbf{Results:} Each reward model is applied to our held-out test trajectories. We compute pairwise preferences, and compare to ground-truth preferences to obtain a test accuracy. Reward model performances are shown in Figure \ref{fig:RMs}.

We find that reward model accuracy increases approximately logarithmically with comparisons, although the random projection reward models have high variance. Additionally, the reward models utilizing the agent model perform better than the reward models using random projections across the full range of comparison sizes, confirming the relevance of the imitation learning representations. Importantly, this transfer makes it possible to train a reward model from visual input to achieve over 90\% preference accuracy with only $\sim100$ comparisons. While this is a relatively simple task, we note that this would correspond to less than 1 hour of labeling time, demonstrating the feasibility of this approach for aligning pixel-based agents from trajectory videos. 

\textbf{In summary, we find evidence of representation transfer between imitation learning and preference modeling, that enables feedback-efficient reward modeling for aligning pixel-based agents.}

\subsection{Aligning the Agent with the Reward Model}\label{subsec:alignment}

\textbf{Procedure: } 
Finally, we align our agent with the preferences captured by our reward models. We run our fine-tuned agent in the environment as for evaluation to generate online trajectories. After each trajectory is complete, we apply our reward model to the trajectory to generate a reward. We use this reward as the return for the trajectory and update our agent policy $\pi$ using REINFORCE \citep{williams_simple_1992}. We discuss the motivation behind this choice in Appendix~\ref{app:reinforce}.

    

For our experiments we ran our agent online for around a day of real-time gameplay, corresponding to 600 parameter updates (see Appendix \ref{app:details} for details). The standard approach in LLM alignment would be to fine-tuning the entire network with an additional KL divergence term to regularize the optimized policy towards the initial fine-tuned policy \citep{touvron_llama_2023, bai_training_2022}, but for our proof of concept we simply fine-tune the last layer for efficiency. We also note that approaches such as Low-Rank Adaption (LoRA) \citep{hu_lora_2021} used for fine-tuning LLMs could also be used here for memory-efficient alignment of a large base model with an individual's preferences.

\subsubsection{Aligning Agent Towards Left Jumppad}\label{sec:align_left}
\begin{figure}[h]
  \centering
  \includegraphics[width=\columnwidth]{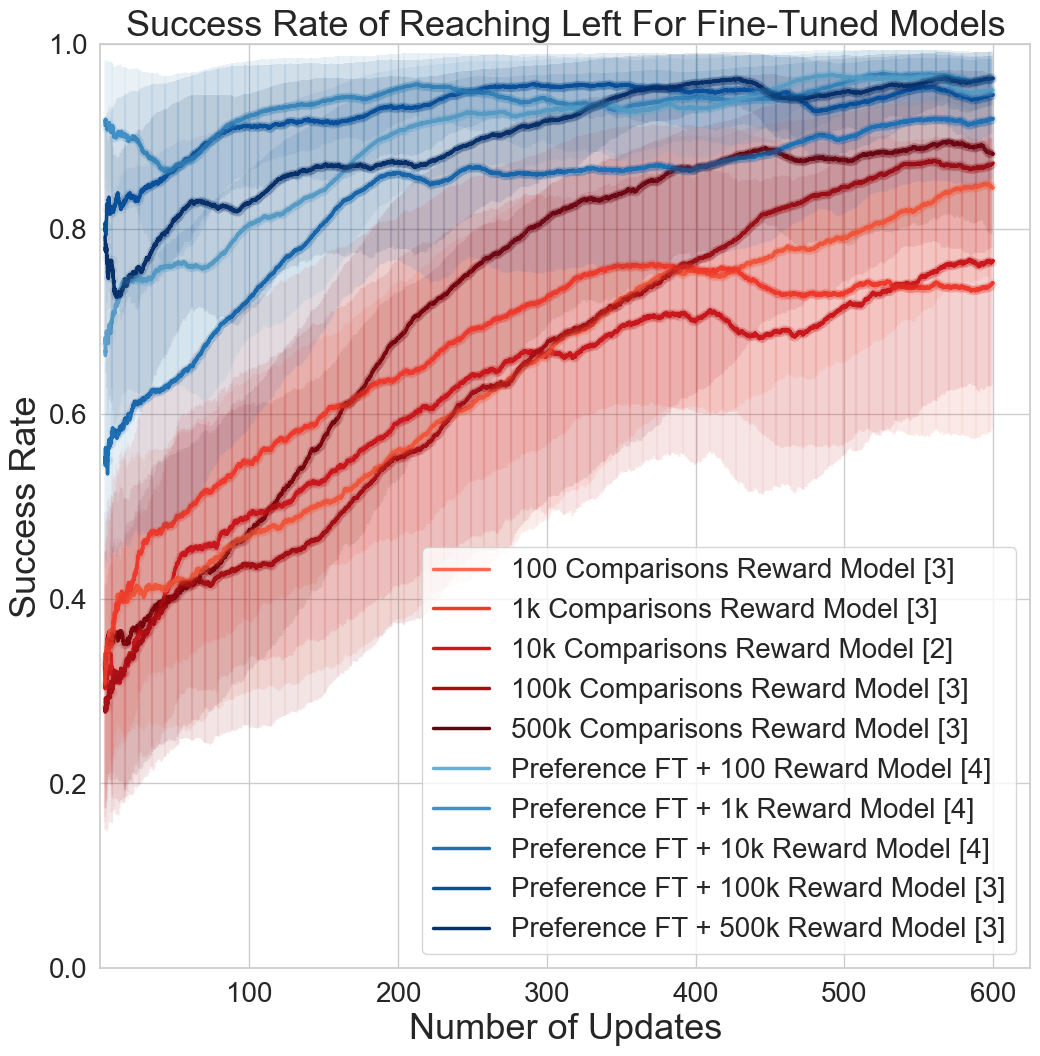}
  \caption{Left jumppad success rate during online alignment. Standard errors shaded, and number of seeds shown in legend. We see: 1) Higher accuracy reward models (darker) generally lead to better alignment, 2) Preference fine-tuning before online alignment (blue) improves performance for all reward models.}
\label{fig:Left_JP_FT_Alignment}
\end{figure}

 We begin by aligning our agent to reach the left jumppad. We plot the average success rate of reaching the left jumppad against the number of episodes in Figure~\ref{fig:Left_JP_FT_Alignment}, for reward models that have been trained on 100 up to 500k comparisons. We see that all of our reward models are sufficient to align our agent to consistently reach the left jumppad, with reward models trained on more data (with higher test performances) generally leading to better alignment. 

However, we see that the agents take most of the 600 updates to fully align, corresponding to a day of real-time training. To improve this efficiency, we again take inspiration from the LLM literature and consider the addition of a \textit{preference fine-tuning} phase in which we first further fine-tune on the preferred trajectories. 
Specifically, we apply the reward model to the training trajectories and take the top 20\% of trajectories by reward (corresponding to the preferred trajectories) and perform additional fine-tuning on these trajectories. This corresponds to a single iteration of Reinforced Self-Training (ReST) \citep{gulcehre_reinforced_2023}, recently introduced to improve the efficiency of language model alignment. We find that this improves online performance for all reward models, shown in Figure \ref{fig:Left_JP_FT_Alignment}. 

\subsubsection{Aligning Agent Towards Right Jumppad}\label{sec:align_right}
\begin{figure}[h!]
\centering
\includegraphics[width=\columnwidth]{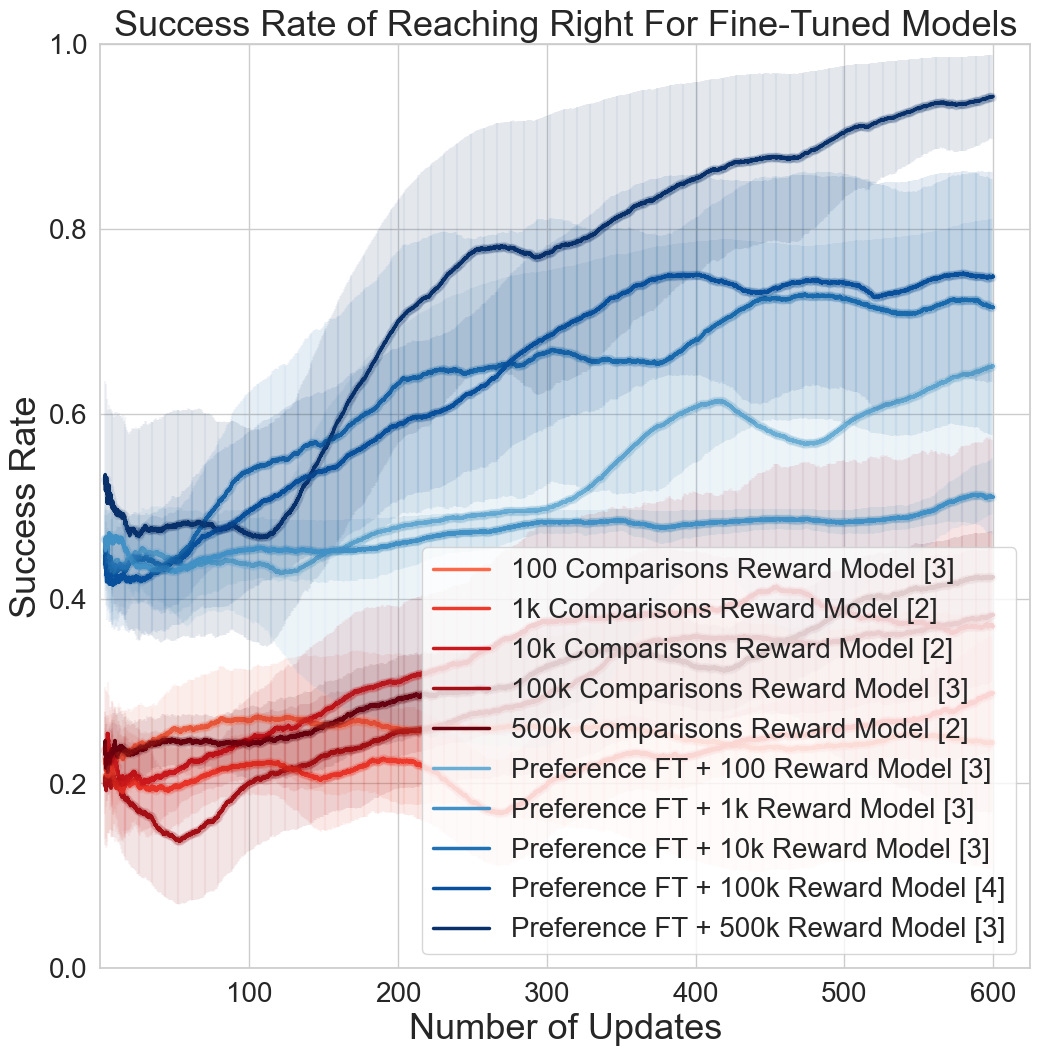}
\caption{Right jumppad success rate during online alignment. Standard errors shaded, and number of seeds shown in legend. Similar trends hold to Figure \ref{fig:Left_JP_FT_Alignment}, with the benefits of preference fine-tuning (blue) even more prominent, but overall alignment is less effective.}
\label{fig:Right_JP_FT_Alignment}
\end{figure}

We now consider aligning our agent towards the right jumppad.
Since this task is seemingly of identical difficulty, we would expect to be able to align the agent similarly.
However, we found that our agents did not align as quickly towards the right jumppad, and in the absence of preference fine-tuning only reached around a $\sim40\%$ success rate after one day of training, as shown in Figure \ref{fig:Right_JP_FT_Alignment}.

We investigate this discrepancy in Appendix~\ref{app:basealignment}, and find that it arises from an imbalance in spawn locations that makes it difficult for the agent to receive positive reward from certain spawns. However, the addition of preference fine-tuning before online alignment helps to mitigate this imbalance, as shown in Figure \ref{fig:Right_JP_FT_Alignment}, since it modifies the agent to produce more of the preferred behaviors that achieve reward before online alignment. This enabled us to completely align the agent right with the same training budget, as shown by the final jumppad distributions in Appendix \ref{app:finaljumppads}. This further demonstrates the benefits of our proposed additional stage of preference fine-tuning. However, this is just one possible strategy to improve alignment efficiency from the LLM literature. Many other approaches, such as Direct Preference Optimization (DPO) \citep{rafailov_direct_2023} could also be incorporated to alleviate the need for costly on-policy, online alignment. However, removing the need for online learning entirely is likely to be challenging, as discussed in Appendix~\ref{app:o2oshift}, and has been recently demonstrated in the context of LLMs \citep{xu_is_2024}.

\textbf{In summary, we find that agents can be aligned to behave in accordance with human preferences by leveraging equivalent approaches to those used for modern LLMs, with similar benefits and effectiveness.}

\subsection{Summary of Agent Alignment}\label{sec:alignment_summary}

\begin{figure}[H]
\vspace{-3mm}
\centering
    \includegraphics[width=\columnwidth]{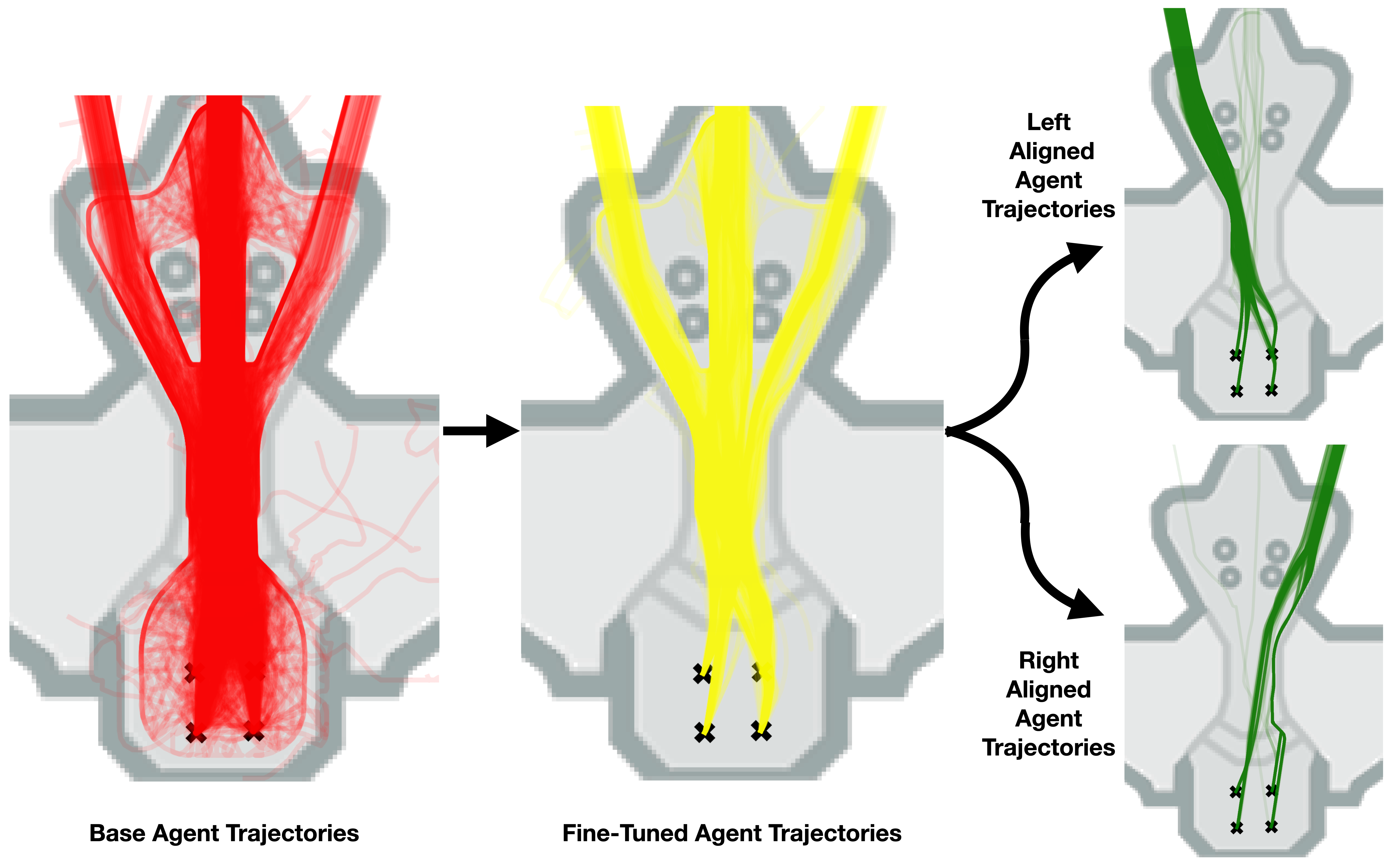}
    \caption{Heatmap of our agent's trajectories at each stage of the alignment pipeline. Each stage shows 1000 trajectories.}\label{fig:heatmap}
    \vspace{-5mm}
\end{figure}

To visualize the gradual alignment of our agent, we provide a heatmap of agent trajectories at each stage of the alignment pipeline in Figure \ref{fig:heatmap}. This demonstrates how pre-training, fine-tuning and RL(HF) can be used to reliably achieve behaviors that would not be possible with imitation learning (due to finite demonstrations) or reinforcement learning (due to lack of dense reward) alone. Videos of our agent at each stage are provided at: \url{https://adamjelley.github.io/aligning-agents-like-llms}.

\section{Alternative Views and Broader Outlook}

Our proof-of-concept provides evidence of the benefits of multi-stage training over imitation or reinforcement learning alone. However, an alternative view to our position might be that rather than training agents like LLMs, \textit{LLMs should be used as agents directly}. There has indeed been significant progress in the development of LLM agents in recent years \citep{xi_rise_2023, wang_survey_2024}, including for low-level control \citep{yang_towards_2024, brohan_rt-2_2023, ahn_as_2022, driess_palm-e_2023, tang_saytap_2023}. 

However, multi-modal LLMs still struggle with low-level decision-making in complex environments such as video games, with recent work demonstrating that these models are a long way from human performance in these environments, even with significant amounts of gameplay data provided for in-context learning \citep{waytowich_atari-gpt_2024}. In particular, \citet{paglieri_balrog_2024} recently demonstrated that LLM agents understandably struggle with experience-based aspects such as spatial reasoning, systematic exploration, long-term planning, understanding environment dynamics, and the `knowing-doing' gap. In fact, LMAct \citep{ruoss_lmact_2024} concludes ``\textit{Our work thus adds another piece to the growing literature that finds difficulties that LMs
can have with translating declarative knowledge into effective acting (know-how) in some interactive
decision-making tasks.}"

We argue that for tasks such as playing a 3D AAA video game, requiring fine-grained control of a gamepad controller at 10Hz, this problem will not simply go away with scale, and obtaining competent embodied agents will instead require training directly on interaction data. Just as human proficiency in video games and physical tasks depends more on experience than intellectual capabilities, our position is that LLM agents will never be as capable as agents trained directly on interaction data, since they lack the common-sense behavioral priors necessary for low-level control.

In our proof-of-concept, we considered training an agent on general gameplay data before fine-tuning on a specific task. However, more generally, we could envision training a gaming foundation agent across a range of games (using our unified pixel observation space and gamepad controller action space), which could then be fine-tuned and aligned for a particular game. Similarly, while we primarily consider video games, we believe the modern LLM training paradigm (described more generally in Appendix \ref{app:procedure}) could also be applied to training decision-making agents in other domains such as robotics. This could enable more general and robust agents, leveraging the impressive progress in LLMs while avoiding the fundamental limitations of LLM agents trained without real-world interaction data.

Finally, as LLMs inevitably continue to evolve, the developments can continue to be transferred to agents. For example, recent progress on improving LLM reasoning via process \citep{lightman_lets_2023} or outcome rewards \citep{lambert_tulu_2024, deepseek-ai_deepseek-r1_2025}, or inference-time scaling \citep{openai_openai_2024}, can be incorporated into this framework simply as conventional RL with environment rewards and search. But by making this stage the `cherry-on-the-cake' \citep{lecun_nips_2016} rather than the sole training procedure, we can avoid many of the difficulties traditionally associated with RL, to obtain more general, robust and useful agents.

\section{Conclusions}

In this paper, we have taken the position that general agents should be trained like large language models. We have provided evidence for our position by following the LLM paradigm to train an agent from pixels to perform a desired behavior on a 3D AAA game as a proof-of-concept. This behavior would be difficult to achieve reliably with imitation learning or RL alone, demonstrating the value of each stage of the standard LLM training pipeline in the context of agents. We believe there is potential for many other recent developments in LLMs to transfer to decision-making agents, such as optimal dataset compositions, scaling laws, architectural and context improvements, post-training alignment procedures, and inference-time scaling. This will likely open up many exciting avenues of research in the coming years that have the potential to address many long-standing limitations and open-problems for decision-making agents. 
We hope our paper provides a framework for such research, and encourages communication and collaboration between the LLM and decision-making agent communities, to enable shared insights and provide a path towards more general and reliable agents for real-world applications.



\section*{Acknowledgements}
This project was done during an internship with the Game Intelligence Team at Microsoft Research Cambridge. We are grateful to Ninja Theory and Microsoft for their support with the project. Thank you to Yuhan Cao and Dave Bignell for implementation and infrastructure support, Sam Devlin for advice and guidance, and particularly Tabish Rashid for supervising the internship and seeing the project through to completion. In addition to the authors, thank you to the whole team, including Anssi Kanervisto, Gunshi Gupta, Katja Hofmann, Lukas Schäfer, Raluca Georgescu, Sergio Valcarcel Macua, Shanzheng Tan, Tarun Gupta, Tim Pearce and Eloi Alonso for a great internship. Thank you also to Amos Storkey, Tom Lee, Trevor McInroe and Rich Turner for insightful discussions. Outside of the internship, Adam Jelley is supported by Microsoft Research and EPSRC through Microsoft’s PhD Scholarship Programme.



\section*{Impact Statement}

This paper presents work whose goal is to advance the field of machine learning, in particular decision-making agents. While these agents could have many positive impacts, ensuring the safety of such agents is crucial. Aligning agents with human preferences is a prominent approach to obtaining helpful and harmless agents, but this procedure has known open problems and limitations \citep{casper_open_2023}. Video games therefore provide an important test bed for such research, and we hope work such as ours can mitigate risks and provide insights in a safe environment that may generalize to other more
real-world applications such as robotics.




\nocite{langley00}

\bibliography{references}

@misc{paglieri_balrog_2024,
	title = {{BALROG}: {Benchmarking} {Agentic} {LLM} and {VLM} {Reasoning} {On} {Games}},
	shorttitle = {{BALROG}},
	url = {http://arxiv.org/abs/2411.13543},
	doi = {10.48550/arXiv.2411.13543},
	urldate = {2025-01-14},
	publisher = {arXiv},
	author = {Paglieri, Davide and Cupiał, Bartłomiej and Coward, Samuel and Piterbarg, Ulyana and Wolczyk, Maciej and Khan, Akbir and Pignatelli, Eduardo and Kucinski, Lukasz and Pinto, Lerrel and Fergus, Rob and Foerster, Jakob Nicolaus and Parker-Holder, Jack and Rocktäschel, Tim},
	month = nov,
	year = {2024},
	note = {arXiv:2411.13543 [cs]},
}

@misc{embodimentcollaboration2024openxembodimentroboticlearning,
	title = {Open x-embodiment: {Robotic} learning datasets and {RT}-x models},
	url = {https://arxiv.org/abs/2310.08864},
	author = {Embodiment-Collaboration and O'Neill, Abby and Rehman, Abdul and Gupta, Abhinav and Maddukuri, Abhiram and Gupta, Abhishek and Padalkar, Abhishek and Lee, Abraham and Pooley, Acorn and Gupta, Agrim and Mandlekar, Ajay and Jain, Ajinkya and Tung, Albert and Bewley, Alex and Herzog, Alex and Irpan, Alex and Khazatsky, Alexander and Rai, Anant and Gupta, Anchit and Wang, Andrew and Kolobov, Andrey and Singh, Anikait and Garg, Animesh and Kembhavi, Aniruddha and Xie, Annie and Brohan, Anthony and Raffin, Antonin and Sharma, Archit and Yavary, Arefeh and Jain, Arhan and Balakrishna, Ashwin and Wahid, Ayzaan and {...} and Zhu, Yuke and Zhang, Yunchu and Jiang, Yunfan and Li, Yunshuang and Li, Yunzhu and Iwasawa, Yusuke and Matsuo, Yutaka and Ma, Zehan and Xu, Zhuo and Cui, Zichen Jeff and Zhang, Zichen and Fu, Zipeng and Lin, Zipeng},
	year = {2024},
	note = {arXiv: 2310.08864 [cs.RO]},
}

@misc{xu_is_2024,
	title = {Is {DPO} {Superior} to {PPO} for {LLM} {Alignment}? {A} {Comprehensive} {Study}},
	shorttitle = {Is {DPO} {Superior} to {PPO} for {LLM} {Alignment}?},
	url = {http://arxiv.org/abs/2404.10719},
	doi = {10.48550/arXiv.2404.10719},
	urldate = {2025-01-29},
	publisher = {arXiv},
	author = {Xu, Shusheng and Fu, Wei and Gao, Jiaxuan and Ye, Wenjie and Liu, Weilin and Mei, Zhiyu and Wang, Guangju and Yu, Chao and Wu, Yi},
	month = apr,
	year = {2024},
	note = {arXiv:2404.10719 [cs]
version: 1},
}

@misc{lewis_retrieval-augmented_2021,
	title = {Retrieval-{Augmented} {Generation} for {Knowledge}-{Intensive} {NLP} {Tasks}},
	url = {http://arxiv.org/abs/2005.11401},
	doi = {10.48550/arXiv.2005.11401},
	urldate = {2025-01-27},
	publisher = {arXiv},
	author = {Lewis, Patrick and Perez, Ethan and Piktus, Aleksandra and Petroni, Fabio and Karpukhin, Vladimir and Goyal, Naman and Küttler, Heinrich and Lewis, Mike and Yih, Wen-tau and Rocktäschel, Tim and Riedel, Sebastian and Kiela, Douwe},
	month = apr,
	year = {2021},
	note = {arXiv:2005.11401 [cs]},
}

@misc{li_llms-as-judges_2024,
	title = {{LLMs}-as-{Judges}: {A} {Comprehensive} {Survey} on {LLM}-based {Evaluation} {Methods}},
	shorttitle = {{LLMs}-as-{Judges}},
	url = {http://arxiv.org/abs/2412.05579},
	doi = {10.48550/arXiv.2412.05579},
	urldate = {2025-01-24},
	publisher = {arXiv},
	author = {Li, Haitao and Dong, Qian and Chen, Junjie and Su, Huixue and Zhou, Yujia and Ai, Qingyao and Ye, Ziyi and Liu, Yiqun},
	month = dec,
	year = {2024},
	note = {arXiv:2412.05579 [cs]},
}

@misc{deepseek-ai_deepseek-r1_2025,
	title = {{DeepSeek}-{R1}: {Incentivizing} {Reasoning} {Capability} in {LLMs} via {Reinforcement} {Learning}},
	shorttitle = {{DeepSeek}-{R1}},
	url = {http://arxiv.org/abs/2501.12948},
	doi = {10.48550/arXiv.2501.12948},
	urldate = {2025-01-24},
	publisher = {arXiv},
	author = {DeepSeek-AI and Guo, Daya and Yang, Dejian and Zhang, Haowei and Song, Junxiao and Zhang, Ruoyu and Xu, Runxin and Zhu, Qihao and Ma, Shirong and Wang, Peiyi and Bi, Xiao and Zhang, Xiaokang and Yu, Xingkai and Wu, Yu and Wu, Z. F. and Gou, Zhibin and Shao, Zhihong and Li, Zhuoshu and Gao, Ziyi and Liu, Aixin and Xue, Bing and Wang, Bingxuan and Wu, Bochao and Feng, Bei and Lu, Chengda and Zhao, Chenggang and Deng, Chengqi and Zhang, Chenyu and Ruan, Chong and Dai, Damai and Chen, Deli and Ji, Dongjie and Li, Erhang and Lin, Fangyun and Dai, Fucong and Luo, Fuli and Hao, Guangbo and Chen, Guanting and Li, Guowei and Zhang, H. and Bao, Han and Xu, Hanwei and Wang, Haocheng and Ding, Honghui and Xin, Huajian and Gao, Huazuo and Qu, Hui and Li, Hui and Guo, Jianzhong and Li, Jiashi and Wang, Jiawei and Chen, Jingchang and Yuan, Jingyang and Qiu, Junjie and Li, Junlong and Cai, J. L. and Ni, Jiaqi and Liang, Jian and Chen, Jin and Dong, Kai and Hu, Kai and Gao, Kaige and Guan, Kang and Huang, Kexin and Yu, Kuai and Wang, Lean and Zhang, Lecong and Zhao, Liang and Wang, Litong and Zhang, Liyue and Xu, Lei and Xia, Leyi and Zhang, Mingchuan and Zhang, Minghua and Tang, Minghui and Li, Meng and Wang, Miaojun and Li, Mingming and Tian, Ning and Huang, Panpan and Zhang, Peng and Wang, Qiancheng and Chen, Qinyu and Du, Qiushi and Ge, Ruiqi and Zhang, Ruisong and Pan, Ruizhe and Wang, Runji and Chen, R. J. and Jin, R. L. and Chen, Ruyi and Lu, Shanghao and Zhou, Shangyan and Chen, Shanhuang and Ye, Shengfeng and Wang, Shiyu and Yu, Shuiping and Zhou, Shunfeng and Pan, Shuting and Li, S. S. and Zhou, Shuang and Wu, Shaoqing and Ye, Shengfeng and Yun, Tao and Pei, Tian and Sun, Tianyu and Wang, T. and Zeng, Wangding and Zhao, Wanjia and Liu, Wen and Liang, Wenfeng and Gao, Wenjun and Yu, Wenqin and Zhang, Wentao and Xiao, W. L. and An, Wei and Liu, Xiaodong and Wang, Xiaohan and Chen, Xiaokang and Nie, Xiaotao and Cheng, Xin and Liu, Xin and Xie, Xin and Liu, Xingchao and Yang, Xinyu and Li, Xinyuan and Su, Xuecheng and Lin, Xuheng and Li, X. Q. and Jin, Xiangyue and Shen, Xiaojin and Chen, Xiaosha and Sun, Xiaowen and Wang, Xiaoxiang and Song, Xinnan and Zhou, Xinyi and Wang, Xianzu and Shan, Xinxia and Li, Y. K. and Wang, Y. Q. and Wei, Y. X. and Zhang, Yang and Xu, Yanhong and Li, Yao and Zhao, Yao and Sun, Yaofeng and Wang, Yaohui and Yu, Yi and Zhang, Yichao and Shi, Yifan and Xiong, Yiliang and He, Ying and Piao, Yishi and Wang, Yisong and Tan, Yixuan and Ma, Yiyang and Liu, Yiyuan and Guo, Yongqiang and Ou, Yuan and Wang, Yuduan and Gong, Yue and Zou, Yuheng and He, Yujia and Xiong, Yunfan and Luo, Yuxiang and You, Yuxiang and Liu, Yuxuan and Zhou, Yuyang and Zhu, Y. X. and Xu, Yanhong and Huang, Yanping and Li, Yaohui and Zheng, Yi and Zhu, Yuchen and Ma, Yunxian and Tang, Ying and Zha, Yukun and Yan, Yuting and Ren, Z. Z. and Ren, Zehui and Sha, Zhangli and Fu, Zhe and Xu, Zhean and Xie, Zhenda and Zhang, Zhengyan and Hao, Zhewen and Ma, Zhicheng and Yan, Zhigang and Wu, Zhiyu and Gu, Zihui and Zhu, Zijia and Liu, Zijun and Li, Zilin and Xie, Ziwei and Song, Ziyang and Pan, Zizheng and Huang, Zhen and Xu, Zhipeng and Zhang, Zhongyu and Zhang, Zhen},
	month = jan,
	year = {2025},
	note = {arXiv:2501.12948 [cs]},
}

@misc{hoffmann_training_2022,
	title = {Training {Compute}-{Optimal} {Large} {Language} {Models}},
	url = {http://arxiv.org/abs/2203.15556},
	doi = {10.48550/arXiv.2203.15556},
	urldate = {2025-01-24},
	publisher = {arXiv},
	author = {Hoffmann, Jordan and Borgeaud, Sebastian and Mensch, Arthur and Buchatskaya, Elena and Cai, Trevor and Rutherford, Eliza and Casas, Diego de Las and Hendricks, Lisa Anne and Welbl, Johannes and Clark, Aidan and Hennigan, Tom and Noland, Eric and Millican, Katie and Driessche, George van den and Damoc, Bogdan and Guy, Aurelia and Osindero, Simon and Simonyan, Karen and Elsen, Erich and Rae, Jack W. and Vinyals, Oriol and Sifre, Laurent},
	month = mar,
	year = {2022},
	note = {arXiv:2203.15556 [cs]},
}

@inproceedings{ross_efficient_2010,
	title = {Efficient {Reductions} for {Imitation} {Learning}},
	url = {https://proceedings.mlr.press/v9/ross10a.html},
	language = {en},
	urldate = {2025-01-23},
	booktitle = {Proceedings of the {Thirteenth} {International} {Conference} on {Artificial} {Intelligence} and {Statistics}},
	publisher = {JMLR Workshop and Conference Proceedings},
	author = {Ross, Stephane and Bagnell, Drew},
	month = mar,
	year = {2010},
	note = {ISSN: 1938-7228},
	pages = {661--668},
}

@misc{lecun_nips_2016,
	title = {{NIPS} {Invited} {Talk} {Predictive} {Learning}},
	url = {https://nips.cc/virtual/2016/invited-talk/6197},
	urldate = {2025-01-20},
	journal = {NIPS Invited Talk},
	author = {LeCun, Yann},
	year = {2016},
}

@misc{lambert_tulu_2024,
	title = {T{ÜLU} 3: {Pushing} {Frontiers} in {Open} {Language} {Model} {Post}-{Training}},
	shorttitle = {T{ÜLU} 3},
	url = {http://arxiv.org/abs/2411.15124},
	doi = {10.48550/arXiv.2411.15124},
	urldate = {2025-01-20},
	publisher = {arXiv},
	author = {Lambert, Nathan and Morrison, Jacob and Pyatkin, Valentina and Huang, Shengyi and Ivison, Hamish and Brahman, Faeze and Miranda, Lester James V. and Liu, Alisa and Dziri, Nouha and Lyu, Shane and Gu, Yuling and Malik, Saumya and Graf, Victoria and Hwang, Jena D. and Yang, Jiangjiang and Bras, Ronan Le and Tafjord, Oyvind and Wilhelm, Chris and Soldaini, Luca and Smith, Noah A. and Wang, Yizhong and Dasigi, Pradeep and Hajishirzi, Hannaneh},
	month = nov,
	year = {2024},
	note = {arXiv:2411.15124 [cs]
version: 1},
}

@misc{lightman_lets_2023,
	title = {Let's {Verify} {Step} by {Step}},
	url = {http://arxiv.org/abs/2305.20050},
	doi = {10.48550/arXiv.2305.20050},
	urldate = {2025-01-20},
	publisher = {arXiv},
	author = {Lightman, Hunter and Kosaraju, Vineet and Burda, Yura and Edwards, Harri and Baker, Bowen and Lee, Teddy and Leike, Jan and Schulman, John and Sutskever, Ilya and Cobbe, Karl},
	month = may,
	year = {2023},
	note = {arXiv:2305.20050 [cs]},
}

@misc{openai_openai_2024,
	title = {{OpenAI} o1 {System} {Card}},
	url = {http://arxiv.org/abs/2412.16720},
	doi = {10.48550/arXiv.2412.16720},
	urldate = {2025-01-20},
	publisher = {arXiv},
	author = {OpenAI and Jaech, Aaron and Kalai, Adam and Lerer, Adam and Richardson, Adam and El-Kishky, Ahmed and Low, Aiden and Helyar, Alec and Madry, Aleksander and Beutel, Alex and Carney, Alex and Iftimie, Alex and Karpenko, Alex and Passos, Alex Tachard and Neitz, Alexander and Prokofiev, Alexander and Wei, Alexander and Tam, Allison and Bennett, Ally and Kumar, Ananya and Saraiva, Andre and Vallone, Andrea and Duberstein, Andrew and Kondrich, Andrew and Mishchenko, Andrey and Applebaum, Andy and Jiang, Angela and Nair, Ashvin and Zoph, Barret and Ghorbani, Behrooz and Rossen, Ben and Sokolowsky, Benjamin and Barak, Boaz and McGrew, Bob and Minaiev, Borys and Hao, Botao and Baker, Bowen and Houghton, Brandon and McKinzie, Brandon and Eastman, Brydon and Lugaresi, Camillo and Bassin, Cary and Hudson, Cary and Li, Chak Ming and Bourcy, Charles de and Voss, Chelsea and Shen, Chen and Zhang, Chong and Koch, Chris and Orsinger, Chris and Hesse, Christopher and Fischer, Claudia and Chan, Clive and Roberts, Dan and Kappler, Daniel and Levy, Daniel and Selsam, Daniel and Dohan, David and Farhi, David and Mely, David and Robinson, David and Tsipras, Dimitris and Li, Doug and Oprica, Dragos and Freeman, Eben and Zhang, Eddie and Wong, Edmund and Proehl, Elizabeth and Cheung, Enoch and Mitchell, Eric and Wallace, Eric and Ritter, Erik and Mays, Evan and Wang, Fan and Such, Felipe Petroski and Raso, Filippo and Leoni, Florencia and Tsimpourlas, Foivos and Song, Francis and Lohmann, Fred von and Sulit, Freddie and Salmon, Geoff and Parascandolo, Giambattista and Chabot, Gildas and Zhao, Grace and Brockman, Greg and Leclerc, Guillaume and Salman, Hadi and Bao, Haiming and Sheng, Hao and Andrin, Hart and Bagherinezhad, Hessam and Ren, Hongyu and Lightman, Hunter and Chung, Hyung Won and Kivlichan, Ian and O'Connell, Ian and Osband, Ian and Gilaberte, Ignasi Clavera and Akkaya, Ilge and Kostrikov, Ilya and Sutskever, Ilya and Kofman, Irina and Pachocki, Jakub and Lennon, James and Wei, Jason and Harb, Jean and Twore, Jerry and Feng, Jiacheng and Yu, Jiahui and Weng, Jiayi and Tang, Jie and Yu, Jieqi and Candela, Joaquin Quiñonero and Palermo, Joe and Parish, Joel and Heidecke, Johannes and Hallman, John and Rizzo, John and Gordon, Jonathan and Uesato, Jonathan and Ward, Jonathan and Huizinga, Joost and Wang, Julie and Chen, Kai and Xiao, Kai and Singhal, Karan and Nguyen, Karina and Cobbe, Karl and Shi, Katy and Wood, Kayla and Rimbach, Kendra and Gu-Lemberg, Keren and Liu, Kevin and Lu, Kevin and Stone, Kevin and Yu, Kevin and Ahmad, Lama and Yang, Lauren and Liu, Leo and Maksin, Leon and Ho, Leyton and Fedus, Liam and Weng, Lilian and Li, Linden and McCallum, Lindsay and Held, Lindsey and Kuhn, Lorenz and Kondraciuk, Lukas and Kaiser, Lukasz and Metz, Luke and Boyd, Madelaine and Trebacz, Maja and Joglekar, Manas and Chen, Mark and Tintor, Marko and Meyer, Mason and Jones, Matt and Kaufer, Matt and Schwarzer, Max and Shah, Meghan and Yatbaz, Mehmet and Guan, Melody Y. and Xu, Mengyuan and Yan, Mengyuan and Glaese, Mia and Chen, Mianna and Lampe, Michael and Malek, Michael and Wang, Michele and Fradin, Michelle and McClay, Mike and Pavlov, Mikhail and Wang, Miles and Wang, Mingxuan and Murati, Mira and Bavarian, Mo and Rohaninejad, Mostafa and McAleese, Nat and Chowdhury, Neil and Chowdhury, Neil and Ryder, Nick and Tezak, Nikolas and Brown, Noam and Nachum, Ofir and Boiko, Oleg and Murk, Oleg and Watkins, Olivia and Chao, Patrick and Ashbourne, Paul and Izmailov, Pavel and Zhokhov, Peter and Dias, Rachel and Arora, Rahul and Lin, Randall and Lopes, Rapha Gontijo and Gaon, Raz and Miyara, Reah and Leike, Reimar and Hwang, Renny and Garg, Rhythm and Brown, Robin and James, Roshan and Shu, Rui and Cheu, Ryan and Greene, Ryan and Jain, Saachi and Altman, Sam and Toizer, Sam and Toyer, Sam and Miserendino, Samuel and Agarwal, Sandhini and Hernandez, Santiago and Baker, Sasha and McKinney, Scott and Yan, Scottie and Zhao, Shengjia and Hu, Shengli and Santurkar, Shibani and Chaudhuri, Shraman Ray and Zhang, Shuyuan and Fu, Siyuan and Papay, Spencer and Lin, Steph and Balaji, Suchir and Sanjeev, Suvansh and Sidor, Szymon and Broda, Tal and Clark, Aidan and Wang, Tao and Gordon, Taylor and Sanders, Ted and Patwardhan, Tejal and Sottiaux, Thibault and Degry, Thomas and Dimson, Thomas and Zheng, Tianhao and Garipov, Timur and Stasi, Tom and Bansal, Trapit and Creech, Trevor and Peterson, Troy and Eloundou, Tyna and Qi, Valerie and Kosaraju, Vineet and Monaco, Vinnie and Pong, Vitchyr and Fomenko, Vlad and Zheng, Weiyi and Zhou, Wenda and McCabe, Wes and Zaremba, Wojciech and Dubois, Yann and Lu, Yinghai and Chen, Yining and Cha, Young and Bai, Yu and He, Yuchen and Zhang, Yuchen and Wang, Yunyun and Shao, Zheng and Li, Zhuohan},
	month = dec,
	year = {2024},
	note = {arXiv:2412.16720 [cs]},
}

@misc{xi_rise_2023,
	title = {The {Rise} and {Potential} of {Large} {Language} {Model} {Based} {Agents}: {A} {Survey}},
	shorttitle = {The {Rise} and {Potential} of {Large} {Language} {Model} {Based} {Agents}},
	url = {http://arxiv.org/abs/2309.07864},
	doi = {10.48550/arXiv.2309.07864},
	urldate = {2025-01-20},
	publisher = {arXiv},
	author = {Xi, Zhiheng and Chen, Wenxiang and Guo, Xin and He, Wei and Ding, Yiwen and Hong, Boyang and Zhang, Ming and Wang, Junzhe and Jin, Senjie and Zhou, Enyu and Zheng, Rui and Fan, Xiaoran and Wang, Xiao and Xiong, Limao and Zhou, Yuhao and Wang, Weiran and Jiang, Changhao and Zou, Yicheng and Liu, Xiangyang and Yin, Zhangyue and Dou, Shihan and Weng, Rongxiang and Cheng, Wensen and Zhang, Qi and Qin, Wenjuan and Zheng, Yongyan and Qiu, Xipeng and Huang, Xuanjing and Gui, Tao},
	month = sep,
	year = {2023},
	note = {arXiv:2309.07864 [cs]},
}

@article{wang_survey_2024,
	title = {A {Survey} on {Large} {Language} {Model} based {Autonomous} {Agents}},
	volume = {18},
	issn = {2095-2228, 2095-2236},
	url = {http://arxiv.org/abs/2308.11432},
	doi = {10.1007/s11704-024-40231-1},
	number = {6},
	urldate = {2025-01-20},
	journal = {Frontiers of Computer Science},
	author = {Wang, Lei and Ma, Chen and Feng, Xueyang and Zhang, Zeyu and Yang, Hao and Zhang, Jingsen and Chen, Zhiyuan and Tang, Jiakai and Chen, Xu and Lin, Yankai and Zhao, Wayne Xin and Wei, Zhewei and Wen, Ji-Rong},
	month = dec,
	year = {2024},
	note = {arXiv:2308.11432 [cs]},
	pages = {186345},
}

@misc{zheng_judging_2023,
	title = {Judging {LLM}-as-a-{Judge} with {MT}-{Bench} and {Chatbot} {Arena}},
	url = {http://arxiv.org/abs/2306.05685},
	doi = {10.48550/arXiv.2306.05685},
	urldate = {2025-01-17},
	publisher = {arXiv},
	author = {Zheng, Lianmin and Chiang, Wei-Lin and Sheng, Ying and Zhuang, Siyuan and Wu, Zhanghao and Zhuang, Yonghao and Lin, Zi and Li, Zhuohan and Li, Dacheng and Xing, Eric P. and Zhang, Hao and Gonzalez, Joseph E. and Stoica, Ion},
	month = dec,
	year = {2023},
	note = {arXiv:2306.05685 [cs]},
}

@misc{geminiteam2024geminifamilyhighlycapable,
	title = {Gemini: a family of highly capable multimodal models},
	url = {https://arxiv.org/abs/2312.11805},
	author = {Gemini-Team, Google and Anil, Rohan and Borgeaud, Sebastian and Alayrac, Jean-Baptiste and Yu, Jiahui and Soricut, Radu and Schalkwyk, Johan and Dai, Andrew M. and Hauth, Anja and Millican, Katie and Silver, David and Johnson, Melvin and Antonoglou, Ioannis and Schrittwieser, Julian and Glaese, Amelia and Chen, Jilin and Pitler, Emily and Lillicrap, Timothy and Lazaridou, Angeliki and Firat, Orhan and Molloy, James and Isard, Michael and Barham, Paul R. and Hennigan, Tom and Lee, Benjamin and ... and Dean, Jeffrey and Vinyals, Oriol},
	year = {2024},
	note = {arXiv: 2312.11805 [cs.CL]},
}

@misc{esser_scaling_2024,
	title = {Scaling {Rectified} {Flow} {Transformers} for {High}-{Resolution} {Image} {Synthesis}},
	url = {http://arxiv.org/abs/2403.03206},
	doi = {10.48550/arXiv.2403.03206},
	urldate = {2025-01-17},
	publisher = {arXiv},
	author = {Esser, Patrick and Kulal, Sumith and Blattmann, Andreas and Entezari, Rahim and Müller, Jonas and Saini, Harry and Levi, Yam and Lorenz, Dominik and Sauer, Axel and Boesel, Frederic and Podell, Dustin and Dockhorn, Tim and English, Zion and Lacey, Kyle and Goodwin, Alex and Marek, Yannik and Rombach, Robin},
	month = mar,
	year = {2024},
	note = {arXiv:2403.03206 [cs]},
}

@misc{pearce_scaling_2024,
	title = {Scaling {Laws} for {Pre}-training {Agents} and {World} {Models}},
	url = {http://arxiv.org/abs/2411.04434},
	doi = {10.48550/arXiv.2411.04434},
	urldate = {2025-01-17},
	publisher = {arXiv},
	author = {Pearce, Tim and Rashid, Tabish and Bignell, Dave and Georgescu, Raluca and Devlin, Sam and Hofmann, Katja},
	month = dec,
	year = {2024},
	note = {arXiv:2411.04434 [cs]},
}

@misc{driess_palm-e_2023,
	title = {{PaLM}-{E}: {An} {Embodied} {Multimodal} {Language} {Model}},
	shorttitle = {{PaLM}-{E}},
	url = {http://arxiv.org/abs/2303.03378},
	doi = {10.48550/arXiv.2303.03378},
	urldate = {2025-01-14},
	publisher = {arXiv},
	author = {Driess, Danny and Xia, Fei and Sajjadi, Mehdi S. M. and Lynch, Corey and Chowdhery, Aakanksha and Ichter, Brian and Wahid, Ayzaan and Tompson, Jonathan and Vuong, Quan and Yu, Tianhe and Huang, Wenlong and Chebotar, Yevgen and Sermanet, Pierre and Duckworth, Daniel and Levine, Sergey and Vanhoucke, Vincent and Hausman, Karol and Toussaint, Marc and Greff, Klaus and Zeng, Andy and Mordatch, Igor and Florence, Pete},
	month = mar,
	year = {2023},
	note = {arXiv:2303.03378 [cs]},
}

@misc{ahn_as_2022,
	title = {Do {As} {I} {Can}, {Not} {As} {I} {Say}: {Grounding} {Language} in {Robotic} {Affordances}},
	shorttitle = {Do {As} {I} {Can}, {Not} {As} {I} {Say}},
	url = {http://arxiv.org/abs/2204.01691},
	doi = {10.48550/arXiv.2204.01691},
	urldate = {2025-01-14},
	publisher = {arXiv},
	author = {Ahn, Michael and Brohan, Anthony and Brown, Noah and Chebotar, Yevgen and Cortes, Omar and David, Byron and Finn, Chelsea and Fu, Chuyuan and Gopalakrishnan, Keerthana and Hausman, Karol and Herzog, Alex and Ho, Daniel and Hsu, Jasmine and Ibarz, Julian and Ichter, Brian and Irpan, Alex and Jang, Eric and Ruano, Rosario Jauregui and Jeffrey, Kyle and Jesmonth, Sally and Joshi, Nikhil J. and Julian, Ryan and Kalashnikov, Dmitry and Kuang, Yuheng and Lee, Kuang-Huei and Levine, Sergey and Lu, Yao and Luu, Linda and Parada, Carolina and Pastor, Peter and Quiambao, Jornell and Rao, Kanishka and Rettinghouse, Jarek and Reyes, Diego and Sermanet, Pierre and Sievers, Nicolas and Tan, Clayton and Toshev, Alexander and Vanhoucke, Vincent and Xia, Fei and Xiao, Ted and Xu, Peng and Xu, Sichun and Yan, Mengyuan and Zeng, Andy},
	month = aug,
	year = {2022},
	note = {arXiv:2204.01691 [cs]},
}

@misc{yang_towards_2024,
	title = {Towards {Unified} {Alignment} {Between} {Agents}, {Humans}, and {Environment}},
	url = {http://arxiv.org/abs/2402.07744},
	doi = {10.48550/arXiv.2402.07744},
	urldate = {2025-01-14},
	publisher = {arXiv},
	author = {Yang, Zonghan and Liu, An and Liu, Zijun and Liu, Kaiming and Xiong, Fangzhou and Wang, Yile and Yang, Zeyuan and Hu, Qingyuan and Chen, Xinrui and Zhang, Zhenhe and Luo, Fuwen and Guo, Zhicheng and Li, Peng and Liu, Yang},
	month = feb,
	year = {2024},
	note = {arXiv:2402.07744 [cs]},
}

@misc{waytowich_atari-gpt_2024,
	title = {Atari-{GPT}: {Investigating} the {Capabilities} of {Multimodal} {Large} {Language} {Models} as {Low}-{Level} {Policies} for {Atari} {Games}},
	shorttitle = {Atari-{GPT}},
	url = {http://arxiv.org/abs/2408.15950},
	doi = {10.48550/arXiv.2408.15950},
	urldate = {2025-01-14},
	publisher = {arXiv},
	author = {Waytowich, Nicholas R. and White, Devin and Sunbeam, M. D. and Goecks, Vinicius G.},
	month = aug,
	year = {2024},
	note = {arXiv:2408.15950 [cs]
version: 1},
}

@misc{ruoss_lmact_2024,
	title = {{LMAct}: {A} {Benchmark} for {In}-{Context} {Imitation} {Learning} with {Long} {Multimodal} {Demonstrations}},
	shorttitle = {{LMAct}},
	url = {http://arxiv.org/abs/2412.01441},
	doi = {10.48550/arXiv.2412.01441},
	urldate = {2025-01-14},
	publisher = {arXiv},
	author = {Ruoss, Anian and Pardo, Fabio and Chan, Harris and Li, Bonnie and Mnih, Volodymyr and Genewein, Tim},
	month = dec,
	year = {2024},
	note = {arXiv:2412.01441 [cs]},
}

@misc{tang_saytap_2023,
	title = {{SayTap}: {Language} to {Quadrupedal} {Locomotion}},
	shorttitle = {{SayTap}},
	url = {http://arxiv.org/abs/2306.07580},
	doi = {10.48550/arXiv.2306.07580},
	urldate = {2025-01-14},
	publisher = {arXiv},
	author = {Tang, Yujin and Yu, Wenhao and Tan, Jie and Zen, Heiga and Faust, Aleksandra and Harada, Tatsuya},
	month = sep,
	year = {2023},
	note = {arXiv:2306.07580 [cs]},
}

@misc{brohan_rt-1_2022,
	title = {{RT}-1: {Robotics} {Transformer} for {Real}-{World} {Control} at {Scale}},
	shorttitle = {{RT}-1},
	url = {http://arxiv.org/abs/2212.06817},
	doi = {10.48550/arXiv.2212.06817},
	urldate = {2023-09-08},
	publisher = {arXiv},
	author = {Brohan, Anthony and Brown, Noah and Carbajal, Justice and Chebotar, Yevgen and Dabis, Joseph and Finn, Chelsea and Gopalakrishnan, Keerthana and Hausman, Karol and Herzog, Alex and Hsu, Jasmine and Ibarz, Julian and Ichter, Brian and Irpan, Alex and Jackson, Tomas and Jesmonth, Sally and Joshi, Nikhil J. and Julian, Ryan and Kalashnikov, Dmitry and Kuang, Yuheng and Leal, Isabel and Lee, Kuang-Huei and Levine, Sergey and Lu, Yao and Malla, Utsav and Manjunath, Deeksha and Mordatch, Igor and Nachum, Ofir and Parada, Carolina and Peralta, Jodilyn and Perez, Emily and Pertsch, Karl and Quiambao, Jornell and Rao, Kanishka and Ryoo, Michael and Salazar, Grecia and Sanketi, Pannag and Sayed, Kevin and Singh, Jaspiar and Sontakke, Sumedh and Stone, Austin and Tan, Clayton and Tran, Huong and Vanhoucke, Vincent and Vega, Steve and Vuong, Quan and Xia, Fei and Xiao, Ted and Xu, Peng and Xu, Sichun and Yu, Tianhe and Zitkovich, Brianna},
	month = dec,
	year = {2022},
	note = {arXiv:2212.06817 [cs]},
}

@misc{sima-team_scaling_2024,
	title = {Scaling {Instructable} {Agents} {Across} {Many} {Simulated} {Worlds}},
	url = {http://arxiv.org/abs/2404.10179},
	doi = {10.48550/arXiv.2404.10179},
	urldate = {2024-05-21},
	publisher = {arXiv},
	author = {SIMA-Team and Raad, Maria Abi and Ahuja, Arun and Barros, Catarina and Besse, Frederic and Bolt, Andrew and Bolton, Adrian and Brownfield, Bethanie and Buttimore, Gavin and Cant, Max and Chakera, Sarah and Chan, Stephanie C. Y. and Clune, Jeff and Collister, Adrian and Copeman, Vikki and Cullum, Alex and Dasgupta, Ishita and de Cesare, Dario and Di Trapani, Julia and Donchev, Yani and Dunleavy, Emma and Engelcke, Martin and Faulkner, Ryan and Garcia, Frankie and Gbadamosi, Charles and Gong, Zhitao and Gonzales, Lucy and Gupta, Kshitij and Gregor, Karol and Hallingstad, Arne Olav and Harley, Tim and Haves, Sam and Hill, Felix and Hirst, Ed and Hudson, Drew A. and Hudson, Jony and Hughes-Fitt, Steph and Rezende, Danilo J. and Jasarevic, Mimi and Kampis, Laura and Ke, Rosemary and Keck, Thomas and Kim, Junkyung and Knagg, Oscar and Kopparapu, Kavya and Lampinen, Andrew and Legg, Shane and Lerchner, Alexander and Limont, Marjorie and Liu, Yulan and Loks-Thompson, Maria and Marino, Joseph and Cussons, Kathryn Martin and Matthey, Loic and Mcloughlin, Siobhan and Mendolicchio, Piermaria and Merzic, Hamza and Mitenkova, Anna and Moufarek, Alexandre and Oliveira, Valeria and Oliveira, Yanko and Openshaw, Hannah and Pan, Renke and Pappu, Aneesh and Platonov, Alex and Purkiss, Ollie and Reichert, David and Reid, John and Richemond, Pierre Harvey and Roberts, Tyson and Ruscoe, Giles and Elias, Jaume Sanchez and Sandars, Tasha and Sawyer, Daniel P. and Scholtes, Tim and Simmons, Guy and Slater, Daniel and Soyer, Hubert and Strathmann, Heiko and Stys, Peter and Tam, Allison C. and Teplyashin, Denis and Terzi, Tayfun and Vercelli, Davide and Vujatovic, Bojan and Wainwright, Marcus and Wang, Jane X. and Wang, Zhengdong and Wierstra, Daan and Williams, Duncan and Wong, Nathaniel and York, Sarah and Young, Nick},
	month = apr,
	year = {2024},
	note = {arXiv:2404.10179 [cs]},
}

@misc{bousmalis_robocat_2023,
	title = {{RoboCat}: {A} {Self}-{Improving} {Generalist} {Agent} for {Robotic} {Manipulation}},
	shorttitle = {{RoboCat}},
	url = {http://arxiv.org/abs/2306.11706},
	doi = {10.48550/arXiv.2306.11706},
	urldate = {2024-05-22},
	publisher = {arXiv},
	author = {Bousmalis, Konstantinos and Vezzani, Giulia and Rao, Dushyant and Devin, Coline and Lee, Alex X. and Bauza, Maria and Davchev, Todor and Zhou, Yuxiang and Gupta, Agrim and Raju, Akhil and Laurens, Antoine and Fantacci, Claudio and Dalibard, Valentin and Zambelli, Martina and Martins, Murilo and Pevceviciute, Rugile and Blokzijl, Michiel and Denil, Misha and Batchelor, Nathan and Lampe, Thomas and Parisotto, Emilio and Żołna, Konrad and Reed, Scott and Colmenarejo, Sergio Gómez and Scholz, Jon and Abdolmaleki, Abbas and Groth, Oliver and Regli, Jean-Baptiste and Sushkov, Oleg and Rothörl, Tom and Chen, José Enrique and Aytar, Yusuf and Barker, Dave and Ortiz, Joy and Riedmiller, Martin and Springenberg, Jost Tobias and Hadsell, Raia and Nori, Francesco and Heess, Nicolas},
	month = dec,
	year = {2023},
	note = {arXiv:2306.11706 [cs]},
}

@misc{devlin_navigation_2021,
	title = {Navigation {Turing} {Test} ({NTT}): {Learning} to {Evaluate} {Human}-{Like} {Navigation}},
	shorttitle = {Navigation {Turing} {Test} ({NTT})},
	url = {http://arxiv.org/abs/2105.09637},
	doi = {10.48550/arXiv.2105.09637},
	urldate = {2023-09-22},
	publisher = {arXiv},
	author = {Devlin, Sam and Georgescu, Raluca and Momennejad, Ida and Rzepecki, Jaroslaw and Zuniga, Evelyn and Costello, Gavin and Leroy, Guy and Shaw, Ali and Hofmann, Katja},
	month = jul,
	year = {2021},
	note = {arXiv:2105.09637},
}

@misc{lee_multi-game_2022,
	title = {Multi-{Game} {Decision} {Transformers}},
	url = {http://arxiv.org/abs/2205.15241},
	urldate = {2023-09-08},
	publisher = {arXiv},
	author = {Lee, Kuang-Huei and Nachum, Ofir and Yang, Mengjiao and Lee, Lisa and Freeman, Daniel and Xu, Winnie and Guadarrama, Sergio and Fischer, Ian and Jang, Eric and Michalewski, Henryk and Mordatch, Igor},
	month = oct,
	year = {2022},
	note = {arXiv:2205.15241},
}

@inproceedings{knox_tamer_2008,
	address = {Monterey, CA},
	title = {{TAMER}: {Training} an {Agent} {Manually} via {Evaluative} {Reinforcement}},
	isbn = {978-1-4244-2661-4},
	shorttitle = {{TAMER}},
	url = {http://ieeexplore.ieee.org/document/4640845/},
	doi = {10.1109/DEVLRN.2008.4640845},
	language = {en},
	urldate = {2023-09-16},
	booktitle = {2008 7th {IEEE} {International} {Conference} on {Development} and {Learning}},
	publisher = {IEEE},
	author = {Knox, Bradley and Stone, Peter},
	month = aug,
	year = {2008},
	pages = {292--297},
}

@misc{loshchilov_decoupled_2019,
	title = {Decoupled {Weight} {Decay} {Regularization}},
	url = {http://arxiv.org/abs/1711.05101},
	doi = {10.48550/arXiv.1711.05101},
	urldate = {2024-02-26},
	publisher = {arXiv},
	author = {Loshchilov, Ilya and Hutter, Frank},
	month = jan,
	year = {2019},
	note = {arXiv:1711.05101 [cs, math]},
}

@misc{hendrycks_gaussian_2023,
	title = {Gaussian {Error} {Linear} {Units} ({GELUs})},
	url = {http://arxiv.org/abs/1606.08415},
	doi = {10.48550/arXiv.1606.08415},
	urldate = {2024-02-26},
	publisher = {arXiv},
	author = {Hendrycks, Dan and Gimpel, Kevin},
	month = jun,
	year = {2023},
	note = {arXiv:1606.08415 [cs]},
}

@techreport{google_deepmind_gemma_2024,
	title = {Gemma: {Open} {Models} {Based} on {Gemini} {Research} and {Technology}},
	shorttitle = {Gemma},
	url = {https://storage.googleapis.com/deepmind-media/gemma/gemma-report.pdf},
	language = {en-us},
	urldate = {2024-02-24},
	author = {Google Deepmind, Gemma Team},
	month = feb,
	year = {2024},
}

@misc{ahmadian_back_2024,
	title = {Back to {Basics}: {Revisiting} {REINFORCE} {Style} {Optimization} for {Learning} from {Human} {Feedback} in {LLMs}},
	shorttitle = {Back to {Basics}},
	url = {http://arxiv.org/abs/2402.14740},
	urldate = {2024-02-24},
	publisher = {arXiv},
	author = {Ahmadian, Arash and Cremer, Chris and Gallé, Matthias and Fadaee, Marzieh and Kreutzer, Julia and Üstün, Ahmet and Hooker, Sara},
	month = feb,
	year = {2024},
	note = {arXiv:2402.14740 [cs]},
}

@misc{liu_convnet_2022,
	title = {A {ConvNet} for the 2020s},
	url = {http://arxiv.org/abs/2201.03545},
	doi = {10.48550/arXiv.2201.03545},
	urldate = {2024-02-23},
	publisher = {arXiv},
	author = {Liu, Zhuang and Mao, Hanzi and Wu, Chao-Yuan and Feichtenhofer, Christoph and Darrell, Trevor and Xie, Saining},
	month = mar,
	year = {2022},
	note = {arXiv:2201.03545 [cs]},
}

@inproceedings{aytemiz_acting_2021,
	title = {Acting with {Style}: {Towards} {Designer}-centred {Reinforcement} {Learning} for the {Video} {Games} {Industry}},
	url = {https://www.microsoft.com/en-us/research/publication/acting-with-style-towards-designer-centred-reinforcement-learning-for-the-video-games-industry/},
	booktitle = {{CHI} 2021 {Workshop} on {Reinforcement} {Learning} for {Humans}, {Computer}, and {Interaction} ({RL4HCI})},
	publisher = {Association for Computing Machinery (ACM)},
	author = {Aytemiz, Batu and Jacob, Mikhail and Devlin, Sam},
	month = may,
	year = {2021},
}

@inproceedings{chen_generative_2020,
	title = {Generative {Pretraining} {From} {Pixels}},
	url = {https://proceedings.mlr.press/v119/chen20s.html},
	language = {en},
	urldate = {2023-09-19},
	booktitle = {Proceedings of the 37th {International} {Conference} on {Machine} {Learning}},
	publisher = {PMLR},
	author = {Chen, Mark and Radford, Alec and Child, Rewon and Wu, Jeffrey and Jun, Heewoo and Luan, David and Sutskever, Ilya},
	month = nov,
	year = {2020},
	note = {ISSN: 2640-3498},
	pages = {1691--1703},
}

@article{radford_improving_2018,
	title = {Improving {Language} {Understanding} by {Generative} {Pre}-{Training}},
	language = {en},
	author = {Radford, Alec and Narasimhan, Karthik and Salimans, Tim and Sutskever, Ilya},
	year = {2018},
}

@misc{hu_lora_2021,
	title = {{LoRA}: {Low}-{Rank} {Adaptation} of {Large} {Language} {Models}},
	shorttitle = {{LoRA}},
	url = {http://arxiv.org/abs/2106.09685},
	doi = {10.48550/arXiv.2106.09685},
	urldate = {2023-09-17},
	publisher = {arXiv},
	author = {Hu, Edward J. and Shen, Yelong and Wallis, Phillip and Allen-Zhu, Zeyuan and Li, Yuanzhi and Wang, Shean and Wang, Lu and Chen, Weizhu},
	month = oct,
	year = {2021},
	note = {arXiv:2106.09685 [cs]},
}

@article{williams_simple_1992,
	title = {Simple statistical gradient-following algorithms for connectionist reinforcement learning},
	volume = {8},
	issn = {1573-0565},
	url = {https://doi.org/10.1007/BF00992696},
	doi = {10.1007/BF00992696},
	language = {en},
	number = {3},
	urldate = {2023-09-17},
	journal = {Machine Learning},
	author = {Williams, Ronald J.},
	month = may,
	year = {1992},
	pages = {229--256},
}

@book{elo_rating_1978,
	title = {The {Rating} of {Chessplayers}, {Past} and {Present}},
	isbn = {978-0-668-04721-0},
	language = {en},
	publisher = {Arco Pub.},
	author = {Elo, Arpad E.},
	year = {1978},
	note = {Google-Books-ID: 8pMnAQAAMAAJ},
}

@article{bradley_rank_1952,
	title = {Rank {Analysis} of {Incomplete} {Block} {Designs}: {I}. {The} {Method} of {Paired} {Comparisons}},
	volume = {39},
	issn = {0006-3444},
	shorttitle = {Rank {Analysis} of {Incomplete} {Block} {Designs}},
	url = {https://www.jstor.org/stable/2334029},
	doi = {10.2307/2334029},
	number = {3/4},
	urldate = {2023-09-17},
	journal = {Biometrika},
	author = {Bradley, Ralph Allan and Terry, Milton E.},
	year = {1952},
	note = {Publisher: [Oxford University Press, Biometrika Trust]},
	pages = {324--345},
}

@article{radford_language_2019,
	title = {Language {Models} are {Unsupervised} {Multitask} {Learners}},
	language = {en},
	author = {Radford, Alec and Wu, Jeffrey and Child, Rewon and Luan, David and Amodei, Dario and Sutskever, Ilya},
	month = feb,
	year = {2019},
}

@misc{openai_introducing_2022,
	title = {Introducing {ChatGPT}},
	url = {https://openai.com/blog/chatgpt},
	language = {en-US},
	urldate = {2023-09-17},
	author = {{OpenAI}},
	year = {2022},
}

@misc{banerjee_benchmarking_2023,
	title = {Benchmarking {LLM} powered {Chatbots}: {Methods} and {Metrics}},
	shorttitle = {Benchmarking {LLM} powered {Chatbots}},
	url = {http://arxiv.org/abs/2308.04624},
	doi = {10.48550/arXiv.2308.04624},
	urldate = {2023-09-17},
	publisher = {arXiv},
	author = {Banerjee, Debarag and Singh, Pooja and Avadhanam, Arjun and Srivastava, Saksham},
	month = aug,
	year = {2023},
	note = {arXiv:2308.04624 [cs]},
}

@misc{spatharioti_comparing_2023,
	title = {Comparing {Traditional} and {LLM}-based {Search} for {Consumer} {Choice}: {A} {Randomized} {Experiment}},
	shorttitle = {Comparing {Traditional} and {LLM}-based {Search} for {Consumer} {Choice}},
	url = {http://arxiv.org/abs/2307.03744},
	doi = {10.48550/arXiv.2307.03744},
	urldate = {2023-09-17},
	publisher = {arXiv},
	author = {Spatharioti, Sofia Eleni and Rothschild, David M. and Goldstein, Daniel G. and Hofman, Jake M.},
	month = jul,
	year = {2023},
	note = {arXiv:2307.03744 [cs]},
}

@misc{hernandez_scaling_2021,
	title = {Scaling {Laws} for {Transfer}},
	url = {http://arxiv.org/abs/2102.01293},
	doi = {10.48550/arXiv.2102.01293},
	urldate = {2023-09-17},
	publisher = {arXiv},
	author = {Hernandez, Danny and Kaplan, Jared and Henighan, Tom and McCandlish, Sam},
	month = feb,
	year = {2021},
	note = {arXiv:2102.01293 [cs]},
}

@misc{bai_training_2022,
	title = {Training a {Helpful} and {Harmless} {Assistant} with {Reinforcement} {Learning} from {Human} {Feedback}},
	url = {http://arxiv.org/abs/2204.05862},
	urldate = {2023-09-17},
	publisher = {arXiv},
	author = {Bai, Yuntao and Jones, Andy and Ndousse, Kamal and Askell, Amanda and Chen, Anna and DasSarma, Nova and Drain, Dawn and Fort, Stanislav and Ganguli, Deep and Henighan, Tom and Joseph, Nicholas and Kadavath, Saurav and Kernion, Jackson and Conerly, Tom and El-Showk, Sheer and Elhage, Nelson and Hatfield-Dodds, Zac and Hernandez, Danny and Hume, Tristan and Johnston, Scott and Kravec, Shauna and Lovitt, Liane and Nanda, Neel and Olsson, Catherine and Amodei, Dario and Brown, Tom and Clark, Jack and McCandlish, Sam and Olah, Chris and Mann, Ben and Kaplan, Jared},
	month = apr,
	year = {2022},
	note = {arXiv:2204.05862 [cs]},
}

@misc{lee_pebble_2021,
	title = {{PEBBLE}: {Feedback}-{Efficient} {Interactive} {Reinforcement} {Learning} via {Relabeling} {Experience} and {Unsupervised} {Pre}-training},
	shorttitle = {{PEBBLE}},
	url = {http://arxiv.org/abs/2106.05091},
	doi = {10.48550/arXiv.2106.05091},
	urldate = {2023-09-17},
	publisher = {arXiv},
	author = {Lee, Kimin and Smith, Laura and Abbeel, Pieter},
	month = jun,
	year = {2021},
	note = {arXiv:2106.05091 [cs]},
}

@article{wirth_survey_2017,
	title = {A {Survey} of {Preference}-{Based} {Reinforcement} {Learning} {Methods}},
	language = {en},
	author = {Wirth, Christian and Akrour, Riad and Neumann, Gerhard and Fürnkranz, Johannes},
	year = {2017},
}

@inproceedings{bennett_netflix_2009,
	title = {The {Netflix} {Prize}},
	author = {Bennett, James and Lanning, Stanley and Netflix, Netflix},
	month = jan,
	year = {2009},
}

@misc{vecerik_leveraging_2018,
	title = {Leveraging {Demonstrations} for {Deep} {Reinforcement} {Learning} on {Robotics} {Problems} with {Sparse} {Rewards}},
	url = {http://arxiv.org/abs/1707.08817},
	doi = {10.48550/arXiv.1707.08817},
	urldate = {2023-09-17},
	publisher = {arXiv},
	author = {Vecerik, Mel and Hester, Todd and Scholz, Jonathan and Wang, Fumin and Pietquin, Olivier and Piot, Bilal and Heess, Nicolas and Rothörl, Thomas and Lampe, Thomas and Riedmiller, Martin},
	month = oct,
	year = {2018},
	note = {arXiv:1707.08817 [cs]},
}

@misc{brohan_rt-2_2023,
	title = {{RT}-2: {Vision}-{Language}-{Action} {Models} {Transfer} {Web} {Knowledge} to {Robotic} {Control}},
	shorttitle = {{RT}-2},
	url = {http://arxiv.org/abs/2307.15818},
	doi = {10.48550/arXiv.2307.15818},
	urldate = {2023-09-17},
	publisher = {arXiv},
	author = {Brohan, Anthony and Brown, Noah and Carbajal, Justice and Chebotar, Yevgen and Chen, Xi and Choromanski, Krzysztof and Ding, Tianli and Driess, Danny and Dubey, Avinava and Finn, Chelsea and Florence, Pete and Fu, Chuyuan and Arenas, Montse Gonzalez and Gopalakrishnan, Keerthana and Han, Kehang and Hausman, Karol and Herzog, Alexander and Hsu, Jasmine and Ichter, Brian and Irpan, Alex and Joshi, Nikhil and Julian, Ryan and Kalashnikov, Dmitry and Kuang, Yuheng and Leal, Isabel and Lee, Lisa and Lee, Tsang-Wei Edward and Levine, Sergey and Lu, Yao and Michalewski, Henryk and Mordatch, Igor and Pertsch, Karl and Rao, Kanishka and Reymann, Krista and Ryoo, Michael and Salazar, Grecia and Sanketi, Pannag and Sermanet, Pierre and Singh, Jaspiar and Singh, Anikait and Soricut, Radu and Tran, Huong and Vanhoucke, Vincent and Vuong, Quan and Wahid, Ayzaan and Welker, Stefan and Wohlhart, Paul and Wu, Jialin and Xia, Fei and Xiao, Ted and Xu, Peng and Xu, Sichun and Yu, Tianhe and Zitkovich, Brianna},
	month = jul,
	year = {2023},
	note = {arXiv:2307.15818 [cs]},
}

@misc{chen_evaluating_2021,
	title = {Evaluating {Large} {Language} {Models} {Trained} on {Code}},
	url = {http://arxiv.org/abs/2107.03374},
	doi = {10.48550/arXiv.2107.03374},
	urldate = {2023-09-16},
	publisher = {arXiv},
	author = {Chen, Mark and Tworek, Jerry and Jun, Heewoo and Yuan, Qiming and Pinto, Henrique Ponde de Oliveira and Kaplan, Jared and Edwards, Harri and Burda, Yuri and Joseph, Nicholas and Brockman, Greg and Ray, Alex and Puri, Raul and Krueger, Gretchen and Petrov, Michael and Khlaaf, Heidy and Sastry, Girish and Mishkin, Pamela and Chan, Brooke and Gray, Scott and Ryder, Nick and Pavlov, Mikhail and Power, Alethea and Kaiser, Lukasz and Bavarian, Mohammad and Winter, Clemens and Tillet, Philippe and Such, Felipe Petroski and Cummings, Dave and Plappert, Matthias and Chantzis, Fotios and Barnes, Elizabeth and Herbert-Voss, Ariel and Guss, William Hebgen and Nichol, Alex and Paino, Alex and Tezak, Nikolas and Tang, Jie and Babuschkin, Igor and Balaji, Suchir and Jain, Shantanu and Saunders, William and Hesse, Christopher and Carr, Andrew N. and Leike, Jan and Achiam, Josh and Misra, Vedant and Morikawa, Evan and Radford, Alec and Knight, Matthew and Brundage, Miles and Murati, Mira and Mayer, Katie and Welinder, Peter and McGrew, Bob and Amodei, Dario and McCandlish, Sam and Sutskever, Ilya and Zaremba, Wojciech},
	month = jul,
	year = {2021},
	note = {arXiv:2107.03374 [cs]},
}

@inproceedings{ibarz_reward_2018,
	title = {Reward learning from human preferences and demonstrations in {Atari}},
	volume = {31},
	url = {https://proceedings.neurips.cc/paper/2018/hash/8cbe9ce23f42628c98f80fa0fac8b19a-Abstract.html},
	urldate = {2023-09-16},
	booktitle = {Advances in {Neural} {Information} {Processing} {Systems}},
	publisher = {Curran Associates, Inc.},
	author = {Ibarz, Borja and Leike, Jan and Pohlen, Tobias and Irving, Geoffrey and Legg, Shane and Amodei, Dario},
	year = {2018},
}

@misc{hester_deep_2017,
	title = {Deep {Q}-learning from {Demonstrations}},
	url = {http://arxiv.org/abs/1704.03732},
	urldate = {2023-09-16},
	publisher = {arXiv},
	author = {Hester, Todd and Vecerik, Matej and Pietquin, Olivier and Lanctot, Marc and Schaul, Tom and Piot, Bilal and Horgan, Dan and Quan, John and Sendonaris, Andrew and Dulac-Arnold, Gabriel and Osband, Ian and Agapiou, John and Leibo, Joel Z. and Gruslys, Audrunas},
	month = nov,
	year = {2017},
	note = {arXiv:1704.03732 [cs]},
}

@misc{openai_gpt-4_2023,
	title = {{GPT}-4 {Technical} {Report}},
	url = {http://arxiv.org/abs/2303.08774},
	urldate = {2023-09-16},
	publisher = {arXiv},
	author = {OpenAI},
	month = mar,
	year = {2023},
	note = {arXiv:2303.08774 [cs]},
}

@misc{milani_towards_2023,
	title = {Towards {Solving} {Fuzzy} {Tasks} with {Human} {Feedback}: {A} {Retrospective} of the {MineRL} {BASALT} 2022 {Competition}},
	shorttitle = {Towards {Solving} {Fuzzy} {Tasks} with {Human} {Feedback}},
	url = {http://arxiv.org/abs/2303.13512},
	urldate = {2023-09-16},
	publisher = {arXiv},
	author = {Milani, Stephanie and Kanervisto, Anssi and Ramanauskas, Karolis and Schulhoff, Sander and Houghton, Brandon and Mohanty, Sharada and Galbraith, Byron and Chen, Ke and Song, Yan and Zhou, Tianze and Yu, Bingquan and Liu, He and Guan, Kai and Hu, Yujing and Lv, Tangjie and Malato, Federico and Leopold, Florian and Raut, Amogh and Hautamäki, Ville and Melnik, Andrew and Ishida, Shu and Henriques, João F. and Klassert, Robert and Laurito, Walter and Novoseller, Ellen and Goecks, Vinicius G. and Waytowich, Nicholas and Watkins, David and Miller, Josh and Shah, Rohin},
	month = mar,
	year = {2023},
	note = {arXiv:2303.13512 [cs]},
}

@misc{kaplan_scaling_2020,
	title = {Scaling {Laws} for {Neural} {Language} {Models}},
	url = {http://arxiv.org/abs/2001.08361},
	urldate = {2023-09-16},
	publisher = {arXiv},
	author = {Kaplan, Jared and McCandlish, Sam and Henighan, Tom and Brown, Tom B. and Chess, Benjamin and Child, Rewon and Gray, Scott and Radford, Alec and Wu, Jeffrey and Amodei, Dario},
	month = jan,
	year = {2020},
	note = {arXiv:2001.08361 [cs, stat]},
}

@article{zhang_recent_2021,
	title = {Recent {Advances} in {Leveraging} {Human} {Guidance} for {Sequential} {Decision}-{Making} {Tasks}},
	volume = {35},
	issn = {1387-2532, 1573-7454},
	url = {http://arxiv.org/abs/2107.05825},
	doi = {10.1007/s10458-021-09514-w},
	number = {2},
	urldate = {2023-09-16},
	journal = {Autonomous Agents and Multi-Agent Systems},
	author = {Zhang, Ruohan and Torabi, Faraz and Warnell, Garrett and Stone, Peter},
	month = oct,
	year = {2021},
	note = {arXiv:2107.05825 [cs]},
	pages = {31},
}

@misc{beltagy_longformer_2020,
	title = {Longformer: {The} {Long}-{Document} {Transformer}},
	shorttitle = {Longformer},
	url = {http://arxiv.org/abs/2004.05150},
	doi = {10.48550/arXiv.2004.05150},
	urldate = {2023-09-16},
	publisher = {arXiv},
	author = {Beltagy, Iz and Peters, Matthew E. and Cohan, Arman},
	month = dec,
	year = {2020},
	note = {arXiv:2004.05150 [cs]},
}

@misc{rafailov_direct_2023,
	title = {Direct {Preference} {Optimization}: {Your} {Language} {Model} is {Secretly} a {Reward} {Model}},
	shorttitle = {Direct {Preference} {Optimization}},
	url = {http://arxiv.org/abs/2305.18290},
	doi = {10.48550/arXiv.2305.18290},
	urldate = {2023-09-16},
	publisher = {arXiv},
	author = {Rafailov, Rafael and Sharma, Archit and Mitchell, Eric and Ermon, Stefano and Manning, Christopher D. and Finn, Chelsea},
	month = may,
	year = {2023},
	note = {arXiv:2305.18290 [cs]},
}

@misc{gulcehre_reinforced_2023,
	title = {Reinforced {Self}-{Training} ({ReST}) for {Language} {Modeling}},
	url = {http://arxiv.org/abs/2308.08998},
	urldate = {2023-09-16},
	publisher = {arXiv},
	author = {Gulcehre, Caglar and Paine, Tom Le and Srinivasan, Srivatsan and Konyushkova, Ksenia and Weerts, Lotte and Sharma, Abhishek and Siddhant, Aditya and Ahern, Alex and Wang, Miaosen and Gu, Chenjie and Macherey, Wolfgang and Doucet, Arnaud and Firat, Orhan and de Freitas, Nando},
	month = aug,
	year = {2023},
	note = {arXiv:2308.08998 [cs]},
}

@misc{ziegler_fine-tuning_2020,
	title = {Fine-{Tuning} {Language} {Models} from {Human} {Preferences}},
	url = {http://arxiv.org/abs/1909.08593},
	urldate = {2023-09-15},
	publisher = {arXiv},
	author = {Ziegler, Daniel M. and Stiennon, Nisan and Wu, Jeffrey and Brown, Tom B. and Radford, Alec and Amodei, Dario and Christiano, Paul and Irving, Geoffrey},
	month = jan,
	year = {2020},
	note = {arXiv:1909.08593 [cs, stat]},
}

@misc{chung_scaling_2022,
	title = {Scaling {Instruction}-{Finetuned} {Language} {Models}},
	url = {http://arxiv.org/abs/2210.11416},
	doi = {10.48550/arXiv.2210.11416},
	urldate = {2023-09-14},
	publisher = {arXiv},
	author = {Chung, Hyung Won and Hou, Le and Longpre, Shayne and Zoph, Barret and Tay, Yi and Fedus, William and Li, Yunxuan and Wang, Xuezhi and Dehghani, Mostafa and Brahma, Siddhartha and Webson, Albert and Gu, Shixiang Shane and Dai, Zhuyun and Suzgun, Mirac and Chen, Xinyun and Chowdhery, Aakanksha and Castro-Ros, Alex and Pellat, Marie and Robinson, Kevin and Valter, Dasha and Narang, Sharan and Mishra, Gaurav and Yu, Adams and Zhao, Vincent and Huang, Yanping and Dai, Andrew and Yu, Hongkun and Petrov, Slav and Chi, Ed H. and Dean, Jeff and Devlin, Jacob and Roberts, Adam and Zhou, Denny and Le, Quoc V. and Wei, Jason},
	month = dec,
	year = {2022},
	note = {arXiv:2210.11416 [cs]},
}

@misc{casper_open_2023,
	title = {Open {Problems} and {Fundamental} {Limitations} of {Reinforcement} {Learning} from {Human} {Feedback}},
	url = {http://arxiv.org/abs/2307.15217},
	urldate = {2023-09-08},
	publisher = {arXiv},
	author = {Casper, Stephen and Davies, Xander and Shi, Claudia and Gilbert, Thomas Krendl and Scheurer, Jérémy and Rando, Javier and Freedman, Rachel and Korbak, Tomasz and Lindner, David and Freire, Pedro and Wang, Tony and Marks, Samuel and Segerie, Charbel-Raphaël and Carroll, Micah and Peng, Andi and Christoffersen, Phillip and Damani, Mehul and Slocum, Stewart and Anwar, Usman and Siththaranjan, Anand and Nadeau, Max and Michaud, Eric J. and Pfau, Jacob and Krasheninnikov, Dmitrii and Chen, Xin and Langosco, Lauro and Hase, Peter and Bıyık, Erdem and Dragan, Anca and Krueger, David and Sadigh, Dorsa and Hadfield-Menell, Dylan},
	month = jul,
	year = {2023},
	note = {arXiv:2307.15217 [cs]},
}

@misc{baker_video_2022,
	title = {Video {PreTraining} ({VPT}): {Learning} to {Act} by {Watching} {Unlabeled} {Online} {Videos}},
	shorttitle = {Video {PreTraining} ({VPT})},
	url = {http://arxiv.org/abs/2206.11795},
	urldate = {2023-09-08},
	publisher = {arXiv},
	author = {Baker, Bowen and Akkaya, Ilge and Zhokhov, Peter and Huizinga, Joost and Tang, Jie and Ecoffet, Adrien and Houghton, Brandon and Sampedro, Raul and Clune, Jeff},
	month = jun,
	year = {2022},
	note = {arXiv:2206.11795 [cs]},
}

@misc{xu_imagereward_2023,
	title = {{ImageReward}: {Learning} and {Evaluating} {Human} {Preferences} for {Text}-to-{Image} {Generation}},
	shorttitle = {{ImageReward}},
	url = {http://arxiv.org/abs/2304.05977},
	urldate = {2023-09-08},
	publisher = {arXiv},
	author = {Xu, Jiazheng and Liu, Xiao and Wu, Yuchen and Tong, Yuxuan and Li, Qinkai and Ding, Ming and Tang, Jie and Dong, Yuxiao},
	month = jun,
	year = {2023},
	note = {arXiv:2304.05977 [cs]},
}

@misc{touvron_llama_2023,
	title = {Llama 2: {Open} {Foundation} and {Fine}-{Tuned} {Chat} {Models}},
	shorttitle = {Llama 2},
	url = {http://arxiv.org/abs/2307.09288},
	doi = {10.48550/arXiv.2307.09288},
	urldate = {2023-09-08},
	publisher = {arXiv},
	author = {Touvron, Hugo and Martin, Louis and Stone, Kevin and Albert, Peter and Almahairi, Amjad and Babaei, Yasmine and Bashlykov, Nikolay and Batra, Soumya and Bhargava, Prajjwal and Bhosale, Shruti and Bikel, Dan and Blecher, Lukas and Ferrer, Cristian Canton and Chen, Moya and Cucurull, Guillem and Esiobu, David and Fernandes, Jude and Fu, Jeremy and Fu, Wenyin and Fuller, Brian and Gao, Cynthia and Goswami, Vedanuj and Goyal, Naman and Hartshorn, Anthony and Hosseini, Saghar and Hou, Rui and Inan, Hakan and Kardas, Marcin and Kerkez, Viktor and Khabsa, Madian and Kloumann, Isabel and Korenev, Artem and Koura, Punit Singh and Lachaux, Marie-Anne and Lavril, Thibaut and Lee, Jenya and Liskovich, Diana and Lu, Yinghai and Mao, Yuning and Martinet, Xavier and Mihaylov, Todor and Mishra, Pushkar and Molybog, Igor and Nie, Yixin and Poulton, Andrew and Reizenstein, Jeremy and Rungta, Rashi and Saladi, Kalyan and Schelten, Alan and Silva, Ruan and Smith, Eric Michael and Subramanian, Ranjan and Tan, Xiaoqing Ellen and Tang, Binh and Taylor, Ross and Williams, Adina and Kuan, Jian Xiang and Xu, Puxin and Yan, Zheng and Zarov, Iliyan and Zhang, Yuchen and Fan, Angela and Kambadur, Melanie and Narang, Sharan and Rodriguez, Aurelien and Stojnic, Robert and Edunov, Sergey and Scialom, Thomas},
	month = jul,
	year = {2023},
	note = {arXiv:2307.09288 [cs]},
}

@misc{abramson_improving_2022,
	title = {Improving {Multimodal} {Interactive} {Agents} with {Reinforcement} {Learning} from {Human} {Feedback}},
	url = {http://arxiv.org/abs/2211.11602},
	urldate = {2023-09-07},
	publisher = {arXiv},
	author = {Abramson, Josh and Ahuja, Arun and Carnevale, Federico and Georgiev, Petko and Goldin, Alex and Hung, Alden and Landon, Jessica and Lhotka, Jirka and Lillicrap, Timothy and Muldal, Alistair and Powell, George and Santoro, Adam and Scully, Guy and Srivastava, Sanjana and von Glehn, Tamara and Wayne, Greg and Wong, Nathaniel and Yan, Chen and Zhu, Rui},
	month = nov,
	year = {2022},
	note = {arXiv:2211.11602 [cs]},
}

@misc{hu_aligning_2023,
	title = {Aligning {Language} {Models} with {Offline} {Reinforcement} {Learning} from {Human} {Feedback}},
	url = {http://arxiv.org/abs/2308.12050},
	urldate = {2023-09-06},
	publisher = {arXiv},
	author = {Hu, Jian and Tao, Li and Yang, June and Zhou, Chandler},
	month = aug,
	year = {2023},
	note = {arXiv:2308.12050 [cs]
version: 1},
}

@article{pomerleau_efficient_1991,
	title = {Efficient {Training} of {Artificial} {Neural} {Networks} for {Autonomous} {Navigation}},
	volume = {3},
	issn = {0899-7667},
	doi = {10.1162/neco.1991.3.1.88},
	number = {1},
	journal = {Neural Computation},
	author = {Pomerleau, Dean A.},
	month = mar,
	year = {1991},
	note = {Conference Name: Neural Computation},
	pages = {88--97},
}

@misc{ba_layer_2016,
	title = {Layer {Normalization}},
	url = {http://arxiv.org/abs/1607.06450},
	doi = {10.48550/arXiv.1607.06450},
	urldate = {2023-04-26},
	publisher = {arXiv},
	author = {Ba, Jimmy Lei and Kiros, Jamie Ryan and Hinton, Geoffrey E.},
	month = jul,
	year = {2016},
	note = {arXiv:1607.06450 [cs, stat]},
}

@inproceedings{stiennon_learning_2020,
	title = {Learning to summarize with human feedback},
	volume = {33},
	url = {https://proceedings.neurips.cc/paper_files/paper/2020/hash/1f89885d556929e98d3ef9b86448f951-Abstract.html},
	urldate = {2023-04-26},
	booktitle = {Advances in {Neural} {Information} {Processing} {Systems}},
	publisher = {Curran Associates, Inc.},
	author = {Stiennon, Nisan and Ouyang, Long and Wu, Jeffrey and Ziegler, Daniel and Lowe, Ryan and Voss, Chelsea and Radford, Alec and Amodei, Dario and Christiano, Paul F},
	year = {2020},
	pages = {3008--3021},
}

@misc{ross_reduction_2011,
	title = {A {Reduction} of {Imitation} {Learning} and {Structured} {Prediction} to {No}-{Regret} {Online} {Learning}},
	url = {http://arxiv.org/abs/1011.0686},
	doi = {10.48550/arXiv.1011.0686},
	urldate = {2023-04-26},
	publisher = {arXiv},
	author = {Ross, Stephane and Gordon, Geoffrey J. and Bagnell, J. Andrew},
	month = mar,
	year = {2011},
	note = {arXiv:1011.0686 [cs, stat]},
}

@misc{ostrovski_difficulty_2021,
	title = {The {Difficulty} of {Passive} {Learning} in {Deep} {Reinforcement} {Learning}},
	url = {http://arxiv.org/abs/2110.14020},
	doi = {10.48550/arXiv.2110.14020},
	urldate = {2023-03-06},
	publisher = {arXiv},
	author = {Ostrovski, Georg and Castro, Pablo Samuel and Dabney, Will},
	month = oct,
	year = {2021},
	note = {arXiv:2110.14020 [cs]},
}

@misc{ouyang_training_2022,
	title = {Training language models to follow instructions with human feedback},
	url = {http://arxiv.org/abs/2203.02155},
	doi = {10.48550/arXiv.2203.02155},
	urldate = {2022-12-20},
	publisher = {arXiv},
	author = {Ouyang, Long and Wu, Jeff and Jiang, Xu and Almeida, Diogo and Wainwright, Carroll L. and Mishkin, Pamela and Zhang, Chong and Agarwal, Sandhini and Slama, Katarina and Ray, Alex and Schulman, John and Hilton, Jacob and Kelton, Fraser and Miller, Luke and Simens, Maddie and Askell, Amanda and Welinder, Peter and Christiano, Paul and Leike, Jan and Lowe, Ryan},
	month = mar,
	year = {2022},
	note = {arXiv:2203.02155 [cs]},
}

@misc{christiano_deep_2017,
	title = {Deep reinforcement learning from human preferences},
	url = {http://arxiv.org/abs/1706.03741},
	urldate = {2022-12-20},
	publisher = {arXiv},
	author = {Christiano, Paul and Leike, Jan and Brown, Tom B. and Martic, Miljan and Legg, Shane and Amodei, Dario},
	month = jul,
	year = {2017},
	note = {arXiv:1706.03741 [cs, stat]},
}

@misc{levine_offline_2020,
	title = {Offline {Reinforcement} {Learning}: {Tutorial}, {Review}, and {Perspectives} on {Open} {Problems}},
	shorttitle = {Offline {Reinforcement} {Learning}},
	url = {http://arxiv.org/abs/2005.01643},
	doi = {10.48550/arXiv.2005.01643},
	urldate = {2022-10-10},
	publisher = {arXiv},
	author = {Levine, Sergey and Kumar, Aviral and Tucker, George and Fu, Justin},
	month = nov,
	year = {2020},
	note = {arXiv:2005.01643 [cs, stat]},
}

@techreport{reed_generalist_2022,
	title = {A {Generalist} {Agent}},
	url = {http://arxiv.org/abs/2205.06175},
	number = {arXiv:2205.06175},
	urldate = {2022-05-16},
	institution = {arXiv},
	author = {Reed, Scott and Zolna, Konrad and Parisotto, Emilio and Colmenarejo, Sergio Gomez and Novikov, Alexander and Barth-Maron, Gabriel and Gimenez, Mai and Sulsky, Yury and Kay, Jackie and Springenberg, Jost Tobias and Eccles, Tom and Bruce, Jake and Razavi, Ali and Edwards, Ashley and Heess, Nicolas and Chen, Yutian and Hadsell, Raia and Vinyals, Oriol and Bordbar, Mahyar and de Freitas, Nando},
	month = may,
	year = {2022},
	doi = {10.48550/arXiv.2205.06175},
	note = {arXiv:2205.06175 [cs]
type: article},
}

@article{silver_mastering_2016,
	title = {Mastering the game of {Go} with deep neural networks and tree search},
	volume = {529},
	copyright = {2016 Nature Publishing Group, a division of Macmillan Publishers Limited. All Rights Reserved.},
	issn = {1476-4687},
	url = {https://www.nature.com/articles/nature16961},
	doi = {10.1038/nature16961},
	language = {en},
	number = {7587},
	urldate = {2021-10-11},
	journal = {Nature},
	author = {Silver, David and Huang, Aja and Maddison, Chris J. and Guez, Arthur and Sifre, Laurent and van den Driessche, George and Schrittwieser, Julian and Antonoglou, Ioannis and Panneershelvam, Veda and Lanctot, Marc and Dieleman, Sander and Grewe, Dominik and Nham, John and Kalchbrenner, Nal and Sutskever, Ilya and Lillicrap, Timothy and Leach, Madeleine and Kavukcuoglu, Koray and Graepel, Thore and Hassabis, Demis},
	month = jan,
	year = {2016},
	note = {Bandiera\_abtest: a
Cg\_type: Nature Research Journals
Number: 7587
Primary\_atype: Research
Publisher: Nature Publishing Group
Subject\_term: Computational science;Computer science;Reward
Subject\_term\_id: computational-science;computer-science;reward},
	pages = {484--489},
}

@article{chen_decision_2021,
	title = {Decision {Transformer}: {Reinforcement} {Learning} via {Sequence} {Modeling}},
	shorttitle = {Decision {Transformer}},
	url = {http://arxiv.org/abs/2106.01345},
	urldate = {2021-09-14},
	journal = {arXiv:2106.01345 [cs]},
	author = {Chen, Lili and Lu, Kevin and Rajeswaran, Aravind and Lee, Kimin and Grover, Aditya and Laskin, Michael and Abbeel, Pieter and Srinivas, Aravind and Mordatch, Igor},
	month = jun,
	year = {2021},
	note = {arXiv: 2106.01345},
}

@article{vinyals_grandmaster_2019,
	title = {Grandmaster level in {StarCraft} {II} using multi-agent reinforcement learning},
	volume = {575},
	issn = {0028-0836, 1476-4687},
	url = {http://www.nature.com/articles/s41586-019-1724-z},
	doi = {10.1038/s41586-019-1724-z},
	language = {en},
	number = {7782},
	urldate = {2021-08-02},
	journal = {Nature},
	author = {Vinyals, Oriol and Babuschkin, Igor and Czarnecki, Wojciech M. and Mathieu, Michaël and Dudzik, Andrew and Chung, Junyoung and Choi, David H. and Powell, Richard and Ewalds, Timo and Georgiev, Petko and Oh, Junhyuk and Horgan, Dan and Kroiss, Manuel and Danihelka, Ivo and Huang, Aja and Sifre, Laurent and Cai, Trevor and Agapiou, John P. and Jaderberg, Max and Vezhnevets, Alexander S. and Leblond, Rémi and Pohlen, Tobias and Dalibard, Valentin and Budden, David and Sulsky, Yury and Molloy, James and Paine, Tom L. and Gulcehre, Caglar and Wang, Ziyu and Pfaff, Tobias and Wu, Yuhuai and Ring, Roman and Yogatama, Dani and Wünsch, Dario and McKinney, Katrina and Smith, Oliver and Schaul, Tom and Lillicrap, Timothy and Kavukcuoglu, Koray and Hassabis, Demis and Apps, Chris and Silver, David},
	month = nov,
	year = {2019},
	pages = {350--354},
}

@article{schulman_proximal_2017,
	title = {Proximal {Policy} {Optimization} {Algorithms}},
	url = {http://arxiv.org/abs/1707.06347},
	urldate = {2021-05-27},
	journal = {arXiv:1707.06347 [cs]},
	author = {Schulman, John and Wolski, Filip and Dhariwal, Prafulla and Radford, Alec and Klimov, Oleg},
	month = aug,
	year = {2017},
	note = {arXiv: 1707.06347},
}
\bibliographystyle{icml2025}

\newpage
\appendix
\onecolumn

\section{Discussion of General Procedure for Aligning Agents}

We break down our procedure for training general decision-making agents (see Figure \ref{fig:Figure1}) into five steps:

\begin{enumerate}
    \item{\textbf{Train a Base Imitation Learning Policy}}\newline
        The first ingredient for training large language models is to train a large, scalable transformer architecture with self-supervised next-token prediction on a diverse dataset of human text, to obtain a language prior. In the context of agents on modern console games, we interpret this as imitation learning to predict the next action taken in human gameplay data, to obtain a behavioral prior. Specifically this involves training the transformer autoregressively to learn a policy $p(a_t|o_t,a_{t-1}...,o_{t-H};e_t)$ with optional text conditioning $e_t$ containing embedded context on the game, task or current goal. For the purpose of this work, we consider an agent trained on diverse data within a particular game, but note that given our unified observation and action spaces (visual observations and gamepad actions), it would also be possible to train across games, as explored in previous work \citep{reed_generalist_2022, lee_multi-game_2022}.
        
    \item{\textbf{Supervised Fine-Tune on a Task Relevant Dataset}}\newline
        The next step in the current LLM pipeline is to fine-tune the pre-trained `foundation' model on task relevant data, such as instruction data \citep{chung_scaling_2022}. Pre-trained transformer models have been shown to fine-tune more effectively, essentially increasing the size of the fine-tuning data compared to training from scratch \citep{hernandez_scaling_2021}. For decision-making agents this involves fine-tuning by imitation learning on the task (or game) of interest. For specific behaviors, this could also consist of fine-tuning on demonstrations of the desired behavior by the game designer. The training procedure is equivalent to Step 1.
    \item{\textbf{ Generate Preference Data on Online Trajectories}}\newline 
        Fine-tuned LLMs are subsequently prompted to generate multiple responses which are compared by human labelers to provide preferences. In the context of agents, the prompt becomes the initial observation (and optional context of previous observations, actions and/or textual instructions). The agent is then rolled out from a given initial start state multiple times to collect multiple trajectories. Similarly to LLMs, the temperature of the softmax sampling of the policy for action selection can be increased to generate more diverse behaviors from the agent for easier comparison. A human (e.g. game designer) then provides preferences on these trajectories, such as preferring trajectories where the agent plays in a certain style.
    
    \item{\textbf{Train a Reward Model on Preferences}}\newline     
        A reward model is then trained on these online trajectories, commonly using a Bradley Terry model \citep{bradley_rank_1952} for pairwise comparisons, so that the reward model provides higher reward for preferred trajectories. While this reward model is usually trained from scratch in the context of agents, the modern LLM procedure utilizes the pre-trained or fine-tuned policy model by replacing the action classification head with a scalar regression head. This enables the reward model to also benefit from the pre-training, and share the same knowledge as the agent to reduce out-of-distribution behaviors \citep{ziegler_fine-tuning_2020}. This reward model should ideally capture general behavior preferences to provide more scalable feedback to the agent.

     \item{\textbf{Align the Fine-tuned Model with the Reward Model}}\newline
        Finally the agent can be trained with online reinforcement learning to maximize the reward provided by the reward model, thereby aligning the agent with the game designer's preferences. Since online reinforcement learning can be inefficient, various alternatives can be used, such as Direct Preference Optimization (DPO) \citep{rafailov_direct_2023} with an offline preference dataset or ReST \citep{gulcehre_reinforced_2023}, but online learning is generally preferred \citep{xu_is_2024}. A common failure mode at this stage is reward model over-optimization, where the agent performs behaviors that maximize the reward model output but are not aligned with preferences (also known as reward hacking). If this occurs, regularization towards the original policy can be added or steps 3-5 can be repeated to generate new preferences on the reward hacking behavior which can be used to update the reward model. This may take multiple iterations (e.g. 5 reward model iterations were used for Llama 2 \citep{touvron_llama_2023}), to eventually obtain an aligned agent.
\end{enumerate}

This procedure combines the benefits of large scale pre-training to obtain a widely generalizable agent, with the benefits of reinforcement learning from preferences to obtain an aligned agent. Furthermore, it also combines the benefits of offline learning to avoid the renown sample inefficiency of online reinforcement learning, with the benefits of online learning to alleviate the well-known problems associated with offline agents going out-of-distribution \citep{ross_efficient_2010, ross_reduction_2011} by enabling some active online learning \citep{ostrovski_difficulty_2021} with a sensible behavioral prior. 
\label{app:procedure}
\newpage
Our procedure for aligning agents with preferences can also be summarized in algorithmic form:

\begin{algorithm}[H]
\caption{Aligning Agents with Preferred Behaviours}\label{alg:overview}
\begin{algorithmic}[1]
\REQUIRE Large diverse dataset $D$, task specific dataset $T$, environment $E$, transformer policy $\pi$
\STATE Train policy $\pi$ on large diverse dataset $D$ with imitation learning \citep{pomerleau_efficient_1991}
\begin{equation*}
    \mathcal{L}_{U}(\pi)=\mathbb{E}_D\left [-\log \pi(a^\tau_t|o^\tau_t, o^\tau_{t-1}, ...)\right ]
\end{equation*}
\STATE Fine-tune policy $\pi$ on task specific dataset or demonstrations $T$
\begin{equation*}
    \mathcal{L}_{FT}(\pi)=\mathbb{E}_T \left [-\log \pi(a^\tau_t|o^\tau_t, o^\tau_{t-1}, ...)\right]
\end{equation*}
\STATE Run fine-tuned policy $\pi$ in environment and provide preferences on pairwise comparisons on online trajectories to obtain preference data $(\tau_A \succ \tau_B) \in P$
\STATE Initialise additional reward model $\hat r$ from $\pi$ and train on preferences $P$ using Bradley-Terry model \citep{bradley_rank_1952}
\begin{equation*}
\mathcal{L}(\hat r) = \sum_{(\tau_w, \tau_l)\in P} -\log\bigg( \sigma\big(\hat r(\tau_w) - \hat r(\tau_l)\big)\bigg)
\end{equation*}
\STATE Train fine-tuned policy $\pi_{RL}$ using online reinforcement learning, using reward model $\hat r$ to provide return for online trajectory $\tau$ to align agent with preferences. For example, using REINFORCE \citep{williams_simple_1992} with Kullback–Leibler divergence regularization parameterized by $\beta$:
\begin{equation*}
    \mathcal{L}_{RL}(\pi_{RL}) = -\mathbb{E}_{\pi_{RL}}\left[\sum_{t=1}^T \gamma^t \hat r (\tau) \log \pi_{RL}(a_t|o_t, ...)  - \beta \log \left(\frac{\pi_{RL}(a_t|o_t,)}{\pi_{FT}(a_t|o_t,)}\right) \right]
\end{equation*}
\end{algorithmic}
\end{algorithm}\label{app:algorithm}
\section{Bleeding Edge Game Human Data Collection}
Human gameplay data was recorded as part of the regular gameplay process, in order to enable in-game functionality as well as to support research. In game, recordings allow players to view their past games to improve their skills and for entertainment. Games were recorded on the servers that hosted the games in the form of replay files, which include a representation of the internal game state and controller actions of all players. 

Data collection was covered by an End User License Agreement (EULA) to which players agreed when logging in to play the game for the first time. Our use of the recorded human gameplay data for this specific research was governed by a data sharing agreement with the game studio, and approved by our institution’s IRB. To minimize risks regarding data privacy, any personally identifiable information (Xbox user ID) was removed when extracting the data used for this study from the original replays.\label{app:data}
\section{Architectures and Training Details}\label{app:details}

\subsection{Base Model}

For our base model ($\sim103$M parameters) we use a GPT-2 causal transformer architecture with $8$ layers with $1024$ hidden dim. Each attention layer has 8 heads, and the feedforward layers have a hidden dim of $4096$.

Each image is resized to be of shape $128\times128\times3$, divided by 255 to put its values in $[0,1]$, and is then fed into a convolutional encoder to map it to a $1024$ dimensional vector.

The first layer of the conv net has kernels of shape $8\times8$, with a stride of $4$, and a padding of $3$ and maps to $16$ channel dimension.
This is followed by 4 lots of ConvNext \citep{liu_convnet_2022} and downsampling blocks (kernel of shape $3\times3$, stride of $2$, padding of $1$, doubling the channel dimension).
Finally, a LayerNorm \citep{ba_layer_2016} is applied to the output. 

The transformer operates on sequences of 32 timesteps using learnt positional encodings. The output of the transformer is layernormed, and then fed into an MLP with a single hidden layer of $1024$ dimensions with a GELU \citep{hendrycks_gaussian_2023} non-linearity.

For our optimiser we use AdamW \citep{loshchilov_decoupled_2019} with a learning rate of $1e-4$ and a weight decay of $1e-4$.
We use a batch size of 256 with a learning rate warmup period of 1000 updates and a gradient clipping value of $1$. We train with the same image augmentations as used by \citep{baker_video_2022}, and filter out all no-op actions. 

We used 8 32GB Nvidia V100 GPUs for approximately 4 days to train the base model.

\subsection{Fine-Tuning of Base Model}

We train for 1500 batches of size 128 with a learning rate of $1e-6$ with the same image augmentations and no-op filtering as for pre-training with $200$ warmup steps.

We used 4 48GB Nvidia A6000 GPUs for less than 1 hour for fine-tuning.

\subsection{Reward Models}

Each trajectory is padded up to the maximum length of 100 (with black images) before being fed into the reward models.

For the \texttt{Random Encoder} model we randomly initialise a linear layer to randomly project the flattened values of the image to a $512$ dimensional vector. This linear layer is not trained.

For the \texttt{Agent Encoder} model, we feed the trajectory into the fine-tuned agent and take the layernormed output of the transformer, corresponding to timesteps $0$ up to $100$, as $1024$ dimensional embeddings. This is larger since it must capture the 32 context steps rather than just a single observation. The parameters of the fine-tuned agent are not trained.

For both models we then feed these vectors into an MLP with a GELU non-linearity and a hidden layer of $256$ and and output dimension of $3$.
Each of the 100 3-dimensional vectors are concatenated together and then fed into another MLP with a GELU, $256$  hidden dimension, and an output of $1$.

To train the reward models we use a minibatch of size $2048$, learning rate of $1e-4$, and an $L_2$ regularisation penalty of $0.1$.
We train all models for $200$ epochs, except for the largest training set size of $1000$ trajectories which we train for $50$ epochs.

After training, we compute the minimum and maximum outputs of the reward model on the training set.
These are then used to normalise the output of the reward model to lie within $[0, 1]$.

We used 4 48GB Nvidia A6000 GPUs for a few hours in total for training all reward models.

\subsection{Alignment Training}

\textbf{Preference Fine-Tuning:}
after training the reward model on its dataset of $M$ trajectories (which result in a dataset of up to $N = (M)(M-1)/2$ comparisions), we compute the reward for each of these $M$ trajectories.
We then sort them by magnitude, and take the top $20\%$ of these as a smaller dataset to perform behaviour cloning on.

For this final step of BC, we use a learning rate of $1e-5$ with $1000$ updates on minibatches of size $256$.
We only train the parameters of the MLP after the transformer layers.

We again used 4 48GB Nvidia A6000 GPUs for less than an hour for preference fine-tuning.

\textbf{Reinforcement Learning:}
in our experiments we use an undiscounted REINFORCE loss for 9600 episodes, using a batch size of 16 to give 600 parameter updates. Each episode is up to 100 timesteps or around 10s, so this corresponds to around 1 day of real time gameplay in total. We use a learning rate of $1e-4$ and once again only train the parameters of the MLP after the transformer layers.
If an error occurs during an episode's rollout, we simply drop the samples from that episode and subsequently use a smaller batch size for the update.

We used 16 Standard NV12 machines (each with 2 Nvidia Tesla M60 GPUs for game rendering) on Azure Batch, requiring one machine around one day per seed for agent alignment.

The total compute used for the entire research project therefore represents of the order of weeks of total compute time, with preliminary experiments not included of the order of days of compute.\label{app:archs}
\section{Discussion of Offline to Online Distribution Shift}

The nature of using a real AAA video game is such that there is significant offline to online distribution shift between our offline training data and our online evaluation, as discussed in Section \ref{subsec:pretraining}. Typical visual distribution shifts in our setting, due to character selection and customizable visual modifications, are shown below in Figure \ref{fig:o2o_screenshots}.  


\begin{figure}[h]
  \begin{subfigure}{0.2\textwidth}
    \includegraphics[width=\linewidth]{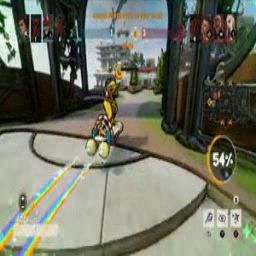}
    \caption{} \label{fig:charachter_1}
  \end{subfigure}%
  \hspace*{\fill}   
  \begin{subfigure}{0.2\textwidth}
    \includegraphics[width=\linewidth]{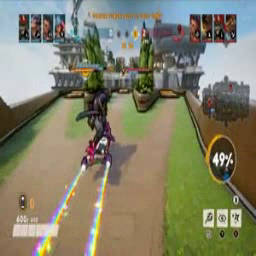}
    \caption{} \label{fig:charachter_2}
  \end{subfigure}%
  \hspace*{\fill}   
  \begin{subfigure}{0.2\textwidth}
    \includegraphics[width=\linewidth]{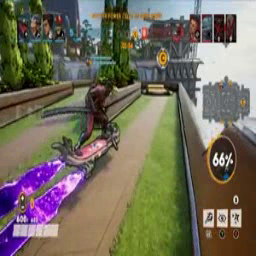}
    \caption{} \label{fig:charachter_3}
      \end{subfigure}%
  \hspace*{\fill}   %
  \begin{subfigure}{0.2\textwidth}
    \includegraphics[width=\linewidth]{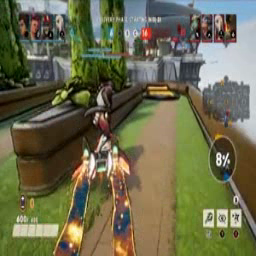}
    \caption{} \label{fig:charachter_4}
    \end{subfigure}%
  \hspace*{\fill}   %
    \begin{subfigure}{0.2\textwidth}
    \includegraphics[width=\linewidth]{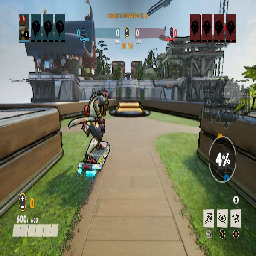}
    \caption{} \label{fig:charachter_ours}
    \end{subfigure}%
  \hspace*{\fill}   %

\caption{Screenshots of various characters with visual modifications contained within the general pre-training data. All are completely representative of the input provided to our agent (only $256\times256$ rather than $128\times128$ resolution). Figure (e) demonstrates the agent we use for online evaluation.} \label{fig:o2o_screenshots} 
\end{figure}

It is also well established that offline-only imitation learning usually suffers from online distribution drift, particularly when learning from pixels in a continuous and partially observable environment \citep{ross_efficient_2010, ross_reduction_2011}. Both of these sources of distribution drift often lead to significant accumulating error in an online trajectory. As a result, imitation learning agents may not consistently perform imitated behaviors online even if the loss has converged on the offline data. This challenge is fundamental to offline learning, as the agent cannot learn from its own observations and actions \citep{ostrovski_difficulty_2021}. This motivates the need to have some online fine-tuning (in our case from preferences) to refine the policy to consistently perform the desired behavior online. 

Figure \ref{fig:FTjumppads} demonstrates that even after supervised fine-tuning on a dataset of trajectories that always reach a jumppad and involve the same character, a significant proportion of trajectories ($\sim10\%$) do not reach any jumppad. However, after online fine-tuning, we are able to increase the overall success rate to nearly 100\% as shown in Appendix~\ref{app:alignedJPdistributions}. We note that this is analogous to recent findings for LLMs that online training (when appropriately configured) can lead to performance improvements over offline-only training \citep{xu_is_2024}.

\label{app:o2oshift}
\newpage
\section{Preliminary Model Scaling and Alignment Analysis}\label{app:scalingablation}

\subsection{Preliminary Model Scaling Analysis}
To further justify the model size and investigate scaling properties of our transformer policy, we also trained smaller models of 4M and 25M parameters on our task-specific dataset, using the same procedure as above. The architecture of our 2 smaller models is identical to that of the base model described in Section \ref{app:details}, except for:

\begin{itemize}
    \item \textbf{$\sim$4M:} $4$ layers with $256$ hidden dims and $4$ heads for each attention layer.
    \item \textbf{$\sim$25M:} $8$ layers with $512$ hidden dims and $8$ heads for each attention layer.
    \item (\textbf{$\sim$103M:} $8$ layers with $1024$ hidden dims and $8$ heads for each attention layer.)
\end{itemize}



\begin{figure}[h]
\vspace{0mm}
    \includegraphics[width=\linewidth]{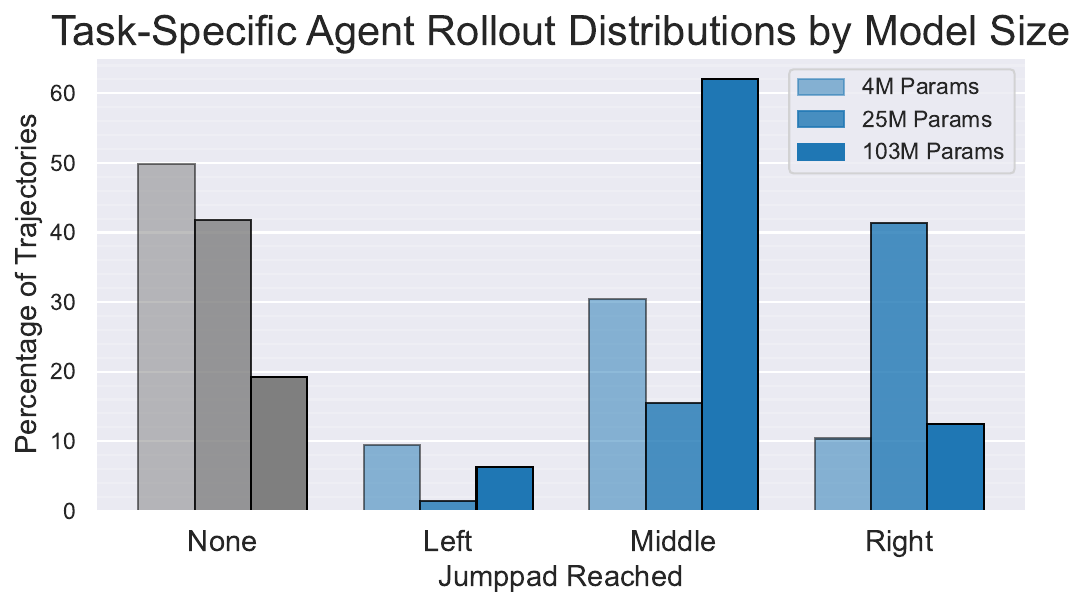}
    
      \vspace{-4mm}
      \caption{Distribution of jumppads reached by various size models (4M, 25M and 103M parameters) trained from scratch on the task-specific dataset used for fine-tuning the base model. Importantly, we see that task failure rates (the grey bars on the left) decrease as the number of parameters increases, even given the same size of dataset and number of training updates.}
\vspace{0mm}
\label{fig:ablationscalingjumppads}
\end{figure}

The jumppad distributions for these models are provided in Figure~\ref{fig:ablationscalingjumppads}. We see that these smaller models have much greater failure rates that the larger 103M parameter models shown in Figure~\ref{fig:ablationFTjumppads}, demonstrating that larger models are beneficial for imitation learning from pixels even on this relatively small task-specific dataset of 300 trajectories. While larger models still may be beneficial (particularly with pre-training on our large unsupervised dataset described in Section \ref{subsec:environment}), larger models would further increase the crucial inference cost and we find that 103M parameters is sufficient for further alignment, as demonstrated in Figure \ref{fig:ablationFTjumppads}.

\subsection{Preliminary Analysis of Model Alignment with Scale}

To complete our ablation, we now investigate how pre-training and model size affects online alignment.
To do so we followed the procedure in Appendix \ref{app:procedure} to generated preferences as in Section \ref{subsec:preferences}. 
We then trained corresponding reward models and compare aligning these models trained from scratch to our pre-trained and fine-tuned model, as shown below in Figure \ref{fig:ablationFTalignment}.

\begin{figure}[h]
\centering
\vspace{0mm}
    \includegraphics[width=0.45\linewidth]{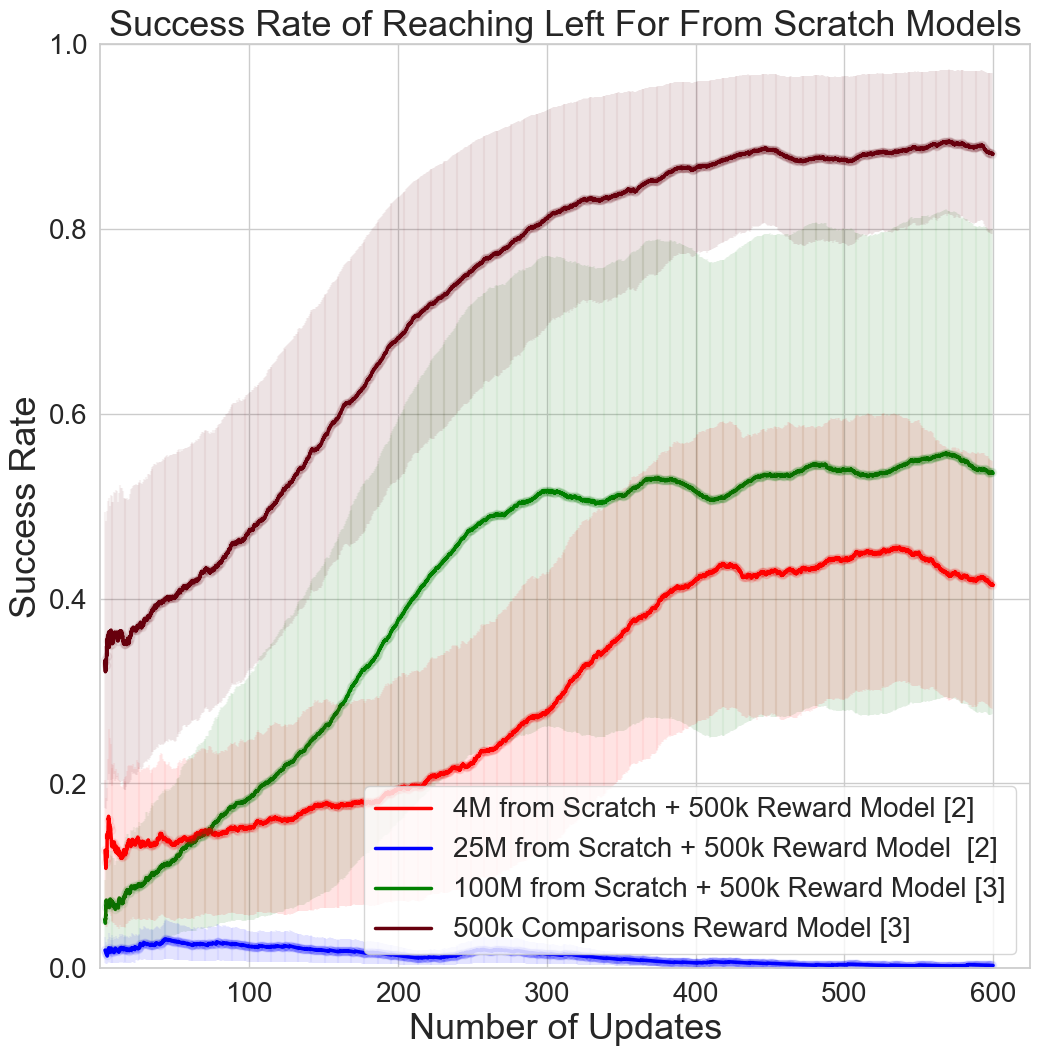}
    \includegraphics[width=0.45\linewidth]{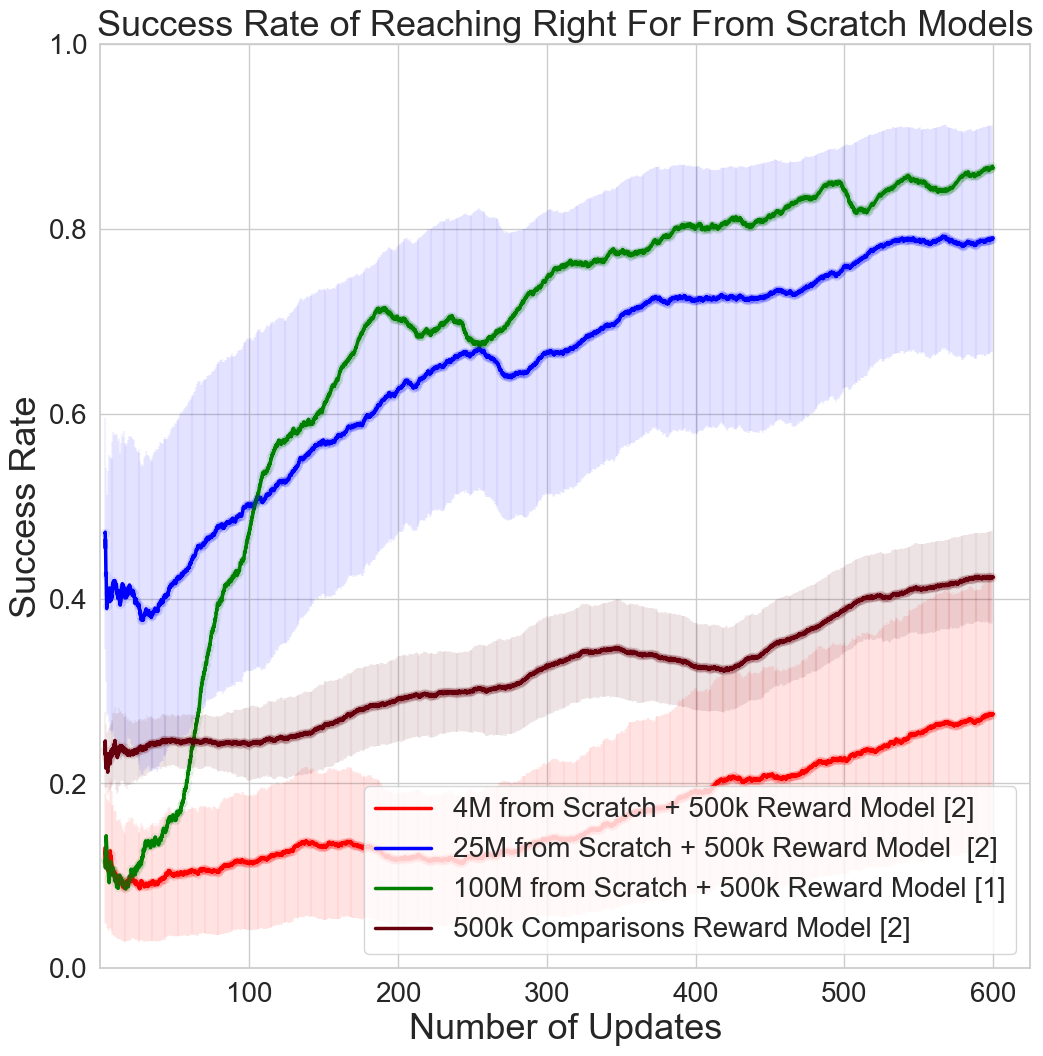}

      \vspace{0mm}
      \caption{Left and right jumppad success rate during online alignment for models of different sizes trained from scratch using corresponding reward models trained on 500k comparisons.}
\vspace{0mm}
\label{fig:ablationFTalignment}
\end{figure}

 Overall, the results are relatively high variance due to the small dataset these results are based on. For example, we see that our pre-trained model (brown) aligns significantly better than the equivalent size model trained from scratch (green) when aligned left, and worse when aligned right, but this is likely just due to the lack of diversity in the pre-trained+fine-tuned model as explained in Section \ref{sec:align_right}. However, bearing that in mind, we can see some general trends emerging. In particular, we see some indications that larger models generally align better (ignoring the left-aligned 25M model and the right-aligned pre-trained 100M model due to the substantial left/right bias associated with these models) and allowing for differences in initial performance due to minor left/right bias, since the success rate generally increases more quickly during alignment as model size increases.


 We also note that in this experiment we only considered using the original reward models trained on 500k comparisons, in order to see whether it was possible to successfully align a model trained solely on our small task-specific dataset.
 As noted in the previous section, these models have significantly less diversity in their behavior than the model first pre-trained and then fine-tuned.
 This makes it much more costly (in terms of human time) to generate sufficient labeled examples of the desired behavior to properly train the corresponding reward model (for example, less than $10\%$ of trajectories reach the left jumppad for the model trained from scratch compared to over $30\%$ for the pre-trained + fine-tuned model).
This means that while these results show an upper bound in terms of reward model performance, we would expect the larger and pre-trained models to have less degradation in their alignment when using reward models trained on smaller quantities of preference data, and therefore for larger pre-trained models to perform even better in more practical settings.
\label{app:ablation}
\newpage
\section{Reward Model Performances by Jumppad}

Test reward model performances against number of comparisons used for training by jumppad are shown below in Figure \ref{fig:reward_model_performance_by_jumppad}. 

\begin{figure}[h]
    \centering
      \includegraphics[width=0.49\textwidth]{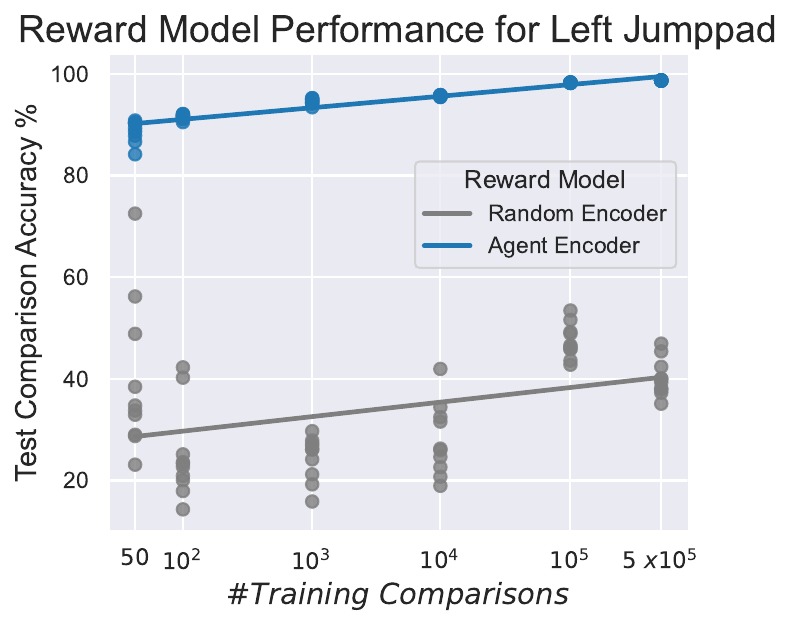}
    \includegraphics[width=0.50\textwidth]{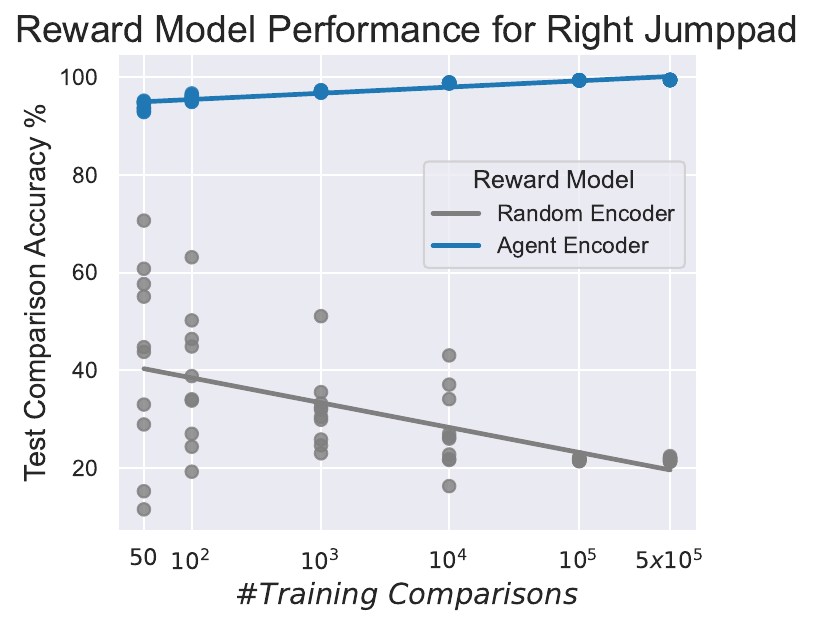}
      \caption[Results caption]%
      {Test reward model performances against number of training comparisons by jumppad.} \label{fig:reward_model_performance_by_jumppad}
\end{figure}

Both jumppads show the same trend for the agent encoder, reaching $\sim90\%$ performance for $100$ comparisons and $100\%$ performance by $500\text{k}$ comparisons. The random encoder is high variance, but generally increases in performance with training comparisons for the left jumppad, while decreasing in performance for the right jumppad. This surprising trend for the right jumppad could be because the reward model begins to overfit to the imbalanced trajectories discussed in Section \ref{sec:align_right} and shown in Figure \ref{fig:ft_agent_rollout_starts}. In other words, the reward model misinterprets the preferences and instead learns to prefer shorter trajectories that begin at the bottom left spawn point (from which 84\% of the successful right-going trajectories arise) leading to the $\sim20\%$ accuracy that we see for higher numbers of comparisons. The agent encoder based reward model is better able to distinguish these trajectories and infer the true underlying preference that better fits the data.\label{app:reward_models}
\section{Motivation for the Use of REINFORCE for Agent Alignment}

Our proof-of-concept uses REINFORCE \citep{williams_simple_1992} for online alignment of our agent, as discussed in Section~\ref{subsec:alignment}. While much of the RLHF literature uses PPO \citep{schulman_proximal_2017}, we note that recent work on LLMs has found PPO to be unnecessarily complex for RLHF \citep{ahmadian_back_2024}. This is due to the fact that in the standard RLHF framework, the reward is only provided at the end of the trajectory, meaning that we generally just want to reinforce entire trajectories rather than attempt credit assignment at each timestep using an advantage function, as is typically done with PPO. Additionally, \citet{ahmadian_back_2024} found that importance clipping rarely occurs in practice in RLHF, since the final aligned policy is kept close to the initial fine-tuned policy.
This is reflected by the use of REINFORCE in recent state-of-the-art LLMs \citep{google_deepmind_gemma_2024}. As a result, we use REINFORCE for simplicity in our proof-of-concept, but note that more complex tasks or feedback mechanisms may warrant the additional complexity of more modern reinforcement algorithms such as PPO.\label{app:reinforce}
\newpage
\section{Investigation into Discrepancy in Aligning Agent to Reach Right Jumppad}\label{app:basealignment}

\subsection{Understanding Discrepancy between Left and Right Jumppad Trajectories}

\begin{figure}[h]
\centering
\includegraphics[width=0.46\columnwidth]{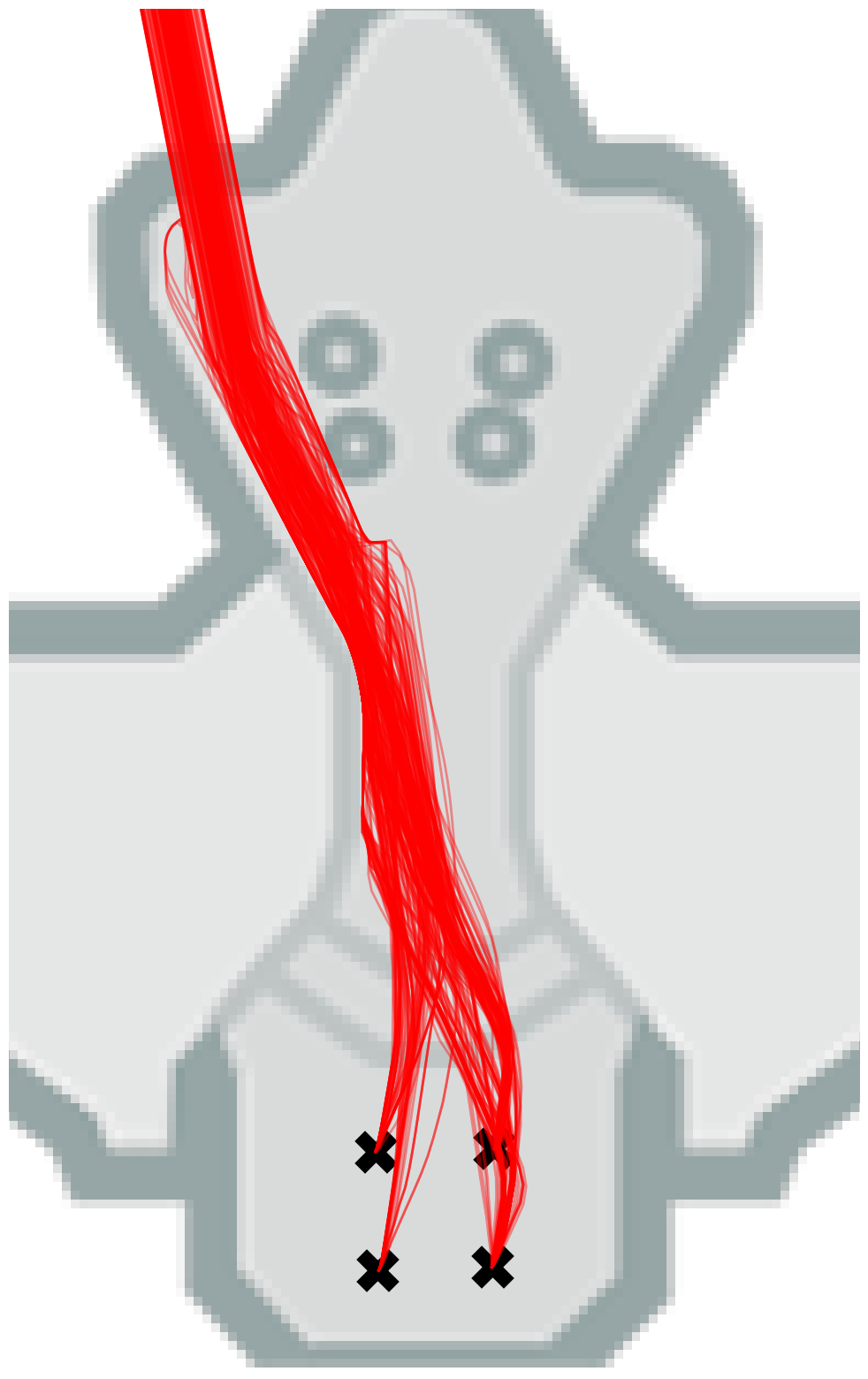}
\hspace{0.02\columnwidth} 
\includegraphics[width=0.46\columnwidth]{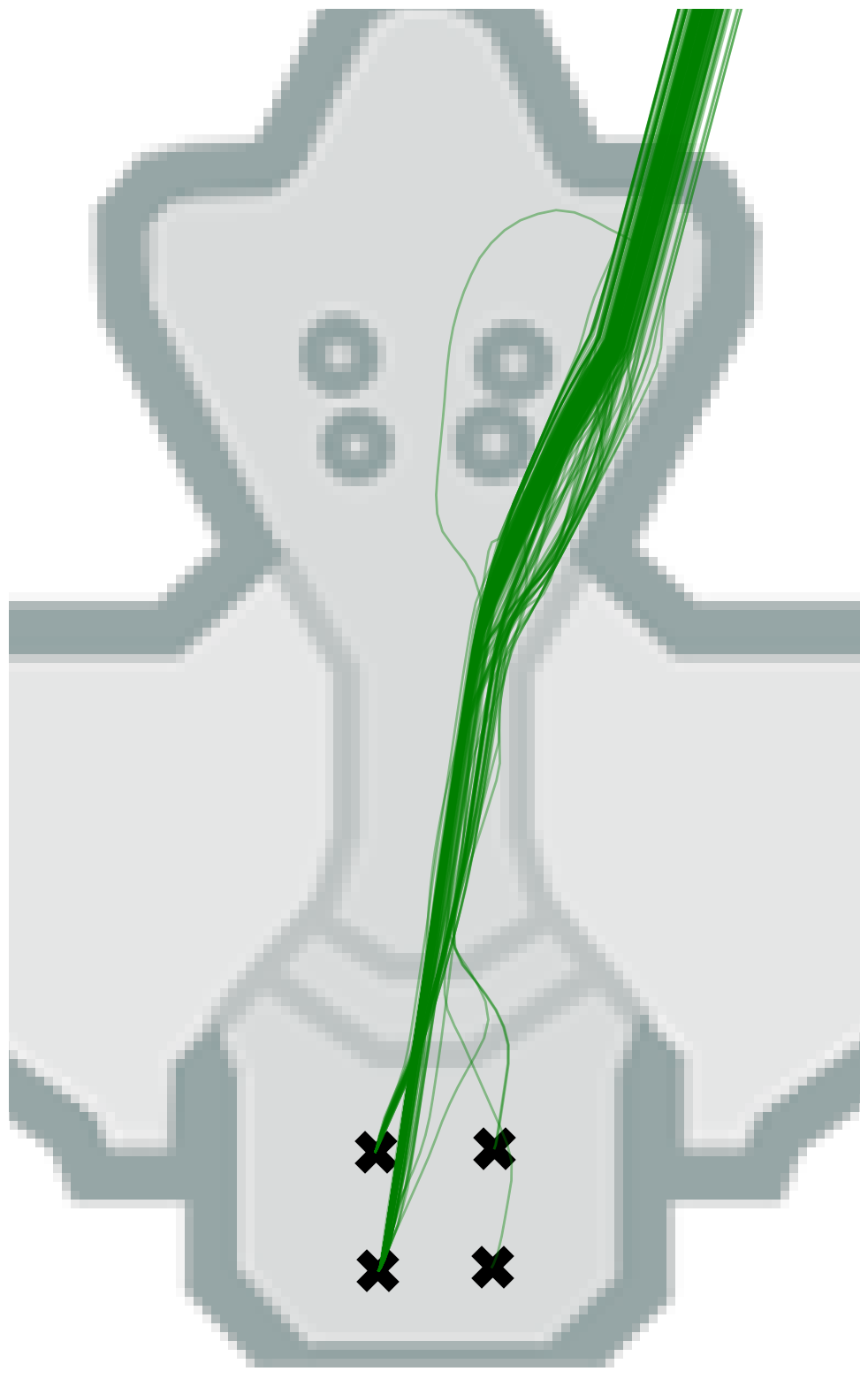}
\caption{Fine-tuned agent trajectories which reached the left (red) or right (green) jumppads. While fine-tuning enables more efficient preference labelling, it reduces diversity, potentially limiting alignment.} \label{fig:ft_agent_rollout_starts}
\end{figure}

Since the task of aligning our agent left or right is seemingly equivalent, we would expect to be able to align the agent similarly. Surprisingly however, we found that our fine-tuned agent did not align as quickly towards the right jumppad as the left, only reaching around a $\sim40\%$ success rate (in the absence of preference fine-tuning) after one day of training (as shown in Figure \ref{fig:Right_JP_FT_Alignment}) compared to around $\sim 90 \%$ for the left jumppad (as shown in Figure \ref{fig:Left_JP_FT_Alignment}). In order to investigate this discrepancy, we analyzed the rollouts of the fine-tuned agent, shown in Figure \ref{fig:ft_agent_rollout_starts}.

We see that the trajectories that reach the left jumppad start relatively evenly from the 4 spawn locations.
However, of the trajectories that successfully reach the right jumppad, only 1\% originated from the spawns on the right hand side. Therefore it is unsurprising that it is more difficult for the agent to learn to go right, given that in half the episodes (from the right spawn locations) the agent rarely navigates to the right jumppad to receive reward to reinforce its behavior. 
To confirm this hypothesis, we instead attempt to align the base agent (before fine-tuning).

\subsection{Alignment of Base Model for Comparison with Fine-Tuned Model}

To understand how trajectory diversity affects alignment, we now consider aligning the base agent using the same reward models (trained on the fine-tuned agent). By plotting the successful trajectories that reach the left and right jumppads we see that the base agent (left) has a greater diversity than the fine-tuned agent (right). However, we note that the success rate for reaching the left and right jumppads is lower for the base model (as in Figures \ref{fig:basejumppads} and \ref{fig:FTjumppads}), so there are fewer trajectories.

\begin{figure}[h]
    \centering
      \includegraphics[width=0.24\textwidth]{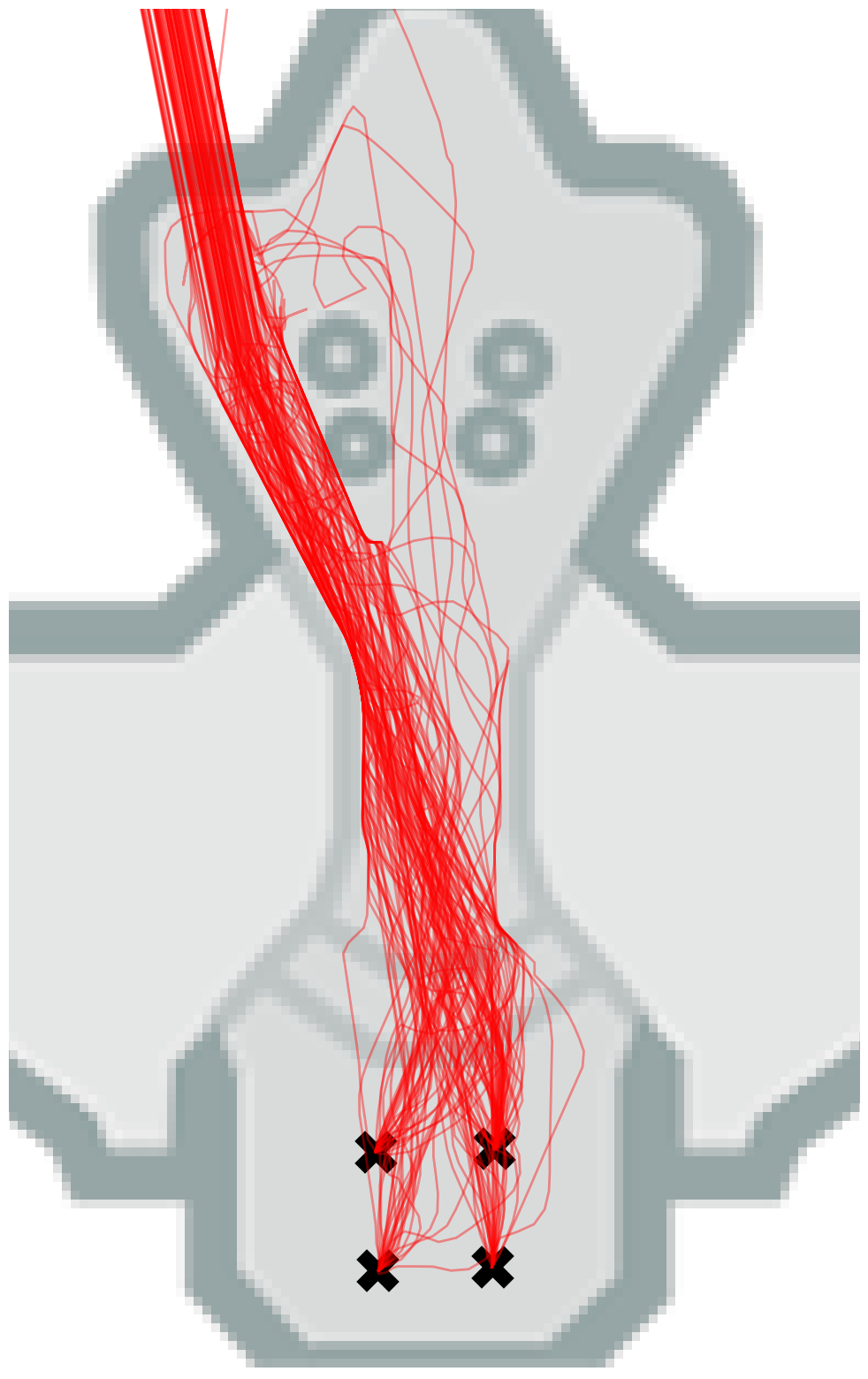}
      \includegraphics[width=0.24\textwidth]{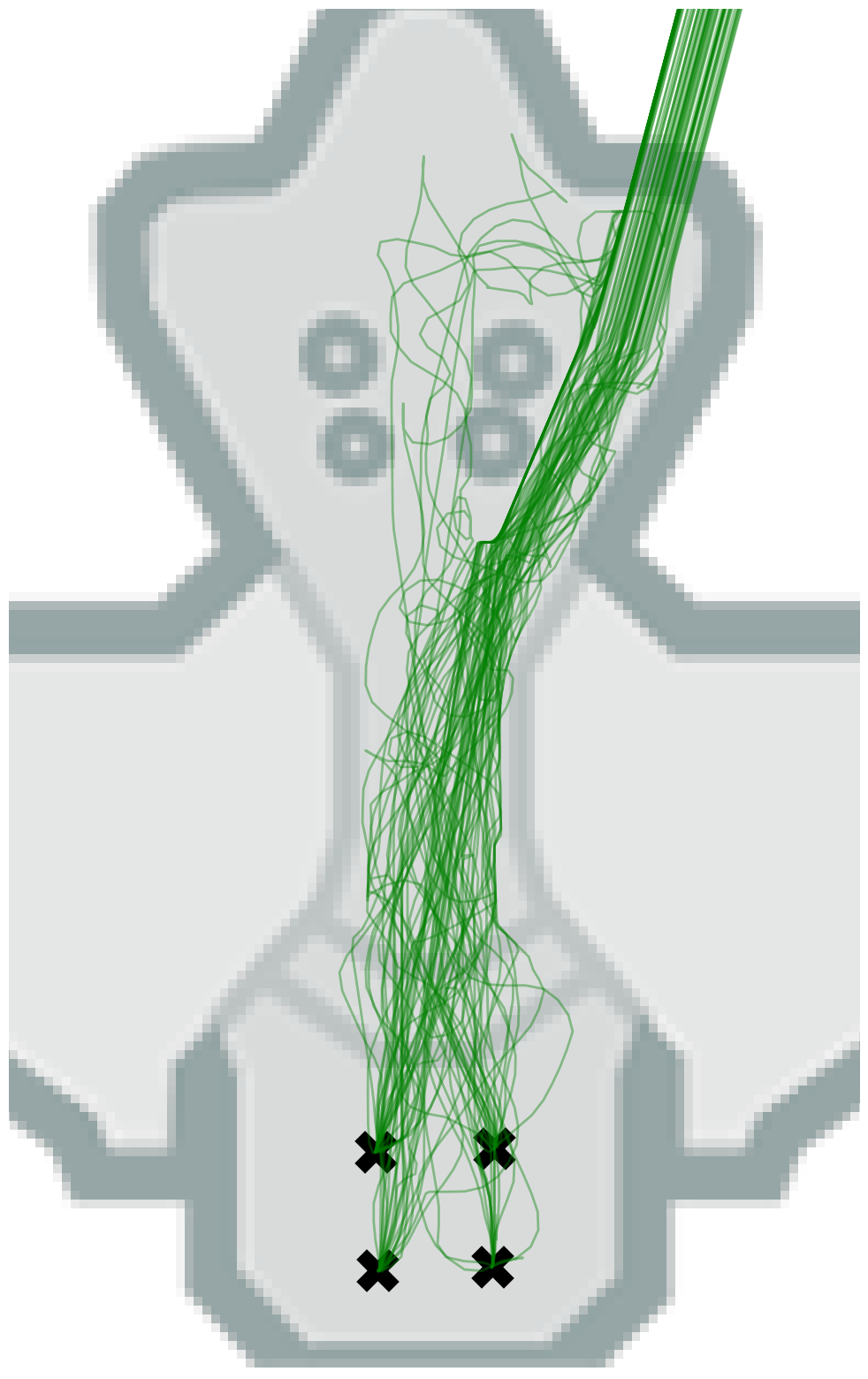}
      \includegraphics[width=0.24\textwidth]{figures/Train_FineTuned_JP1_trajectories.png}
      \includegraphics[width=0.24\textwidth]{figures/Train_FineTuned_JP3_trajectories.png}
      \caption[Results caption]%
      {Base agent (left) and fine-tuned agent (right) trajectories for which the agents successfully reached the left (red) or right (green) jumppads. While fine-tuning provides a greater success rate, enabling more efficient preference labelling, it also reduces diversity of trajectories.} \label{fig:base_ft_agent_rollout_starts}
\end{figure}

We now align the base agent to reach the left and right jumppads using the reward models trained on the fine-tuned agent, as shown below. We find that the agent can be aligned more symmetrically with both the left and the right jumppads, although alignment with the left jumppad still has a slightly better performance. However, in comparison to alignment of the fine-tuned agent (Figures \ref{fig:Left_JP_FT_Alignment} and \ref{fig:Right_JP_FT_Alignment}), we see that the alignment is much slower, starting at a lower success rate, not increasing as quickly during training, and reaching a lower final success rate. As before, we see the same general trend that higher accuracy reward models (using more preference data) lead to better alignment.

\begin{figure}[h]
\centering
\begin{minipage}{.49\textwidth}
  \centering
  \includegraphics[width=0.9\textwidth]{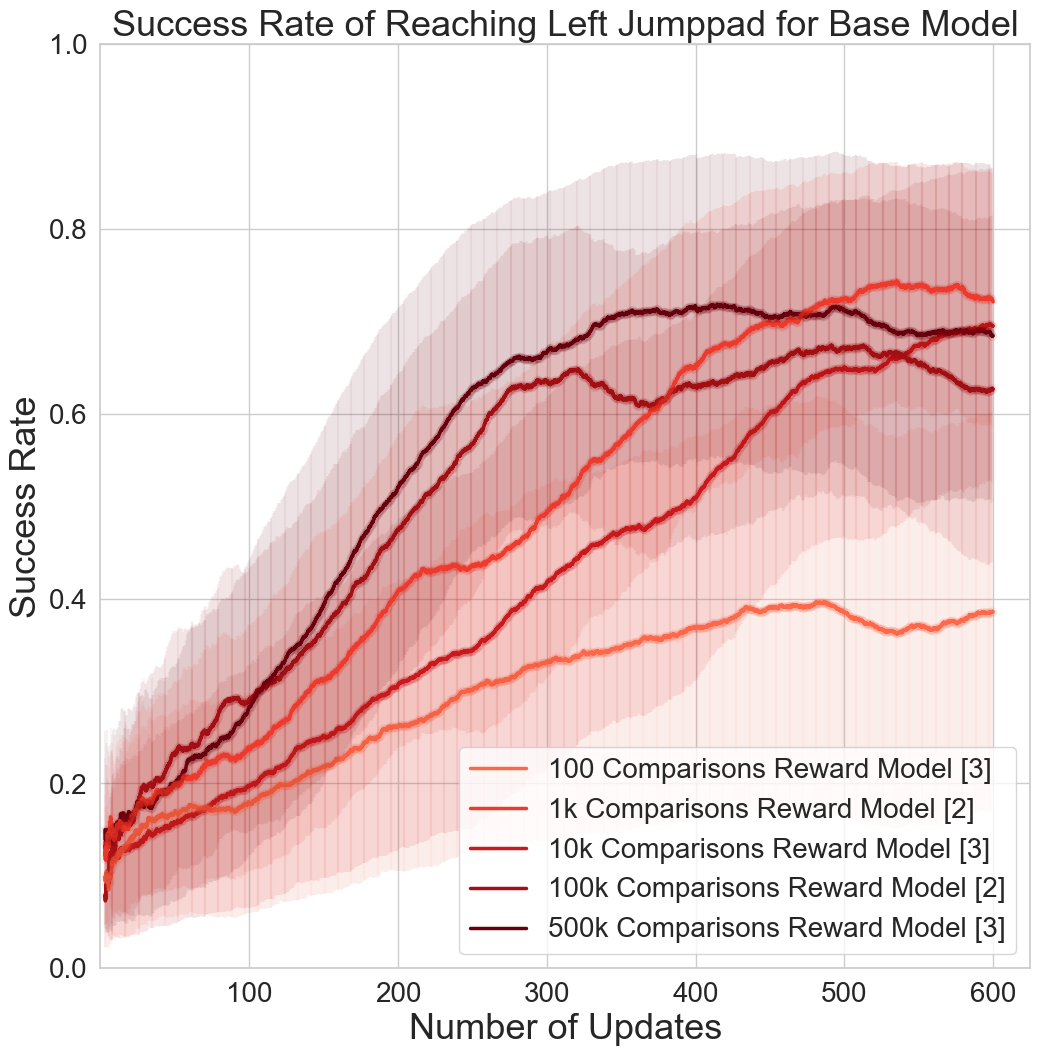}
  \captionsetup{width=0.5\textwidth}
  \caption[Results caption]%
  {Left jumppad success rate for base agent using\\ fine-tuned agent reward models.}
\label{fig:left_base_align}
\end{minipage}%
\begin{minipage}{.49\textwidth}
  \centering
  \includegraphics[width=0.9\textwidth]{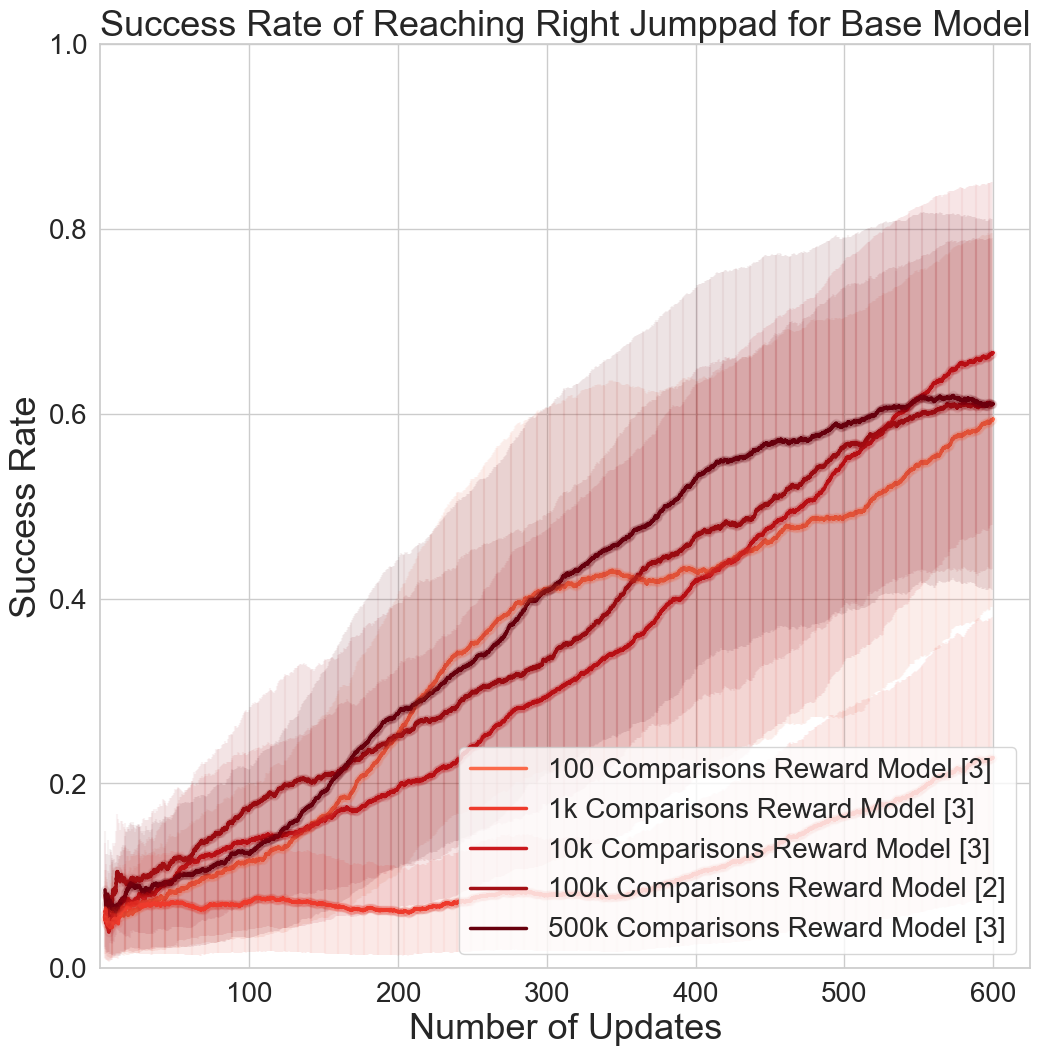}
  \captionsetup{width=0.5\textwidth}
\caption[Results caption]%
  {Right jumppad success rate for base agent using\\ fine-tuned agent reward models} \label{fig:right_base_align}
\end{minipage}
\end{figure}
\newpage
\section{Final Aligned Agent Jumppad Distributions}\label{app:alignedJPdistributions}
Jumppad distributions for our final agents aligned to go left and right using preference fine-tuning and online alignment with reward models trained on 500k preferences are shown below. 

\begin{figure}[h]
\centering
\begin{minipage}{.49\textwidth}
  \centering
  \includegraphics[width=\textwidth]{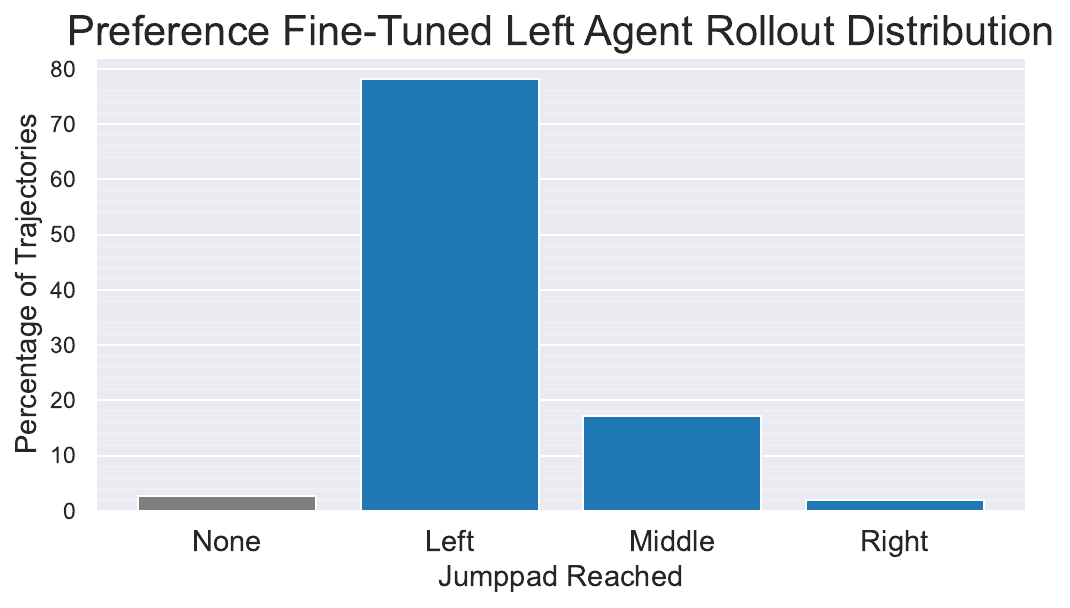}
\end{minipage}%
\begin{minipage}{.49\textwidth}
  \centering
  \includegraphics[width=\textwidth]{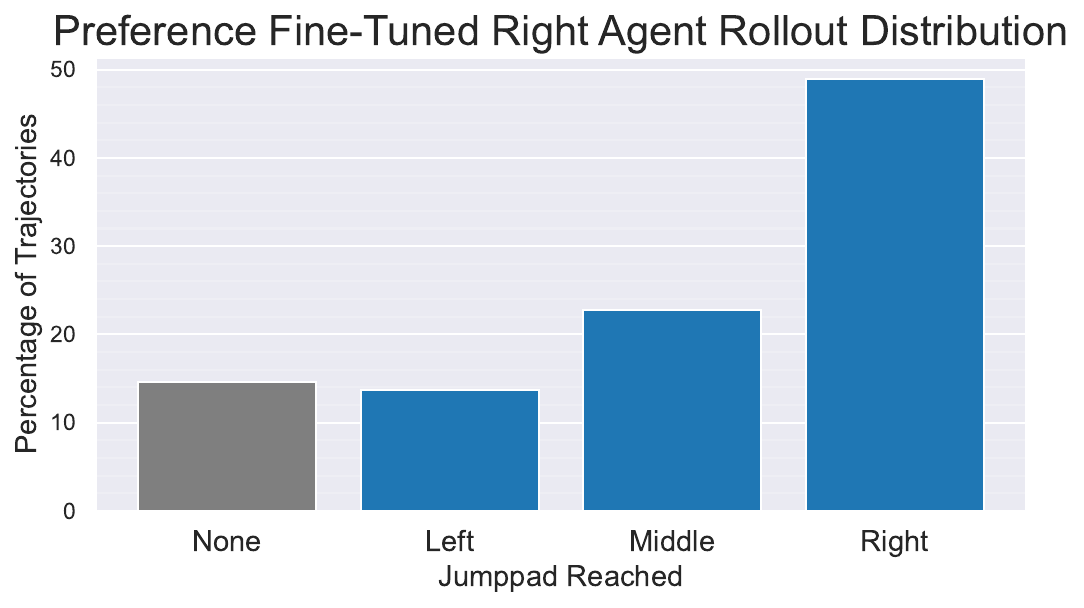}
\end{minipage}
\caption[Results caption]%
  {Left and right jumppad success rate for agents partially aligned with preference fine-tuning with reward models trained on 500k preferences.}\label{fig:preferenceFT_align_jumppads} 
\end{figure}

First we partially align our agents with preference fine-tuning using our (500k comparison) reward models, so that the behaviour distributions are closer to the desired behaviour distributions to reduce the alignment required online, as shown in Figure \ref{fig:preferenceFT_align_jumppads}.

We see that preference fine-tuning starts to align our agents towards the desired behaviour, but does not fully align our agents. Therefore we then perform online reinforcement learning using our (500k comparison) reward models until our agents are fully aligned. 

\begin{figure}[h]
\centering
\begin{minipage}{.49\textwidth}
  \centering
  \includegraphics[width=\textwidth]{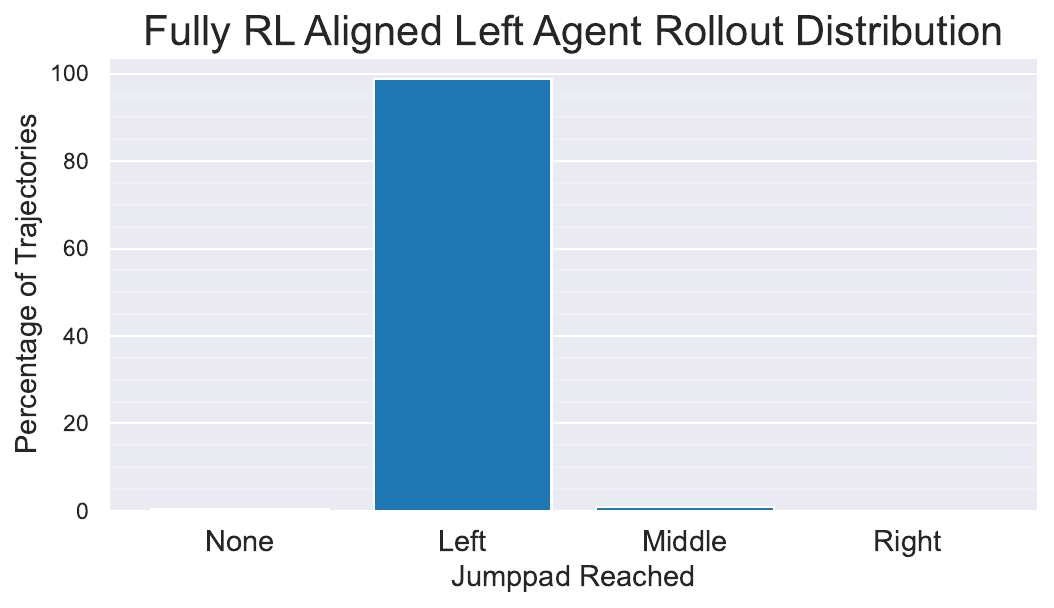}
\end{minipage}%
\begin{minipage}{.49\textwidth}
  \centering
  \includegraphics[width=\textwidth]{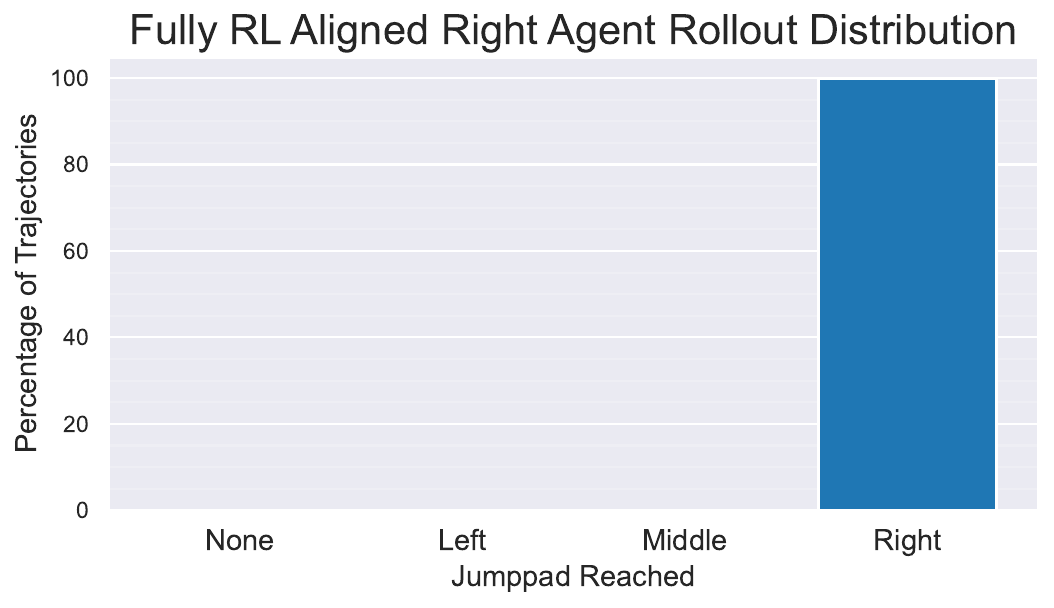}
\end{minipage}
\caption[Results caption]%
  {Left and right jumppad success rate for our fully aligned fine-tuned agents using preference fine-tuning and online reinforcement learning with reward models trained on 500k preferences.}\label{fig:RL_align_jumppads} 
\end{figure}
Final evaluation shows that our agents have now been effectively fully aligned with our desired behaviour.\label{app:finaljumppads}
\newpage
\newpage
\newpage
\section{Limitations and Further Work}

Our work provides a proof-of-concept of an approach for training embodied agents to act as desired in complex 3D environments. For this procedure to be applied in practice in video games and other domains such as robotics, the efficiency of this pipeline will be essential.

One limitation of our approach is that we assume access to a large amount of human data on the environment of interest to pre-train a base model. In practice however, when designing a game, large amounts of human data may not yet be available, and even for existing games it may not be possible to access. 
A potential solution may be to use a `gaming foundation model' trained on similar games (with pixel input and action controller output). Equivalently in robotics, the collection of general cross-objective and embodiment datasets is becoming more important for training general robotic agents \citep{embodimentcollaboration2024openxembodimentroboticlearning}. We believe the potential transfer of models pre-trained across environments will be an exciting direction for future work.


Investigating developments in LLM architectures and pre-training could also be of interest for further work. For example, the context window used by the transformer $H$ corresponds to the memory of the agent, so developments in increasing or augmenting the context window in LLMs such as Longformer attention \citep{beltagy_longformer_2020} or Retrieval-Augmented Generation (RAG) \citep{lewis_retrieval-augmented_2021}, could similarly be incorporated to increase the memory of agents trained with this approach, helping to address the issues with partial observability in complex 3D environments. 

Another limitation of our proof-of-concept that we discuss in Section \ref{subsec:preferences} is that we utilize synthetic preferences. While this may be a realistic option for many use cases, potentially using VLMs-as-a-judge \citep{zheng_judging_2023, li_llms-as-judges_2024}, real human feedback may be required in general. As well as potentially introducing noise to the preference labels, this process can be costly in human time, especially for agents on modern console games which must often be run in real-time in order to be rendered. Therefore supervised fine-tuning on relevant behaviors to improve preference labeling efficiency could help to reduce this human time requirement. However, knowledge transfer from training LLMs could again be relevant here. Recent work on providing human feedback for LLMs has used a hybrid form of feedback that combines preference and evaluative feedback, in which responses are grouped into a batch of size $N$ and simultaneously compared on a preference scale of size $P$, which enables larger numbers of comparisons to be extracted from a given number of responses, improving the time efficiency of providing feedback. For example, InstructGPT \citep{ouyang_training_2022} uses $4\leq N\leq9$, $P=7$ while Llama2 \citep{touvron_llama_2023} uses $N=2$, $P=4$. As $N$ and $P$ increase, more information can be extracted from the provided human feedback. For example, with $N=5$, $P=5$ and assuming no category duplication (no trajectories are considered equal), ${5\choose2} = 10$ comparisons can be extracted, which is equivalent to 2 bits of information per trajectory watched, compared to just 0.5 bits per trajectory for default pairwise comparisons with $N=2$, $P=2$. This results in a $4\times$ improvement in feedback efficiency for human labellers, at the cost of potential label noise due to the additional mental overhead required. Similar strategies could be applied to providing feedback to agents to make the process more time efficient.

A final limitation we consider in our proof-of-concept is the ability to run agents online, since embodied agents are much more challenging to run online than LLMs, even in the context of video games. For more complex behaviors and tasks this may become prohibitively expensive. Fortunately, developments in RLHF for language models can also be utilized here. We demonstrated in Section \ref{subsec:alignment} that an initial step of preference fine-tuning, corresponding to Reinforced Self-Training (ReST) \citep{gulcehre_reinforced_2023} with a single iteration, can improve alignment efficiency. However, the full procedure using multiple iterations could be used to further improve the sample efficiency of aligning with preferences while maintaining the benefits of online exploration. Other recent work has investigated aligning models completely offline. Direct Preference Optimization \citep{rafailov_direct_2023} provides an approach for optimizing a model to align with preferences without the need for a reward model or online training, while \citep{hu_aligning_2023} uses offline reinforcement learning with pre-generated samples and rewards. Additionally, efficient fine-tuning strategies such as LoRA \citep{hu_lora_2021} could be used to reduce the hardware requirements for the fine-tuning of a large base model for individual use-cases. These new approaches provide promising directions for further work in the context of agents to improve the efficiency or potentially even remove the need for online training completely. However, the extent to which active online learning is required for true generality is still an open question \citep{ostrovski_difficulty_2021} as we discuss in Appendix~\ref{app:o2oshift}.

In summary, we believe there is significant potential for transfer from recent developments in large language models to agents, that could help to address many long-standing limitations and open-problems for decision-making agents. We believe these will open up exciting new research directions in the coming years.\label{app:limitations}


\end{document}